\RequirePackage{fix-cm}
\documentclass[twocolumn]{svjour3}          
\smartqed  
\usepackage{bbding}
\usepackage{graphicx}
\usepackage{amssymb}
\usepackage[cmex10]{amsmath}
\usepackage{array}
\usepackage{url}
\usepackage{color}
\usepackage[tight,footnotesize]{subfigure}
\usepackage{natbib}
\usepackage{hyphenat}
\usepackage{multirow}
\journalname{International Journal of Computer Vision}
\definecolor{darkmag}{rgb}{0.55,0,0.55}
\begin{document}

\title{Facial Feature Point Detection: A Comprehensive Survey
}


\author{Nannan Wang         \and
        Xinbo Gao \and
        Dacheng Tao \and
        Xuelong Li
}



\institute{N. Wang \at
              VIPS Lab, School of Electronic Engineering, Xidian University, 710071, Xi'an, P. R. China \\
              \email{nannanwang.xidian@gamil.com}
           \and
           X. Gao \at
           VIPS Lab, School of Electronic Engineering, Xidian University, 710071, Xi'an, P. R. China \\
           \email{xbgao@mail.xidian.edu.cn}
           \and
           D. Tao \at
              Centre for Quantum Computation \& Intelligent Systems, Faculty of Engineering \& Information Technology, University of Technology, Sydney, 235 Jones Street, Ultimo, NSW 2007, Australia \\
              \email{Dacheng.Tao@uts.edu.au}
           \and
            X. Li \at
           School of Computer Science and Information Systems, Birkbeck College, University of London, Malet Street, London, WC1E 7HX, U.K. \\
           \email{xuelong@dcs.bbk.ac.uk}
}

\date{Received: date / Accepted: date}

\maketitle

\begin{abstract}
This paper presents a comprehensive survey of facial feature point detection with the assistance of abundant manually labeled images. Facial feature point detection favors many applications such as face recognition, animation, tracking, hallucination, expression analysis and 3D face modeling. Existing methods can be categorized into the following four groups: constrained local model (CLM)-based, active appearance model (AAM)-based, regression-based, and other methods. CLM-based methods consist of a shape model and a number of local experts, each of which is utilized to detect a facial feature point. AAM-based methods fit a shape model to an image by minimizing texture synthesis errors. Regression-based methods directly learn a mapping function from facial image appearance to facial feature points. Besides the above three major categories of methods, there are also minor categories of methods which we classify into \textit{other} methods: graphical model-based methods, joint face alignment methods, independent facial feature point detectors, and deep learning-based methods. Though significant progress has been made, facial feature point detection is limited in its success by wild and real-world conditions: variations across poses, expressions, illuminations, and occlusions. A comparative illustration and analysis of representative methods provide us a holistic understanding and deep insight into facial feature point detection, which also motivates us to explore promising future directions.
\keywords{Active appearance model \and active shape model \and constrained local face alignment \and facial feature point detection \and facial landmark localization}
\end{abstract}

\section{Introduction}
\label{sec:Introduction}

\begin{figure}[!t]
\centering
\includegraphics[width=0.5\columnwidth]{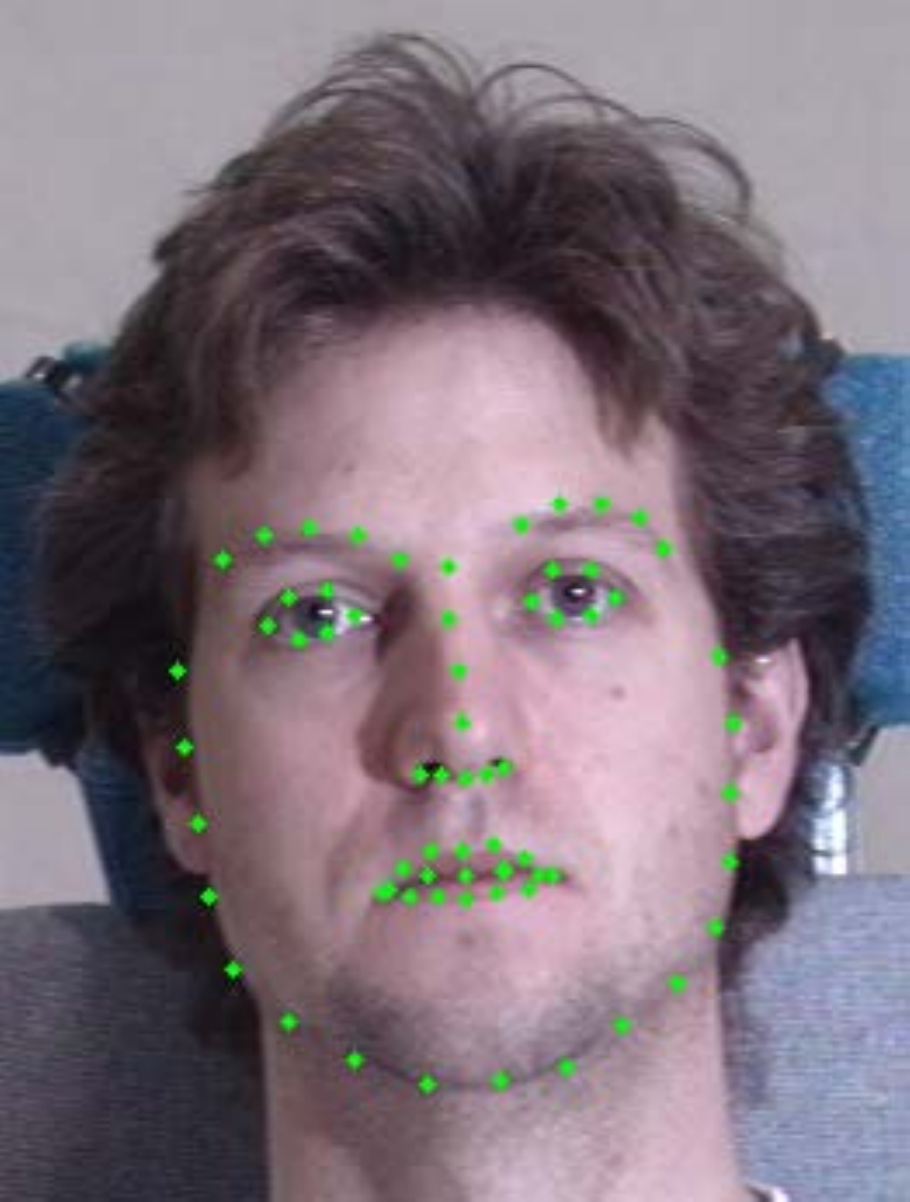}
\caption{Illustration of an example image with 68 manually labeled points from the Multi-PIE database \citep{DB2010MultiPIE}.}
\label{fig:Fig 1}
\end{figure}

{Facial} feature points, also known as facial landmarks or facial fiducial points, have semantic meaning. Facial feature points are mainly located around facial components such as eyes, mouth, nose and chin (see Fig. \ref{fig:Fig 1}). Facial feature point detection (FFPD) refers to a supervised or semi-supervised process using abundant manually labeled images. FFPD usually starts from a rectangular bounding box returned by face detectors \citep{RPIJCV2004Viola, RPPAMI2002Yang} which implies the location of a face. This bounding box can be employed to initialize the positions of facial feature points. Facial feature points are different from keypoints for image registration \citep{RPPAMI2010Ozuysal} and keypoint detection is usually an unsupervised procedure.

Suggested by \citep{PBCVIU1995Cootes}, facial feature points can be reduced to three types: points labeling parts of faces with application-dependent significance, such as the center of an eye or the sharp corners of a boundary; points labeling application-independent elements, such as the highest point on a face in a particular orientation, or curvature extrema (the highest point along the bridge of the nose); and points interpolated from points of the previous two types, such as points along the chin. According to various application scenarios, different numbers of facial feature points are labeled as, for example, a 17-point model, 29-point model or 68-point model. Whatever the number of points is, these points should cover several frequently-used areas: eyes, nose, and mouth. These areas carry the most important information for both discriminative and generative purposes. Generally speaking, more points indicate richer information, although it is more time-consuming to detect all the points.

The points shown in Fig. \ref{fig:Fig 1} can be concatenated to represent a shape $\textbf{x}=(x_1,\cdots,x_\textit{N},y_1,\cdots,y_\textit{N})^T$ where $(x_\textit{i},y_\textit{i})$ denotes the location of the $i$-th point and $\textit{N}$ is the number of points ($\textit{N}$ is 68 in this figure).  Given a sufficiently large number of manually labeled points and corresponding images as the training data, the target of facial feature point detection is to localize the shape of an input testing image according to the facial appearance. Detecting the shape of a facial image is a challenging problem due to both the rigid (scale, rotation, and translation) and non-rigid (such as facial expression variation) face deformation. FFPD generally consists of two phases: in the training phase, a model is learned from the appearance variations to the shape variations; and in the testing phase, the learned model is applied to an input testing image to localize facial feature points (shape). Normally the shape search process starts from a coarse initialization, following which the initial shape is moved to a better position step by step until convergence. According to the method of modeling the shape variation and the appearance variation, existing FFPD methods can be grouped into four categories: constrained local model (CLM)-based methods (here, the term CLM should be not confused with that in \citet{BMVC2006Cristinacce} which is a special case of CLM in our nomenclature), active appearance model (AAM)-based methods, regression-based methods and other methods.

CLM-based methods consider the appearance variation around each facial feature point independently. One response map can therefore be calculated from the appearance variation around each facial feature point with the assistance of a corresponding local expert. Facial feature points are then predicted from these response maps refined by a shape prior which is generally learned from training shapes. AAM-based methods model the appearance variation from a holistic perspective. In addition, both the shape and appearance variation model are usually constructed from a linear combination of some bases learned from training shapes and images. Regression-based methods estimate the shape directly from the appearance without learning any shape model or appearance model. There are also other FFPD methods which do not fall into any of the aforementioned categories and are classified into the category of 'other methods'. This category can be further divided into four sub-categories: graphical model-based methods, joint face alignment methods, independent facial feature point detectors, and deep learning-based methods. Table \ref{tab:table 01} and Fig.2 present the development timeline of the four categories of methods. As shown in the table and figure, the topic has attracted growing interest.

\begin{table*}
\centering
\caption{Development timeline of four categories of methods.}
\label{tab:table 01}
\scalebox{0.7}{
\begin{tabular}{|c|p{4cm}<{\centering}|p{4cm}<{\centering}|p{3cm}<{\centering}|p{2cm}<{\centering}|p{2cm}<{\centering}|p{2cm}<{\centering}|p{2cm}<{\centering}|}
\hline
\multirow{2}{*}{Year} & \multirow{2}{*}{CLM} & \multirow{2}{*}{AAM} & \multirow{2}{*}{Regression} & \multicolumn{4}{c|}{\textit{Other} Methods} \\
\cline{5-8}
& & & & GM & Joint & Independent & DL \\
\hline
1992 &[1] \citet{PBBMVC1992Cootes} & & & & & & \\
\hline
1993 &[2] \citet{PBBMVC1993Cootes}& & & & & & \\
\hline
1994 &[3] \citet{PBBMVC1994Cootes}& & & & & & \\
\hline
1995 &[4] \citet{PBCVIU1995Cootes}; [5] \citet{PBIVC1995Sozou}& & & & & & \\
\hline
1997 &[6] \citet{PBIVC1997Sozou}& & & & & & \\
\hline
1998 & &[7] \citet{PBECCV1998Cootes}; [8] \citet{PBBMVC1998Cootes}& & & & & \\
\hline
2001 & &[9] \citet{PBPAMI2001Cootes}; [10] \citet{PBICCV2001Cootes}; [11] \citet{PBCVPR2001Hou}& & & & & \\
\hline
2002 & &[12] \citet{PBIVC2002Cootes} & &[13] \citet{PBECCV2002Coughlan} & & & \\
\hline
2003 &[14] \citet{PBBMVC2003Cristinacce}; [15] \citet{PBCVPR2003Zhou}&[16] \citet{PBCVPR2003Batur} & & & & & \\
\hline
2004 &[17] \citet{PBFG2004Cristinacce}; [18] \citet{PBBMVC2004Cristinacce}&[19] \citet{PBIJCV2004Matthews} & & & & & \\
\hline
2005 & &[20] \citet{PBTIP2005Batur}; [21] \citet{PBIVC2005Gross}& & & &[22] \citet{PBSMC2005Vukadinovic} & \\
\hline
2006 &[23] \citet{FG2006Cristinacce}; [24] \citet{BMVC2006Cristinacce}&[25] \citet{BMVC2006Cootes}; [26] \citet{ECCV2006Dedeoglu}; [27] \citet{PAMI2006Donner}; [28] \citet{BMVC2006Liu}& & [29] \citet{CVPR2006Liang}; [30] \citet{ECCV2006Liang}& & & \\
\hline
2007 &[31] \citet{BMVC2007Cristinacce}; [32] \citet{PAMI2007sukno}; [33] \citet{ICCV2007Vogler}&[34] \citet{ICCV2007Gonzalez}; [35] \citet{CVPR2007Kahraman}; [36] \citet{IJCV2007Matthews}; [37] \citet{BMVC2007Peyras}; [38] \citet{BMVC2007Roberts}; [39] \citet{ICCV2007Saragih}; [40] \citet{IJCV2007Sung}&[41] \citet{IPMI2007Zhou} &[42] \citet{ICCV2007Huang} & & & \\
\hline
2008 &[43] \citet{PR2008Cristinacce}; [44] \citet{ECCV2008Gu}; [45] \citet{ECCV2008Liang}; [46] \citet{ECCV2008Milborrow}; [47] \citet{CVPR2008Wang}; [48] \citet{FG2008Wang}; [49] \citet{PAMI2008Wimmer}&[50] \citet{CVPR2008Nguyen}; [51] \citet{FG2008Nguyen}; [52] \citet{CVPR2008Papandreou}; [53] \citet{FG2008Saragih}; [54] \citet{IJCV2008Sung}&[55] \citet{CVPR2008Kozakaya}; [56] \citet{FG2008Kozakaya}& & &[57] \citet{CVPR2008Ding} & \\
\hline
2009 &[58] \citet{CVPR2009Li}; [59] \citet{IVC2009Lucey}; [60] \citet{CVPR2009Paquet}; [61] \citet{ICCV2009Saragih02}; [62] \citet{ICCV2009Saragih01}; [63] \citet{ICCV2009Saragih03}; [64] \citet{BMVC2009Tresadern}&[65] \citet{CVPR2009Brian}; [66] \citet{CVPR2009Asthana}; [67] \citet{ICCV2009Hamsici}; [68] \citet{PAMI2009Lee}; [69] \citet{PAMI2009Liu}; [70] \citet{PR2009Saragih}& & &[71] \citet{CVPR2009Tong} &[72] \citet{PR2009Asteriadis} & \\
\hline
2010 & &[73] \citet{CVPR2010Ashraf}; [74] \citet{BMVC2010Martins}; [75] \citet{IJCV2010Nguyen}; [76] \citet{BMVC2010Tresadern}&[77] \citet{FG2008Kozakaya}; [78] \citet{CVPR2010Valstar}& & &[79] \citet{PAMI2010Ding} & \\
\hline
2011 &[80] \citet{CVPR2011Belhumeur}; [81] \citet{FG2011Chew}; [82] \citet{PAMI2011Li}; [83] \citet{FG2011Roh}; [84] \citet{CVPR2011Saragih}; [85] \citet{IJCV2011Saragih}&[86] \citet{PR2011Asthana}; [87] \citet{BMVC2011Hansen}; [88] \citet{ICCV2011Navarathna}; [89] \citet{BMVC2011Sauer}&[90] \cite{BMVC2011Kazemi} & &[91] \citet{JACVPR2011Zhao} & & \\
\hline
2012 &[92] \citet{CVPR2012Baltrusaitis}; [93] \citet{ECCV2012Cootes}; [94] \citet{ECCV2012Le}; [95] \citet{ECCV2012Martins}; [96] \citet{BMVC2012Martins}&[97] \citet{CVIU2012Huang}; [98] \citet{ECCV2012Kinoshita}; [99] \citet{IJCV2012Tresadern}; [100] \citet{ACCV2012Tzimiropoulos}&[101] \citet{CVPR2012Cao}; [102] \citet{CVPR2012Dantone}; [103] \citet{PR2012Rivera}; [104] \citet{ECCV2012Lozano}; [105] \citet{ACCV2012Yang}&[106] \citet{VISAPP2012Michal}; [107] \citet{CVPR2012Zhu}&[108] \citet{JAECCV2012Smith}; [109] \citet{JACVIU2012Tong}; [110] \citet{JAECCV2012Zhao}& &[111] \citet{CVPR2012Luo} \\
\hline
2013 &[112] \citet{CVPR2013Asthana}; [113] \citet{PAMI2013Belhumeur}; [114] \citet{ICCV2013Yu}&[115] \cite{CVPR2013Anderson}; [116] \citet{FG2013Fanelli}; [117] \citet{CVIU2013Martins}; [118] \citet{ICCV2013Tzimiropoulos}; [119] \citet{PAMI2013Lucey}&[120] \citet{ICCV2013Burgos}; [121] \citet{PAMI2013Martinez}; [122] \citet{CVPR2013Xiong}; [123] \citet{ICCV2013Yang}&[124] \citet{ICCV2013Zhao} & &[125] \citet{CVPR2013Shen} &[126] \citet{CVPR2013Smith}; [127] \citet{CVPR2013Sun}; [128] \citet{CVPR2013Wu}\\
\hline
\end{tabular}}
\\
Note: Representative methods after 2006 are surveyed very carefully. Important works before 2006 are also included.
In the table, "GM" denotes the sub-category of graphical model-based methods, "Joint" represents the sub-category of joint face alignment, "Independent" means the sub-category of independent facial feature point detectors, and "DL" is the abbreviation of deep learning.
\end{table*}

\begin{figure*}
\centering
\includegraphics[width=1.8\columnwidth]{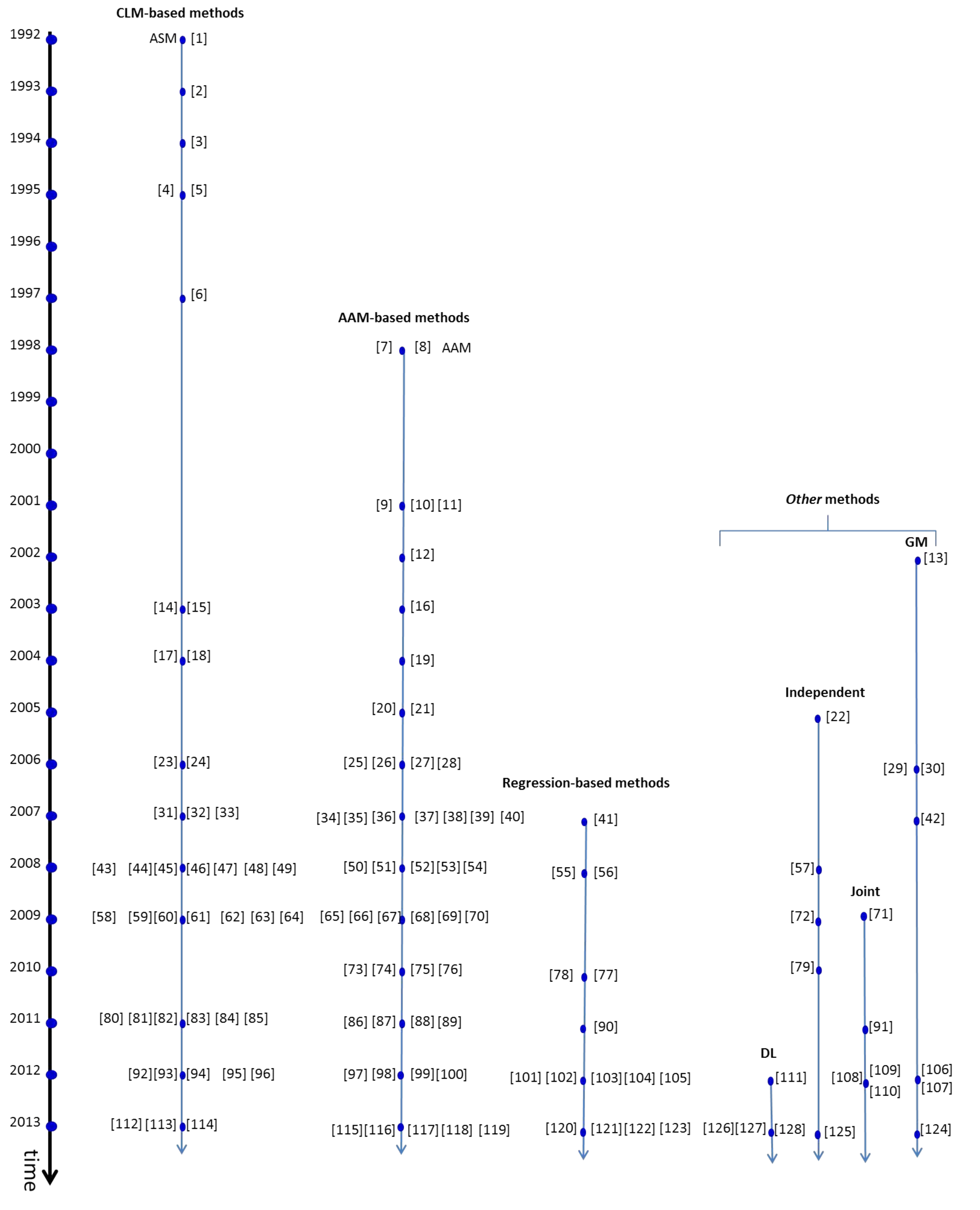}
\caption{Development timeline of four categories of methods. The reference number in this figure is indexed according to Table \ref{tab:table 01}.}
\label{fig:Fig 2}
\end{figure*}

Many related research topics and real-world applications could benefit from the accurate detection of facial feature points. Lee and Kim \citep{PAMI2009Lee} explored the fitted shape and shape-normalized appearance of the proposed tensor-based active appearance model (AAM) \citep{PBECCV1998Cootes} to transform the input image into a normalized image (frontal pose, neural expression, and normal illumination) to conduct variation-robust face recognition. \citet{PBTMI2003Stegmann} applied AAM to medical image analysis. \citep{APAMI2005Zhou} proposed a fusion strategy to incorporate subspace model constraints for robust shape tracking. \citet{AICCV2001Chen} applied active shape model (ASM) \citep{PBBMVC1992Cootes} to separate the shape from the texture to favor the sketch generation process. FFPD for face alignment is an essential preprocessing step in face hallucination \citep{AIJCV2013Wang} and facial swapping \citep{ASIGGRAPH2008Bitouk}. Facial animation \citep{ASIGGRAPH2011Weise} generally detects facial feature points to control the variation of facial appearance. The combination of 2D and 3D view-based AAM is utilized to robustly describe the variation of facial expression across different poses \citep{ASMCA2008Sung}. The correspondence of facial feature points plays an important role in 3D face modeling \citep{RPSIGGRAPH1999Blanz}. \citet{CVPR2013Anderson} applied AAM to track robustly and quickly over a very large corpus of expressive facial data and to synthesize video realistic renderings in the visual text-to-speech system.

Table \ref{tab:table 1} shows general notations commonly appearing in this paper. The remainder of this paper is organized as follows: Sections \ref{sec:2} to \ref{sec:5} investigate the aforementioned four categories of methods, respectively. Section \ref{sec:6} evaluates and analyzes the performance of several representative methods. Finally, Section \ref{sec:7} summarizes the paper, and discusses some promising future directions and tasks regarding FFPD.

\begin{table*}[!t]
\centering
\caption{Notations}
\label{tab:table 1}
\begin{tabular}{|c|c|}
\hline
Symbols & Descriptions \\
\hline
${\textit{N}}$ & The number of landmarks labeled in each image \\
\hline
${\textit{m}}$ & The number of principal modes in texture models of AAM \\
\hline
$\textit{n}$ & The number of principal modes in the point distribution model \\
\hline
$\pmb{\alpha}=(\alpha_1,\cdots,\alpha_i,\cdots,\alpha_N)^T$ & Shape parameters in the point distribution model (PDM)\\
\hline
$\pmb{\beta}=(\beta_1,\cdots,\beta_i,\cdots,\beta_N)$ & Texture parameters in the texture model of AAM \\
\hline
$\lambda_i$ & The $i$-th eigenvalue in the PDM \\
\hline
$\textbf{P}_s$ & Shape projection matrix in PDM \\
\hline
$\textbf{P}_a$ & Texture projection matrix in the texture model of AAM \\
\hline
$\textbf{I}$ & An input testing image \\
\hline
$\textbf{I}^{(e)}$ & Identity matrix whose order is determined in the context \\
\hline
$\textbf{x}$ & A shape represented in the image frame\\
\hline
$(x_i,y_i)^T$ or $\textbf{x}_{(i)}$ & Coordinate of the i-th point in the image frame \\
\hline
$\textbf{s}$ & A shape represented in the reference (mean-shape) frame\\
\hline
$\textbf{s}_i$ & The $i$-th shape basis in the PDM\\
\hline
$\textbf{a}$ & A texture representation in the reference frame \\
\hline
$\textbf{a}_i$ & The $i$-th texture basis in the texture model of AAM \\
\hline
$\textbf{s}_0$ & The mean shape \\
\hline
$\textbf{a}_0$ & The mean texture in the reference frame \\
\hline
$\textbf{c}$ & Appearance parameters in AAM \\
\hline
$s$ & Sclae in the rigid transformation \\
\hline
$\textbf{R}$ & $2\times2$ rotation matrix with orientation $\theta$ \\
\hline
$\textbf{t}$ & $2\times1$ translation vector in rigid transformation \\
\hline
$\textbf{q}$ & Pose parameters ($4\times1$ vector: $s$, $\theta$, and $\textbf{t}$)\\
\hline
$\textbf{p}$ & Rigid and non-rigid shape parameters $(\textbf{p}=(\textbf{q}^T;\pmb{\alpha}^T)^T)$ \\
\hline
\end{tabular}
\end{table*}

\section{Constrained Local Model-Based Methods}
\label{sec:2}
CLM-based methods fit an input image for the target shape through optimizing an objective function, which is comprised of two terms: shape prior $\mathcal{R}(\textbf{p})$ and the sum of response maps $\mathcal{D}_i{(\textbf{x}_i;\textbf{I})}$, $(i=1,\cdots,N)$ obtained from $N$ independent local experts \citep{IJCV2011Saragih}:

\begin{equation}
\label{eq 1}
\mathrm{min}_{\textbf{p}}\mathcal{R}(\textbf{p})+\sum_{i=1}^{N}\mathcal{D}_i{(\textbf{x}_i;\textbf{I})}.
\end{equation}

A shape model is usually learned from training facial shapes and is taken as the prior refining the configuration of facial feature points. Each local expert is trained from the facial appearance around the corresponding feature point and is utilized to compute the response map which measures detection accuracy. The CLM objective function in the equation (\ref{eq 1}) can be interpreted from a probabilistic perspective:

\begin{equation}
\label{eq 2}
\mathrm{max}_{\textbf{p}}p(\textbf{p})\prod_{i=1}^N{p(l_i=1|\textbf{x}_i,\textbf{I})}.
\end{equation}
where $l_i\in\{1,-1\}$ indicates whether the $i$-th point is aligned or misaligned, $\mathcal{R}(\textbf{p})=-ln\{p(\textbf{p})\}$ and $\mathcal{D}_i(\textbf{x}_i;\textbf{I})=-ln{p(l_i=1|\textbf{x}_i,\textbf{I})}$. The CLM objective function (either (\ref{eq 1}) or (\ref{eq 2})) implicitly assumes that $N$ response maps are calculated independently.

In the offline phase, a shape model and local experts should be learned from training shapes and corresponding images. Then in the online phase, given an input image, the output shape can be solved from the optimization of equation (\ref{eq 1}). We will investigate commonly used shape models and local experts sequentially. Finally, methods on how to combine the shape model and local experts for optimization are investigated.

\subsection{Shape Model}
Fig. \ref{fig:Fig 3} illustrates the statistical distribution of facial feature points sampled from 600 facial images. Regarding the shape prior, multivariate Gaussian distribution is commonly assumed, otherwise known as the point distribution model (PDM) proposed by Cootes and Taylor \citep{PBBMVC1992Cootes}:

\begin{figure}[!t]
\centering
\includegraphics[width=0.8\columnwidth]{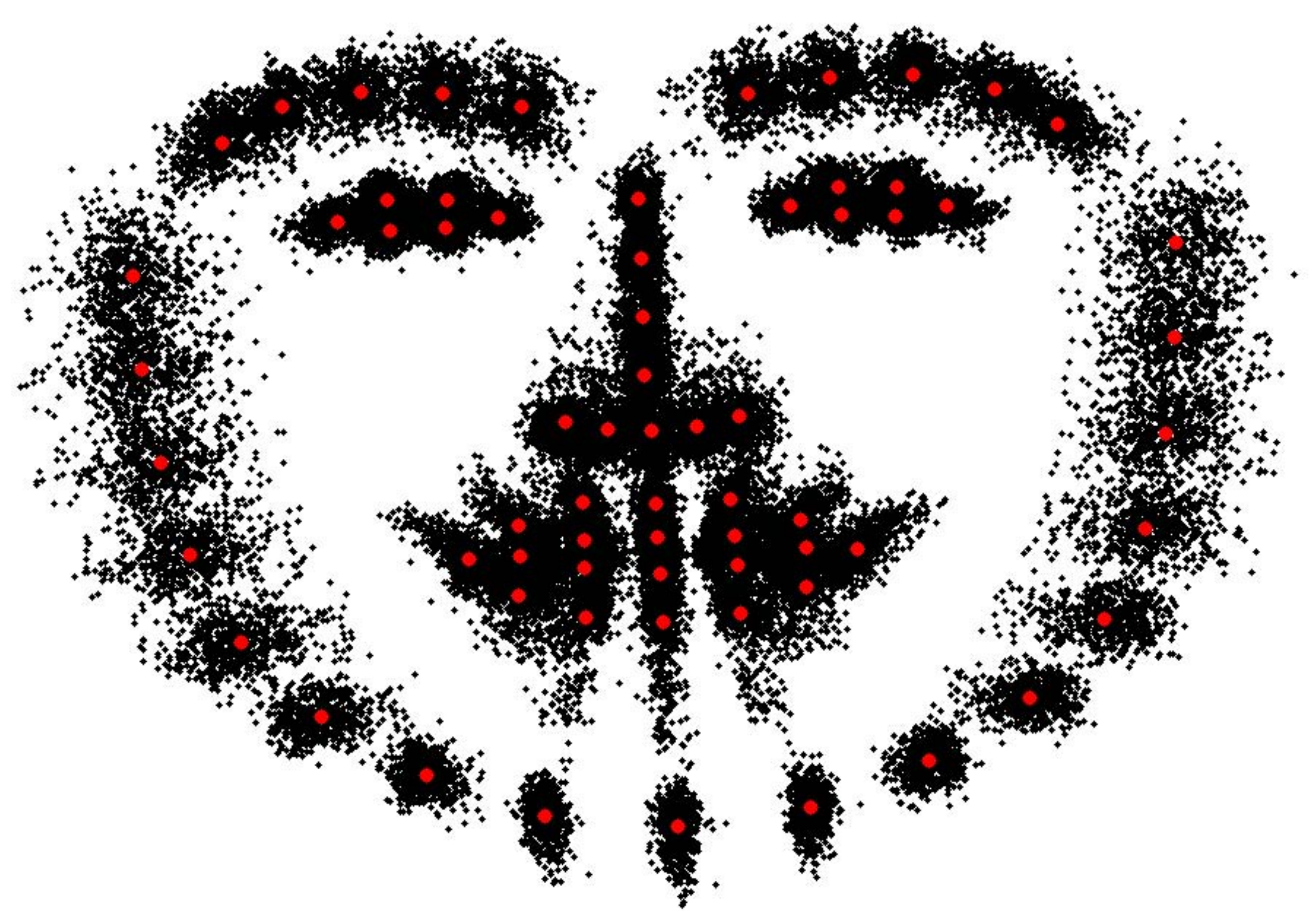}
\caption{Illustration of statistical distribution of facial feature points. There are 600 shapes (smaller dot points in black) normalized by Procrustes analysis. The larger dot points in red indicate the mean shape of all shapes.}
\label{fig:Fig 3}
\end{figure}

\begin{equation}
\label{eq 3}
\mathbf{s}=\textbf{s}_0+\textbf{P}_s\pmb{\alpha}=\textbf{s}_0+\sum_{i=1}^n\alpha_i\textbf{s}_i,
\end{equation}
where $\textbf{s}_i (i=0,\cdots,n)$ can be estimated by the principal component analysis (PCA) on all aligned training shapes. Actually, $\textbf{s}_0$ is the mean of all these shapes and $\textbf{s}_1, \cdots, \textbf{s}_n$ are the eigenvectors corresponding to the $n$ largest eigenvalues of the covariance matrix of all aligned training shapes. $n$ is usually determined by preserving $90\%\sim98\%$ variance (the ratio between the sum of n largest eigenvalues and sum of all eigenvalues). \citet{ECCV2008Mei} suggested the above rule to determine whether the value of $n$ is reliable or not and further explored bootstrap stability analysis to improve reliability. To remove the effect of rigid transformation, all training shapes are aligned by Procrustes analysis before learning the shape model. We call this rigid transformation-free shape $\textbf{s}$ in a reference frame. We apply rigid transformation to $\textbf{s}$ to generate a shape $\textbf{x}$ in the image frame:

\begin{equation}
\label{eq 4}
\mathbf{x}_{point}=s\textbf{R}\textbf{s}_{point}+\textbf{t}_{point},
\end{equation}
where $\textbf{t}_{point}$ consists of $N$ replications of $\textbf{t}$ and $\textbf{s}_{point}$ denotes a rearranged $2\times{N}$ matrix with each column corresponding to one point in $\textbf{s}$. Similarly, $\textbf{x}$ is the rearrangement of $\textbf{x}_{point}$.

An eigenspace shown in equation (\ref{eq 3}) can be represented by a quadruple: mean vector, matrix of eigenvectors, eigenvalues, and the number of observations to construct the eigenspace. Eigenspace fusion \citep{RPPAMI2000Hall} merges two eigenspaces into one eigenspace, which has great significance for online updating. Butakoff and Frangi \citep{PAMI2006Butakoff} generalized the eigenspace fusion model \citep{RPPAMI2000Hall} to a weighted version and applied it to merge multiple ASMs (or AAMs). Their experimental results show that fused ASMs have similar performance to full ASMs (model constructed from full set of observations) in terms of both segmentation error and time cost. They also applied the above fusion model to multi-view face segmentation \citep{CVIU2010Butakoff}, which can be casted as a two-model fusion problem: the fusion of a frontal view model and a left profile model; and the fusion of a frontal view model and a right profile model. Faces in intermediate view can be interpolated through fusion weight estimation.

In addition to PDM, there are several improvements on the prior shape distribution. Considering that PCA can only model the linear structure of shapes, Cootes et al. \citep{PBIVC1995Sozou, PBIVC1997Sozou} generalized the linear PDM to a nonlinear version by exploring polynomial regression and multi-layer perceptron respectively. \citet{CVPR2006Gu} proposed a 3D face alignment method in a single testing image based on a 3D PDM. To project the 3D shapes to the 2D plane, a weak perspective projection is assumed between the 3D space and the 2D plane. \citet{CVPR2008De} proposed a kernel PCA-based nonlinear shape model. Since a single Gaussian is inadequate for modeling the distribution over facial feature points, a mixture of Gaussian has been explored \citep{PBIVC1999Cootes, BMVC2006Everingham, CVPR2009Sivic}. In PDM (see equation (\ref{eq 3})), shapes are constrained on the subspace spanned by principal components. \citet{CVPR2011Saragih} exploited the principal regression analysis to span a constrained subspace. Since PDM assumes Gaussian observation noise and learns a shape model using all the training data, it is vulnerable to gross feature detection errors due to partial occlusions or spurious background features. Li et al. \citep{CVPR2009Li, PAMI2011Li} thus presented a robust shape model exploring random sample consensus \citep{RPCACM1981Fischler}. This method is in a hypothesis-and-test form, i.e. given a set of hypotheses, the one that satisfies certain optimal conditions should be chosen. Object shape and pose hypotheses (parameters) are first generated from randomly sampled partial shape-subsets of feature points. Subsequently, the hypotheses are tested to find the one that minimizes the shape prediction error.

Besides the above explicit shape models, \citet{PBBMVC2004Cristinacce} proposed an implicit shape model known as pairwise reinforcement of feature responses, which models a shape by learning the pairwise distribution of all ground truth feature point locations relative to the optimal match of each corresponding individual feature detector.

\subsection{Local Expert}
A local expert functions to compute a response map on the local region around corresponding facial feature points, i.e. we have $N$ local experts in a FFPD model. The region that supports a local expert could be either one-dimensional (i.e. a line) or two-dimensional (such as a rectangular region). A local expert can be a distance metric such as the Mahalanobis distance \citep{PBCVIU1995Cootes}, a classifier such as linear support vector machine \citep{CVPR2008Wang}, or a regressor \citep{BMVC2007Cristinacce, ICCV2009Saragih03}.

\begin{figure}[!t]
\centering
\includegraphics[width=0.8\columnwidth]{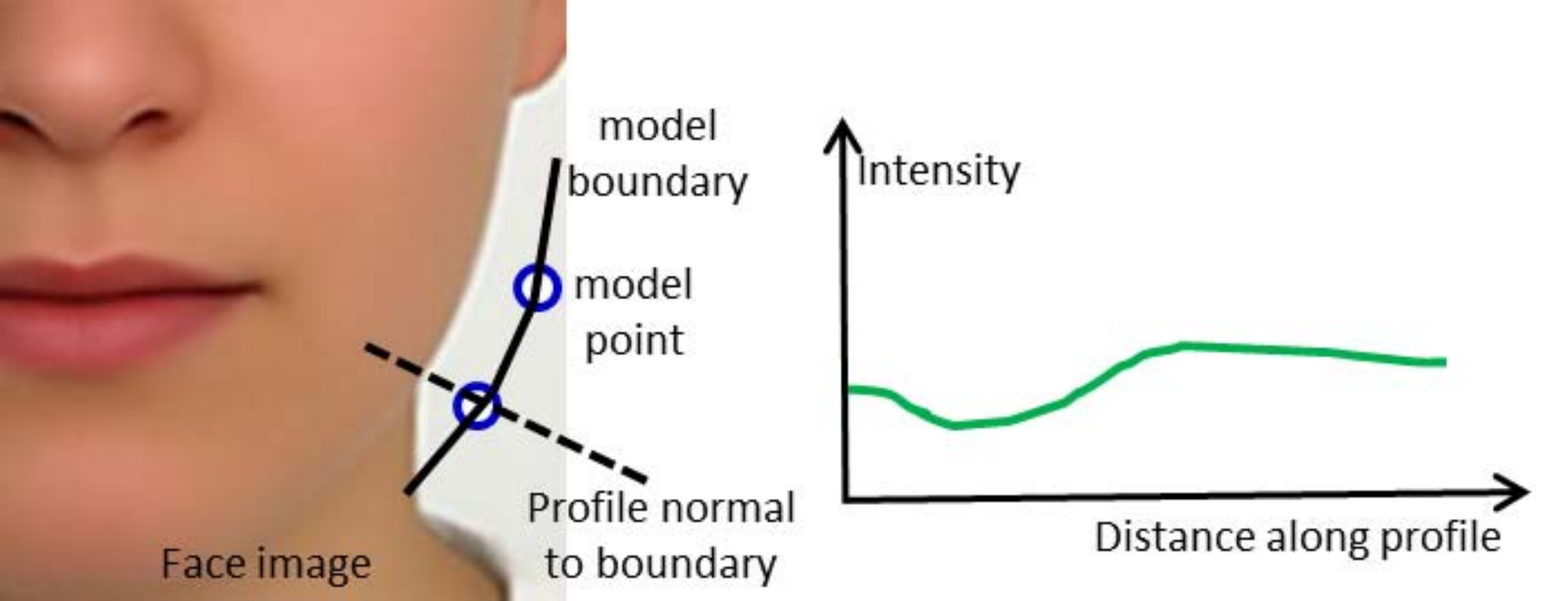}
\caption{ASM search profile.}
\label{fig:Fig 4}
\end{figure}

Regarding ASM, \citet{PBCVIU1995Cootes} defined the support region as the profile normal to the model boundary through each shape model point (see Fig. \ref{fig:Fig 4}). Along the profile, $k$ pixels are sampled from both sides of the model point in the $i$-th training image. Then $2k+1$ samples (actually gradients of these pixels) can be concatenated into a column vector $\textbf{g}_i$. After being normalized by the sum of the absolute value of elements in the vector, the mean $\bar{\textbf{g}}$ and the covariance $\textbf{S}_g$ can be estimated from all training vectors $\{\textbf{g}_i\}$. They adopted the multivariate Gaussian distribution assumption for the vectors. The fitting response for a new sample vector $\textbf{g}_s$ is given by

\begin{equation}
\label{eq 5}
(\textbf{g}_s-\bar{\textbf{g}})^T{\textbf{S}_g^{-1}}(\textbf{g}_s-\bar{\textbf{g}}),
\end{equation}
which is also known as the Mahalanobis distance of the sample vector from the model mean. The authors then provided a quantitative evaluation of the active shape model search using these local grey-level models \citep{PBBMVC1993Cootes}.

The aforementioned Mahalanobis distance-based methods assume that the local appearance is Gaussian distributed. This Gaussian distribution assumption does not always hold and thus may result in inferior performance. Classifier-based local experts separate aligned from misaligned locations, and so they ignore the local appearance variations. These experts are trained from positive image patches (centered at corresponding facial feature points) and negative image patches (with their centers displaced from the correct facial feature point positions). The linear support vector machine is frequently chosen due to its efficiency \citep{IJCV2011Saragih, CVPR2008Wang}. Taking the local expert corresponding to the $i$-th facial feature point as an example:
\begin{equation}
\label{eq 6}
\mathcal{C}_i(\textbf{x};\textbf{I})=\textbf{w}_i{\textbf{f}}+\gamma_i,
\end{equation}
where $\{\textbf{w}_i,\gamma_i\}$ denote the gain and the bias, respectively and $\textbf{f}$ represents the normalized patch vector with zero mean and unit variance. To reformulate the output of the classifier in a probability form, the logistic regression is employed to refine the equation (\ref{eq 6}):

\begin{equation}
\label{eq 7}
\frac{1}{1+e^{a_i{\mathcal{C}_i(\textbf{x};\textbf{I})}+b_i}}.
\end{equation}

\begin{figure}[!t]
\centering
\includegraphics[width=0.8\columnwidth]{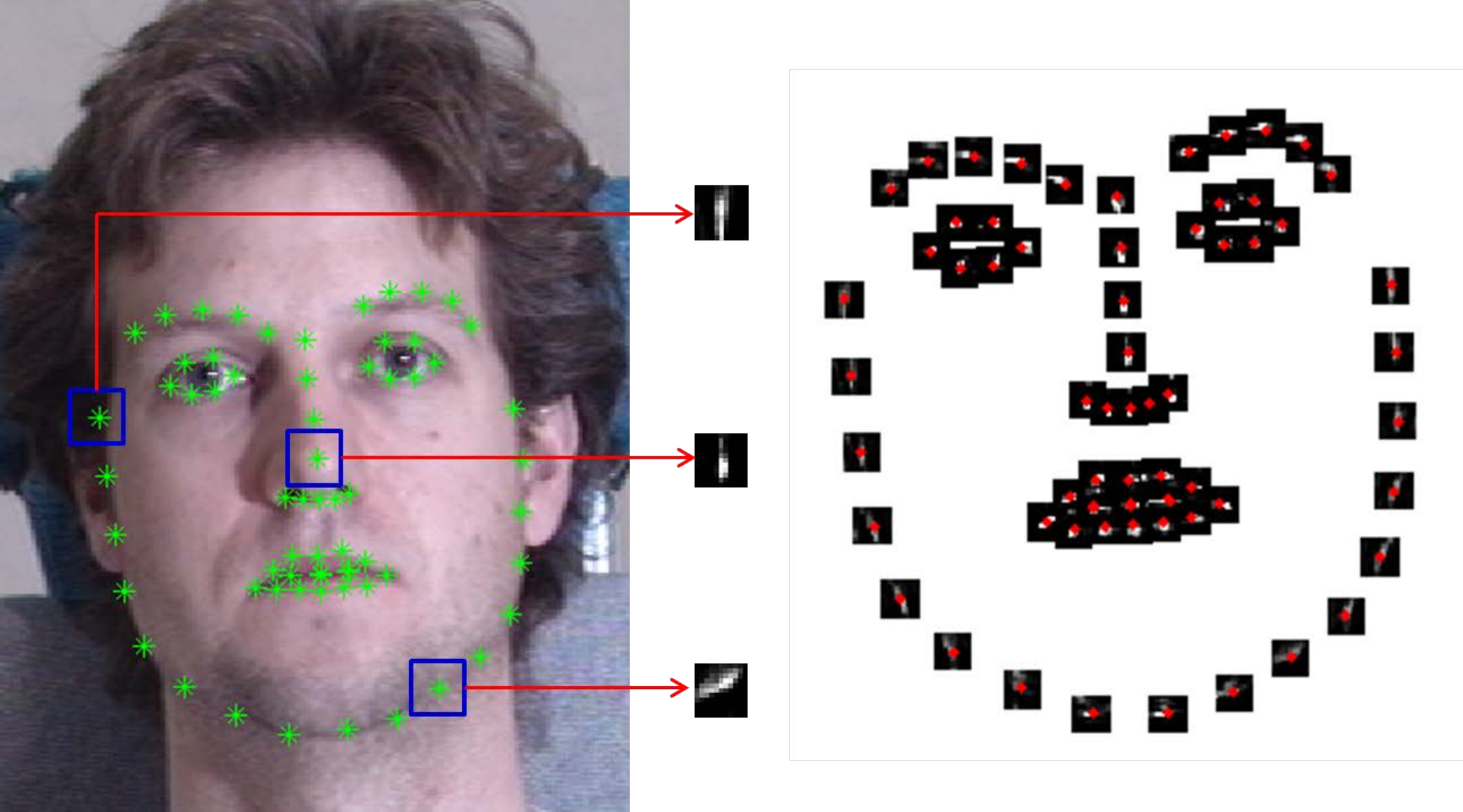}
\caption{Illustration of response maps. The area within the blue box is the region used to calculate the response maps. The red dots indicate the ground truth facial feature point locations.}
\label{fig:Fig 5}
\end{figure}

Given an estimated shape, we can calculate the response map within the region around each facial feature point according to equation (\ref{eq 7}). Fig. \ref{fig:Fig 5} shows the response maps of $66$ classifiers (specifically, linear support vector machines).

An alternative way to model the local expert is to exploit regressors instead of classifiers. \citet{BMVC2007Cristinacce} explored GentleBoost \citep{RPAS2000Friedman} to learn a regressor from the local neighborhood appearance to the displacement between the center of the local neighborhood and the true facial feature point location. \citet{ICCV2009Saragih03} claimed that a fixed mapping function (regressor) would take a complex form to incorporate the issues of generalizability and computational complexity. Considering a fixed mapping function cannot adapt to face variations in identity, pose, illumination and expression, they developed a bilinear model. \citet{ECCV2012Cootes} introduced random forest \citep{ RPML2001Breiman} to the CLM framework. Random forest learns response maps taking Haar-like features as the regressor input. PDM statistically models the shape models and regularizes the global shape configuration. The motivation behind the regressor rather than the classifier is that the regressor can potentially provide more useful information, such as the distance of negative patches from a positive patch, while classifiers only determine whether an image patch is positive or negative. However, learning a regressor is more difficult than constructing a classifier.

\subsection{Improvements and Extensions}
The fitting of CLM-based methods consists of two main steps: (1) predicting local displacements of shape model points; (2) constraining the configuration of all point to adhere to the shape model. These two steps are iterated until they satisfy a convergence criterion.

Cootes and Taylor \citep{PBBMVC1992Cootes, PBCVIU1995Cootes} proposed to search the "better" candidate point locations along profiles normal to the boundary. Corresponding displacements from current point locations to sought "better" locations should then be refined to adapt the PDM. The fitting objective function can be written in a similar form as equation (\ref{eq 1}) \citep{IJCV2011Saragih}:

\begin{equation}
\label{eq 8}
\mathrm{min}_{\textbf{p}}\left\|\pmb{\alpha}\right\|_{\pmb{\Lambda}^{-1}}^2+\sum_{i=1}^N\omega_i\left\|\textbf{x}_{(i)}-\pmb{\mu}_{(i)}\right\|^2,
\end{equation}
where $\pmb{\Lambda}=diag\{[\lambda_1;\cdots;\lambda_n]\}$, $\pmb{\mu}_{(i)}$ is the sought location of the $i$-th facial feature point corresponding to the peak response (the maximum value of the equation (\ref{eq 4})), the weights $\{\omega_i\}_{i=1}^N$ measure the significance of the peak response coordinates. In the above optimization problem, the first term neglects regularization on rigid transformation parameters $\textbf{q}$ by assuming a non-informative prior. To minimize the problem (\ref{eq 8}), the first order Taylor expansion of the PDM's points is applied:

\begin{equation}
\label{eq 9}
\textbf{x}_{(i)}\approx \textbf{x}_{(i)}^c+\textbf{J}_i\triangle{\textbf{p}},
\end{equation}
where $\textbf{x}^c=[\textbf{x}_{(1)}^c;\cdots;\textbf{x}_{(N)}^c]$ denotes the current approximated PDM shape, and $\textbf{J}=[\textbf{J}_1;\cdots;\textbf{J}_N]$ is the PDM's Jacobian matrix. Substituting the equation (\ref{eq 9}) into (\ref{eq 8}) we can then obtain the increment for updating the parameters:

\begin{equation}
\label{eq 10}
\triangle\mathbf{p}=-\textbf{H}^{-1}\left(\tilde{\pmb{\Lambda}}^{-1}\textbf{p}+\sum_{i=1}^N{\omega_i}\textbf{J}_i^T{(\textbf{x}_i^c-\pmb{\mu}_i)}\right),
\end{equation}
where $\textbf{H}=\tilde{\pmb{\Lambda}}^{-1}+\sum_{i=1}^N{\omega_i}\textbf{J}_i^T\textbf{J}_i$ is the Gauss-Newton Hessian and $\tilde{\pmb{\Lambda}}=diag\{[\textbf{0};\lambda_1;\cdots;\lambda_n]\}$. The parameters can be updated in an additive manner: $\textbf{p}\leftarrow{\textbf{p}+\triangle{\textbf{p}}}$. Indeed, from a probabilistic perspective \citep{IJCV2011Saragih}, the ASM's fitting procedure is equivalent to modeling the response maps by the isotropic Gaussian estimators $\{\mathcal{N}(\textbf{x}_{(i)};\pmb{\mu}_i,\omega_i^{-1}\textbf{I}^{(e)})\}_{i=1}^N$.

Since the emergence of the seminal work ASM \citep{PBCVIU1995Cootes}, quantity variants have been proposed. \citet{PBBMVC1994Cootes} proposed a multi-resolution strategy to improve the fitting performance from coarse to fine. The optimized solution on a low resolution image is taken as the initialization of the next higher resolution image. This strategy overcomes the sensitivity to initialization to some extent. \citet{FG2011Roh} found that the least squares in the equation (\ref{eq 7}) determines whether the problem will have optimal results only when the assumption of Gaussian noise is satisfied. However, since non-Gaussian noise is regularly encountered, they proposed to employ two strategies (M-estimator and random sampling) to robustly estimate these parameters. \citet{PBCVPR2003Zhou} formulated FFPD into a maximum a posterior (MAP) problem in the tangent space and designed an expectation-maximization (EM)-based fitting algorithm to solve the MAP optimization. The tangent shape is iteratively updated by a weighted average of model shapes and tangent projection of the observed shape while the shape is reconstructed from model shapes in ASM. Furthermore, continuous regularization of shape parameters was applied while traditional ASM discontinuously truncated shape parameters to constrain the shape variation, which could result in unstable estimation. \citet{ICCV2007Vogler} proposed to combine a 3D deformable model based on ASM for reliable real-time tracking. The 3D deformable model mainly governs the overall variation of a face (such as shape, orientation, and location). Several ASMs are trained, each corresponding to a viewpoint to govern the 2D facial feature variations. \citet{DB2010MUCT} extended the 1D profile to the 2D profile (actually a squared area) which outperformed the traditional ASM. \citet{PAMI2008Wimmer} investigated how to learn local objective functions for face model fitting. They claimed that a best local objective function should have the following two properties: (1) a global minimum corresponding to the best model fit; (2) no local extrema or saddle points. They then learned objective functions under the framework of ASM. Rather than using the Mahalanobis distance, they explored tree-based regression to learn an objective function mapping from the extracted Haar-like and edge features.

To further facilitate the localization of facial feature points, some component-based methods have also been proposed. \citet{ECCV2008Liang} utilized the component locations as constraints to regularize the configuration of facial features. This method first detects $11$ components by cascaded boosting classifier \citep{RPIJCV2004Viola}. By solving a fitting objective similar to that of ASM except for an additional constraint term of component locations, a fitted shape can be resolved. To further improve the detection accuracy of components, the authors proposed to utilize direction classifiers to determine the search direction for component locations. These direction classifiers (3 classifiers for left/right profile, brows, and upper/lower lips and 9 classifiers for other components) are trained from positive and negative samples with respect to corresponding components. To determine the appropriate position along the above detected direction, a customized searching strategy is designed. Since new positions of components are found, an updated shape can be achieved by solving the aforementioned fitting objective function. Through several such iterations, a reasonable shape can ultimately be reached. \citet{ECCV2012Le} presented a component-based ASM model and an interactive refinement algorithm. According to the aforementioned descriptions, ASM consists of two models: a profile model (local expert) and a shape model. Unlike the ASM method, which models all points by a single multivariate Gaussian distribution, this approach separates the whole face into seven components and constructs a Gaussian model for each component. To obtain a reasonable configuration of these components, the locations of these components (centroids of these components) are further modeled by Gaussian distribution. In other words, the shape model is decomposed into two modules: component shape fitting and configuration model fitting. For the profile model, besides the unary scores (owing to the fact that $N$ local detectors are independent), binary constraint (tree-structure) is introduced to refine each pair of neighboring landmarks. Similar binary constraints have been imposed by tree-structure \citep{ECCV2006Zheng} and MRF (graph structure with loops) in \citet{BMVC2009Tresadern}. Lastly, dynamic programming is explored to solve the fitting problem. Moreover, the authors introduced a user-assisted facial feature point localization strategy to further decrease localization error.

\citet{PBTMI2002Ginneken} substituted the fixed normalized first derivative profile \citep{PBCVIU1995Cootes} with a distinct set of optimal features for each facial feature point. A nonlinear $k$-nearest neighbor (kNN) classifier instead of linear Mahalanobis distance has also been explored to search the optimal displacement for points. \citet{PAMI2007sukno} proposed a generalization of optimal features ASM \citep{PBTMI2002Ginneken}. A reduced set of differential invariant features is taken as the local appearance descriptors, which are invariant to rigid transformation. In the fitting phase, a sequential features selection method is adopted to choose a subset of features for each point. To further speed the procedure, multivalued neurons (MVN) \citep{RPBook2000Aizenberg} are adopted to replace the kNN classifier in \citet{PBTMI2002Ginneken}.

Inspired by the cascaded face detection method \citep{RPIJCV2004Viola}, \citet{PBBMVC2003Cristinacce} proposed to detect each local facial feature point by trained an Adaboost classifier. To constrain the global configurations of these points and reliably locate each point, multivariate Gaussian was assumed for the shape point distribution. They then further \citep{PBFG2004Cristinacce} extended this model by utilizing three templates to compute the response map for each individual feature point: normalized correlation template, orientation maps and boosted classifier. The fitting objective is to maximize the sum of all these response maps under the constraint that each PDM shape parameter should be under the threshold of three standard deviations. The Nelder-Mead simplex method \citep{RPComputerJournal1965} was explored to optimize this problem. They then \citep{FG2006Cristinacce} proposed an adaptive strategy to update the template to compute the response maps. They \citep{BMVC2006Cristinacce, PR2008Cristinacce} first proposed the term "constrained local model" consisting of two steps: the first step is to calculate the response maps for each facial feature point; the second step is to maximize the sum of response scores under the Gaussian prior constraint, as in equation (\ref{eq 7}). Given an input testing image, templates are generated through an appearance model (AAM \citep{PBPAMI2001Cootes}) constructed from vectors, each of which is the concatenation of image patches extracted from each facial feature point in a training image. These templates are iteratively updated in the fitting process. The Nelder-Mead simplex method \citep{RPComputerJournal1965} is utilized to optimize the problem.

\citet{IVC2009Lucey} proposed an improved version of the method \citep{PR2008Cristinacce}, named exhaustive local search, in the following aspects: (1) substitute the original generative patch experts with a discriminative expertt trained by linear support vector machine; (2) decompose the complex fitting function into $N$ independent fitting problems, which greatly favors real-time performance; (3) exploit a composite rather than additive warp update step. However, this method only utilizes the maximum of a response map for each facial feature point, neglecting the distribution of response maps. Furthermore, constraints to the shape configuration are not taken into account, which may lead to an invalid shape.

Although isotropic Gaussian estimation to the response maps leads to an efficient and simple approximation, it may fail in some cases if the response maps cannot be modeled by isotropic Gaussian distributions.

\citet{CVPR2008Wang} proposed to approximate the response map by: $\{\mathcal{N}(\textbf{x}_{(i)};\pmb{\mu}_i,\pmb{\Sigma}_i)\}_{i=1}^N$, anisotropic Gaussian estimators. Here $\pmb{\Sigma}_i$ is the full covariance matrix. $\pmb{\mu}_i$ and $\pmb{\Sigma}_i$ can be inferred from a convex quadratic function fitted to the negative log of the response maps (obtained from a linear support vector machine). The fitting problem can be written as \citep{IJCV2011Saragih}:

\begin{equation}
\label{eq 11}
\mathrm{min}_{\textbf{p}}\left\|\pmb{\alpha}\right\|_{\pmb{\Lambda}^{-1}}^2+\sum_{i=1}^N\left\|\textbf{x}_{(i)}-\pmb{\mu}_{(i)}\right\|_{\pmb{\Sigma}_i^{-1}}^2,
\end{equation}
Then the Gauss-Newton update is:

\begin{equation}
\label{eq 12}
\triangle\textbf{p}=-\textbf{H}_{ani}^{-1}\left(\tilde{\pmb{\Lambda}}^{-1}\textbf{p}+\sum_{i=1}^N\textbf{J}_i^T\pmb{\Sigma}_i^{-1}(\textbf{x}_i^c-\pmb{\mu}_i)\right),
\end{equation}
where $\textbf{H}_{ani}=\tilde{\pmb{\Lambda}}^{-1}+\sum_{i=1}^N\textbf{J}_i^T\pmb{\Sigma}_i^{-1}\textbf{J}_i$. They subsequently applied this strategy to non-rigid face tracking \citep{FG2008Wang}. Paquet proposed a Bayesian version of the method \citep{CVPR2008Wang} which can be seen as the maximum likelihood solution of the proposed Bayesian method.

Considering that response maps may be multimodal, a single Gaussian estimator cannot model the density distribution. \citet{ECCV2008Gu} employed a Gaussian mixture model (GMM) to approximate the response maps: $\{\sum_{k=1}^{K_i}\pi_{ik}\mathcal{N}(\textbf{x}_{(i)};\pmb{\mu}_{ik},\pmb{\Sigma}_{ik})\}_{i=1}^N$ where $K_i$ is the number of Gaussian components to model the response map corresponding to the $i$-th point and $\pi_{ik}$ are the mixing coefficients. GMM parameters are estimated from the GMM fitting process to the response maps. Finally the optimization problem is \citep{IJCV2011Saragih}:

\begin{equation}
\label{eq 13}
\mathrm{min}_{\textbf{p}}\left\|\pmb{\alpha}\right\|_{\pmb{\Lambda}^{-1}}^2+\sum_{i=1}^N\sum_{k=1}^{K_i}\omega_{ik}\left\|\textbf{x}_{(i)}-\pmb{\mu}_{(ik)}\right\|_{\pmb{\Sigma}_{ik}^{-1}}^2,
\end{equation}
where $\omega_{ik}=\frac{\pi_{ik}\mathcal{N}(\textbf{x}_{(i)};\pmb{\mu}_{ik},\pmb{\Sigma}_{ik})}{\sum_{j=1}^{K_i}\pi_{jk}\mathcal{N}(\textbf{x}_{(i)};\pmb{\mu}_{ij},\pmb{\Sigma}_{ij})}$. Through Gaussian-Newton optimization, the update can be calculated:

\begin{equation}
\label{eq 14}
\triangle\textbf{p}=-\textbf{H}_{GMM}^{-1}\left(\tilde{\pmb{\Lambda}}^{-1}\textbf{p}+\sum_{i=1}^N\sum_{k=1}^{K_i}\omega_{ik}\textbf{J}_i^T\pmb{\Sigma}_{ik}^{-1}(\pmb{\mu}_{ik}-\textbf{x}_i^c)\right),
\end{equation}
where $\textbf{H}_{GMM}=\tilde{\pmb{\Lambda}}^{-1}+\sum_{i=1}^N\sum_{k=1}^{K_i}\omega_{ik}\textbf{J}_i^T\pmb{\Sigma}_{ik}^{-1}\textbf{J}_i$. \citet{ICCV2009Saragih02} utilized GMM to approximate response maps of the proposed mixture of local experts. They have a similar fitting objective as in equation (\ref{eq 13}) except without the shape prior regularization.

Unlike previous methods approximating response maps in parametric forms, Saragih et al. \citep{ICCV2009Saragih01, IJCV2011Saragih} proposed a non-parametric estimate in the form of a homoscedastic isotropic Gaussian kernel density estimate: $\{\sum_{\textbf{y}_{(j)}\in\pmb{\Psi}_i}\pi_{\textbf{y}_{(j)}}\mathcal{N}(\textbf{x}_{(i)};\textbf{y}_{(j)},\rho\textbf{I}^{(e)})\}_{i=1}^N$. Here $\pmb{\Psi}_i$ represents all integer pixel locations within the rectangular region around $i$-th facial feature point, $\pi_{\textbf{y}_{(i)}}$ denotes the likelihood that the $i$-th point is aligned at location $\textbf{y}_{(i)}$ which can be estimated from the equation (\ref{eq 7}), and $\rho$ denotes the variance of the noise on point locations which can be determined from training data as $\rho=\frac{1}{N-n}\sum_{i=n+1}^{N}\lambda_i$. The fitting objective function is as follows:

\begin{equation}
\label{eq 15}
\mathrm{min}_{\textbf{p}}\left\|\pmb{\alpha}\right\|_{\pmb{\Lambda}^{-1}}^2+\sum_{i=1}^N\sum_{\textbf{y}_{(j)}\in\pmb{\Psi}_i}\frac{\omega_{\textbf{y}_{(j)}}}{\rho}\left\|\textbf{x}_{(i)}-\textbf{y}_{(j)}\right\|^2,
\end{equation}
where $\omega_{\textbf{y}_{(j)}}=\frac{\pi_{\textbf{y}_{(i)}}\mathcal{N}(\textbf{x}_{(i)};\textbf{y}_{(i)},\rho\textbf{I}^{(e)})}{\sum_{\textbf{z}_{(j)}}\pi_{\textbf{z}_{(j)}}\mathcal{N}(\textbf{x}_{(i)};\textbf{z}_{(j)},\rho\textbf{I}^{(e)})}$.
The update is:

\begin{equation}
\label{eq 16}
\triangle\textbf{p}=-(\rho\tilde{\pmb{\Lambda}}^{-1}+\textbf{J}^T\textbf{J})^{-1}(\rho\tilde{\pmb{\Lambda}}^{-1}\textbf{p}-\textbf{J}^T\textbf{v}),
\end{equation}
where $\textbf{v}=[\textbf{v}_1;\cdots;\textbf{v}_N]$ and \\ $\textbf{v}_i=\left(\sum_{\textbf{y}_{(j)}\in\pmb{\Psi}_i}\frac{\pi_{\textbf{y}_{(j)}}\mathcal{N}(\textbf{x}_{(i)};\textbf{y}_{(j)},\rho\textbf{I}^{(e)})}{\sum_{\textbf{z}_{(j)}}\pi_{\textbf{z}_{(j)}}\mathcal{N}(\textbf{x}_{(i)};\textbf{z}_{(j)},\rho\textbf{I}^{(e)})}\right)-\textbf{x}_i^c$
\\which is in a similar form with mean-shift. To further handle partial occlusions, they used an M-estimator to substitute the least square in the equation (\ref{eq 15}).

The above method \citep{IJCV2011Saragih} has been extensively investigated due to its effectiveness and efficiency. \citet{FG2011Chew} have applied this method to facial expression detection. Excluding the influence of occluded points through random sample consensus \citep{RPCACM1981Fischler} hypothesis-and-test strategy, \citet{FG2011Roh} proposed an algorithm robust to occlusion. Response maps achieved from linear SVM are represented in a multi-model fashion resulting from the mean-shift segmentation (each segmented region is modeled by a 2D-Gaussian distribution). \citet{CVPR2012Baltrusaitis} extended the above method \citep{IJCV2011Saragih} to a 3D version. In addition to general face images, they explored the information of depth images. The mean of response maps estimated from the general image and corresponding depth image is taken as the final response map. \citet{ICCV2013Yu} explored the mean-shift method \citep{IJCV2011Saragih} to rapidly approach the global optimum in their proposed two-stage cascaded deformable shape model and then utilized comp\hyp onent-wise active contours to discriminatively refine the subtle shape variation.

Unlike the aforementioned non-parametric and parametric approximations to response maps, \citet{CVPR2013Asthana} directly regressed the PDM shape update parameters from the low-dimensional representation of response maps through a series of weak learners. The response maps can be obtained from linear support vector machines and the low-dimensional representation is obtained from the PCA projection. Linear support vector regression plays the role of the weak learner.

\citet{ECCV2012Martins} claimed that the above subspace-constrained mean-shift method \citep{ICCV2009Saragih01, IJCV2011Saragih} is vulnerable to outliers owing to a least squares projection. They formulated the objective as maximum a posterior (MAP) of PDM shape parameters $p(\pmb{\alpha}|\textbf{s})$ and pose parameters $p(\textbf{q}|\textbf{s})$ respectively conditioned on the observed shape. The observed shape here is obtained according to response maps. The MAP $p(\pmb{\alpha}|\textbf{s})$ can be decomposed into $p(\textbf{s}|\pmb{\alpha})p(\pmb{\alpha})$. According to PDM (see equation (\ref{eq 3})), $p(\textbf{s}|\pmb{\alpha})$ can be modeled as a Gaussian distribution:

\begin{equation}
\label{eq 17}
\begin{split}
&p(\textbf{s}|\pmb{\alpha})\propto\\
&{exp\Big(-\frac{1}{2}\big(\textbf{s}-(\textbf{s}_0+\sum_{i=1}^n\alpha_i\textbf{s}_i)\big)^T\pmb{\Sigma}_{\textbf{s}}^{-1}\big(\textbf{s}-(\textbf{s}_0+\sum_{i=1}^n\alpha_i\textbf{s}_i)\big)\Big)},
\end{split}
\end{equation}
where $\pmb{\Sigma}_{\textbf{s}}$ indicates the uncertainty of the spatial localization of all points and can be estimated from response maps. To simplify the optimization procedure, they adopted the conjugate prior \citep{BMVC2012Martins}, i.e. $p(\pmb{\alpha})$ distributes as a Gaussian. The MAP problem $p(\textbf{q}|\textbf{s})$ was processed in the same way. Finally the MAP problem was optimized by linear dynamical systems.

\citet{CVPR2011Belhumeur} proposed a method that combines the output of local experts with a non-parametric global model. The local expert is applied through a support vector machine taking the SIFT feature \citep{RPIJCV2004Lowe} as the input. Based on the response maps of these local experts, the objective is to maximize the posterior probability $p(\textbf{x}|\textbf{d})$ where $\textbf{d}$ represents the response maps of all local experts. Since the location corresponding to the highest response map value is not always the correct location due to occlusions and appearance ambiguities, they further designed a non-parametric set of global models from the similarity transformation of training exemplar images to constrain the configurations of these facial feature points. The random sample consensus method \citep{RPCACM1981Fischler} is explored to optimize the global model.

\citet{ICCV2011Amberg} casted the FFPD detection problem as a discrete programming problem given a number of candidate positions for each point. They utilized decision forest \citep{RPBook1984Breiman} to detect a number of candidate locations for each point. The facial feature localization problem is actually to determine the indexes of points in corresponding candidate points which minimize the distance between the shape model and the image points. A fixed 3D shape projected according to a weak perspective camera is taken as the shape model. The objective is globally minimized by the branch and bound method.

\section{Active Appearance Model-Based Methods}
\label{sec:3}
\subsection{Active Appearance Model}
An active appearance model (AAM) \citep{SMCC2010Gao} can be decoupled into a linear shape model and a linear texture model. The linear shape is obtained in the same way as in the CLM framework (see equation (\ref{eq 3})). To construct the texture model, all training faces should be warped to the mean-shape frame by triangulation or thin plate spline method; the resultant images should be free of shape variation, called shape-free textures. Each shape-free texture is raster scanned into a grey-level vector $\textbf{z}_i$. To eliminate the effect of global lighting variation, $\textbf{z}_i$ is normalized by a scaling $u$ and offset $v$:

\begin{equation}
\label{eq 18}
\textbf{a}=\frac{\textbf{z}_i-v\cdot\textbf{1}}{u},
\end{equation}
where $u$ and $v$ represent the variance and the mean of the texture $\textbf{z}_i$ respectively and $\textbf{1}$ is a vector of all 1s with the same length as $\textbf{z}_i$. The texture model can be generated by applying PCA on all normalized textures as follows:
\begin{equation}
\label{eq 19}
\textbf{a}=\textbf{a}_0+\textbf{P}_a\pmb{\beta}=\textbf{a}_0+\sum_{i=1}^m\beta_i\textbf{a}_i,
\end{equation}

The coupled relationship between the shape model and the texture model is bridged by PCA on shape and texture parameters:

\begin{equation}
\label{eq 20}
\left(
\begin{array}{c}
\textbf{W}_s\pmb{\alpha} \\
\pmb{\beta} \\
\end{array}
\right)=
\left(
\begin{array}{c}
\textbf{W}_s\textbf{P}_{{s}}^T(\textbf{s}-\textbf{s}_0) \\
\textbf{P}_a^T(\textbf{a}-\textbf{a}_0) \\
\end{array}
\right)=
\left(
\begin{array}{c}
\textbf{Q}_s \\
\textbf{Q}_a \\
\end{array}
\right)
\textbf{c}
\end{equation}
where $\textbf{W}_s$ is a diagonal weighting matrix measuring the difference between the shape and texture parameters. The appearance parameter vector $\textbf{c}$ governs both the shape and texture variation. To simplify the parameter representation, here we still utilize $\textbf{p}$ to incorporate all necessary parameters: appearance parameters $\textbf{c}$, pose parameters $\textbf{q}$ and texture transformation parameters $u$ and $v$.

The fitting objective of AAM is to minimize the difference between the texture $\textbf{a}_s$ sampled from the testing image and the texture $\textbf{a}_m$ synthesized by the model. Let $\textbf{r}(\textbf{p})=\textbf{a}_s-\textbf{a}_m$. \citet{PBECCV1998Cootes} first proposed to model the relationship between $\textbf{r}(\textbf{p})$ (it was warped to the image frame in \citet{PBECCV1998Cootes}) and parameter update $\delta\textbf{p}$ by linear regression:

\begin{equation}
\label{eq 21}
\pmb{\delta}\textbf{p}=\textbf{A}\cdot\textbf{r}(\textbf{p}),
\end{equation}
where $\textbf{A}$ was solved by multiple multivariate linear regression on a sample of known model displacements $\pmb{\delta}\textbf{p}$ and the corresponding difference texture $\textbf{r}(\textbf{p})$. \citet{PBPAMI2001Cootes} later developed a Gaussian-Newton optimization method. Applying a first order Taylor expansion to $\textbf{r}(\textbf{p})$:

\begin{equation}
\label{eq 22}
\textbf{r}(\textbf{p}+\pmb{\delta}\textbf{p})=\textbf{r}(\textbf{p})+\frac{\partial\textbf{r}}{\partial\textbf{p}}\pmb{\delta}{\textbf{p}},
\end{equation}
Through solving the optimization problem: $\mathrm{min}_{\pmb{\delta}\textbf{p}}\|\textbf{r}(\textbf{p}+\pmb{\delta}\textbf{p})\|^2$, we receive the optimal solution:

\begin{equation}
\label{eq 23}
\pmb{\delta}\textbf{p}=-\textbf{B}\textbf{r}(\textbf{p}), \textbf{B}=\left(\frac{\partial\textbf{r}^T}{\partial\textbf{p}}\frac{\partial\textbf{r}}{\partial\textbf{p}}\right)^{(-1)}\frac{\partial\textbf{r}^T}{\partial\textbf{p}}.
\end{equation}
Considering the fact that updating $\frac{\partial\textbf{r}^T}{\partial\textbf{p}}$ at every iteration is expensive, the authors fixed it in a constant matrix which can be estimated from training images by numeric differentiation.

\subsection{Improvements and Extensions}
Due to the flexible and simple framework of AAM, it has been extensively investigated and improved. However, many difficulties are encountered when AAM is applied to real applications. These difficulties are generally encountered from the following three aspects: the low efficiency for real-time applications, the less discrimination for classification, and the lack of robustness under inconstant circumstances. As our previous work \citep{SMCC2010Gao} does, we review the developments of AAM from these three aspects.

\subsubsection{Efficiency}
Due to the high-dimensional texture representation and the unconstrained optimization, the original AAM suffers from low efficiency in real-time systems. We investigate improvements from these two aspects respectively.

1) Texture representation

To reduce the redundancy information contained in the texture, \citet{PBBMVC1998Cootes} only subsampled a number of pixels. Pixels corresponding to a number of the largest elements in the regression matrix are assumed to be helpful and are preserved. This procedure decreases the dimension of texture representation. However, since the assumption is not always tenable, it cannot be guaranteed to obtain reasonable results.

Since learning the regression matrix in (\ref{eq 21}) or (\ref{eq 23}) is time- and memory-consuming, \citet{PBCVPR2001Hou} learned the regression from the low-dimensional representation (PCA projection) of texture difference to the position displacement. Moreover, considering that the mapping from the texture to the shape is many-to-one, they proposed to linearly model the relationship between the texture and the shape to cater for this.

\citet{IJCV2012Tresadern} explored Haar-like features to provide a computationally inexpensive linear projection for efficiency to facilitate facial feature point tracking on a mobile device. To provide high accuracy, a hierarchical model that utilizes tailored training data is designed.

2) Optimization

In order to improve the efficiency of the fitting process, \citet{PBIJCV2004Matthews} considered the AAM as an image alignment problem and optimized it by inverse compositional method \citep{PBLK20Part3} based on independent AAM. Here, independent AAM indicates that the linear shape model and linear texture model are not combined, as in the original literature. The method aims to minimize the following objective function:

\begin{equation}
\label{eq 24}
\sum_{\textbf{s}_{(k)}\in{\textbf{s}_0}}\left[\textbf{a}_0(\textbf{s}_{(k)})+\sum_{i=1}^m\beta_i\textbf{a}(\textbf{s}_{(k)})-\textbf{I}\Big(\pmb{\mathcal{Q}}\big(\pmb{\mathcal{W}}(\textbf{s}_{(k)};\textbf{p});\textbf{q}\big)\Big)\right]^2,
\end{equation}
where $\textbf{s}_{(k)}$ denotes any pixel location with the area enclosed by the mean shape $\textbf{s}_0$, $\pmb{\mathcal{W}}(\textbf{s}_{(k)};\textbf{p})$ represents the pixel location after warping $\textbf{s}_{(k)}$ with a warp $\pmb{\mathcal{W}}(\cdot;\textbf{p})$, $\pmb{\mathcal{Q}}(\textbf{s}_{(k)};\textbf{q})$ has a similar meaning and a composition relation exists: $\pmb{\mathcal{Q}}\circ\pmb{\mathcal{W}}(\textbf{s}_{(k)};\textbf{p},\textbf{q})$. The advantage of the inverse compositional method is that in the fitting process, many variants such as the Jacobian matrix and the Hessian matrix can be precomputed.

The inverse compositional method has had many variants since its birth. It has been applied to solve the robust and efficient FFPD objective \citep{ICCV2011Tzimiropoulos} which aims to detect points under occlusion and illumination changes. \citet{PBIVC2005Gross} proposed a simultaneous inverse compositional algorithm which simultaneously updates the warp parameters $\textbf{p}$ and the texture parameters $\pmb{\beta}$. Moreover, they also claimed that: (1) the person specific AAM is much easier to build and fit than the generic AAM (person-independent AAM) and can also achieve better performance; (2) the generic shape model is far easier to build than the generic texture model; (3) the origin of the idea that fitting the generic AAM is far harder than the person specific AAM lies in fact that the effective dimensionality of the generic shape model is far higher than that of the person-specific shape models. \citet{CVPR2008Papandreou} presented two improvements to the inverse compositional AAM fitting method to overcome significant appearance variation: fitting algorithm adaptation through fitting matrix adjustment and AAM mean template update by incorporating the prior information to constrain the fitting process. \citet{FG2008Saragih} applied a mixed inverse-compositional-forward-additive parameter update scheme to optimize the objective subject to soft correspondence constraints between the image and the model. \citet{CVPR2009Brian} claimed that the inverse compositional method has a small convergence radius and proposed two improvements to enlarge the radius at the expense of it being four times slower and preserving the same time consumption respectively. Lucey et al. \citep{CVPR2010Ashraf, PAMI2013Lucey} extended the inverse compositional method in the Fourier domain for image alignment and applied this method specifically to the case of AAM fitting \citep{ICCV2011Navarathna}. \citet{ACCV2012Tzimiropoulos} proposed a generative model called active orientation model which is as computationally efficient as the standard project-out inverse compositional algorithm. Subsequently, \citet{ICCV2013Tzimiropoulos} proposed a framework for efficiently solving AAM fitting problem in both forward and inverse coordinate frames. Benefiting from the efficiency of proposed framework, they trained and fitted AAM in-the-wild and the trained model could achieve promising performance.

\citet{PAMI2006Donner} claimed that the multivariate regression technology explored in conventional AAM neglects the correlations between the response variables. This results in slow convergence (more iterations) in the fitting procedure. Since canonical correlation analysis models the correlations between response variables, it is employed to calculate a more accurate gradient matrix.

\citet{BMVC2010Tresadern} utilized an additive update model (boosting model, both linear and nonlinear) to substitute for the original linear predictors in AAM taking Haar-like features as the regression input. They found that the linear additive model is faster than original linear regression \citep{PBECCV1998Cootes} but it preserves comparable accuracy and is also as effective as nonlinear models when close to the true solution. Therefore, they suggested a hybrid AAM which utilizes a nonlinear additive update model at the first several iterations and then a linear additive update model in the last several iterations.

Although the linear regression strategy achieves some success in obtaining the updated parameters, it is a coarse approximation of the nonlinear relation between texture residuals and warp parameters. When the parameters are initialized far away from the right place, this linear assumption is invalid. To this end, \citet{ICCV2007Saragih} deployed a nonlinear boosting procedure to learn the multivariate regression. Each parameter is updated by a strong regressor consisting of an ensemble of weak learners \citep{AS2001Friedman}. Each weak learner is fed with Haar-like features to output the parameter. This nonlinear modeling results in a more accurate fitting than linear procedures.

\citet{CVPR2007Liu, PAMI2009Liu} explored GentleBoost classifier \citep{RPAS2000Friedman} to model the nonlinear relationship between texture and parameter updates. A strong classifier consists of an ensemble of weak classifiers (arc-tangent functions). Haar-like rectangular features are fed into each weak classifier. The goal of the fitting procedure is to find the PDM parameter updates which maximize the score of the strong classifier. \citet{ACCV2009Zhang} utilized granular features to replace the rectangular Haar-like feature to improve computational efficiency, discriminability and a larger search space. In addition, they explored the evolutionary search process to overcome the deficiency searching problem in the large feature space. Because the weak classifier in \citet{CVPR2007Liu, PAMI2009Liu} is actually utilized to classify the right PDM parameters from the wrong ones, it cannot guarantee that the fitting objective will converge to the optimum solution. Consequently, instead of discriminatively classifying corrected alignment from incorrect alignment, \citet{CVPR2008Wu} learned classifiers (GentleBoost) to determine whether to switch from one shape parameters to another parameter corresponding to an improved alignment. Based on the ranking appearance model \citep{CVPR2008Wu}, \citet{BMVC2012Gao} preferred to use gradient boosted regression trees \citep{AS2001Friedman} instead of GentleBoost classifiers. Modified census transform features and pseudo census transform features \citep{BMVC2011Gao} are fed to the regression trees.

\citet{BMVC2011Sauer} compared the performance of linear predictor, boosting-based predictor and random regression-based predictor. Their experimental results illustrate the random regression-based method achieves the best generalization ability. Furthermore, it can achieve performance that is as efficient as boosting procedures without significant reduction in accuracy.

\subsubsection{Discrimination}
Regarding discrimination, here we mainly refer to the ability to accurately fit a model to an image \citep{SMCC2010Gao}. Many aspects may affect this, such as prior knowledge, texture representation and nonlinear modeling the relation between texture residuals and parameters.

Instead of simply minimizing a sum of square measure, \citet{PBICCV2001Cootes} reformulated the AAM problem in a MAP form $p(\textbf{p}|\textbf{I})=p(\textbf{I}|\textbf{p})p(\textbf{p})$ is a zero-mean Gaussian with covraiance matrix $\textbf{S}_p^{-1}$, the MAP problem can be simplified in a log-probability form to minimize the following problem:

\begin{equation}
\label{eq 25}
E(\textbf{p})=\sigma_r^2\textbf{r}^T\textbf{r}+\textbf{p}^T\textbf{S}_p^{-1}\textbf{p}.
\end{equation}
Following a similar procedure to that described in Section3.1, the above problem can be resolved. To deploy the prior knowledge such as the position of certain points, a further prior can be added to equation (\ref{eq 25}) and the optimization is still a similar procedure.

Traditional AAM \citep{PBPAMI2001Cootes} fixed the gradient matrix (Jacobian of the residual to the parameters) which may lead to poor performance when the texture of a testing image differs dramatically from the mean texture. \citet{PBCVPR2003Batur, PBTIP2005Batur} update the gradient matrix in each iteration by adding a linear combination of basis matrixes to a fixed basic matrix. \citet{BMVC2006Cootes} presented a strategy-like quasi-Newton method to update the gradient matrix. The update of the gradient matrix will increase the fitting accuracy to some extent but will also reduce efficiency. \citet{PR2009Saragih} pointed out that a fixed linear update model, as in traditional AAM, has limited ability to account for the various error terrains about the optimum in different images and an adaptive update model according to the image at hand requires a time consuming process of re-calculating it for every iteration. They adopted a compromise manner: learning a set of fixed linear update models to be applied to the image sequentially in the fitting process.

\citet{ECCV2006Zheng} proposed a rank-based non-rigid shape detection method through RankBoost \citep{RPJMLR2003Freund}. RankBoost is utilized to learn a ranking model from Haar-like features extracted from warped training images. The ranking model is then applied to Haar-like features extracted from images warped from the testing image to calculate the response scores. The final shape is achieved from a linear combination of the $K$ training shapes corresponding to previously calculated top $K$ response scores. One disadvantage of this method is that the detection efficiency is seriously affected by the number of images in the training set.

Standard AAM achieves limited accuracy in fitting a face image for an individual unseen in the training set. This is mainly because the appearance model of AAM (created in a generative manner) has limited generalization ability. \citet{BMVC2007Peyras} proposed a multi-level segmented method which constructs multiple AAMs, each corresponding to different parts of a face, e.g. eye, mouth, and nose. The whole fitting strategy is in a coarse-to-fine fashion (multi-resolution) and a different number of AAMs are correspondingly constructed.

Nguyen et al. \citep{CVPR2008Nguyen, FG2008Nguyen, IJCV2010Nguyen} claimed that AAMs are easily converged to local minima in the fitting process and that the local minima seldom correspond to acceptable solutions. They proposed a parameterized appearance model which learns a cost function having local minima at and only at desired places. This is guaranteed by a quadratic error function with a symmetric positive semidefinite coefficient matrix corresponding to the quadratic term. The objective function is optimized by a variant of the Newton iteration method.

\subsubsection{Robustness}

The robustness of AAM is generally influenced by the inconstant circumstances, e.g. pose variations, resolutions, illumination changes, occlusion, and any other wild conditions.

\textbf{Pose}: \citet{PBIVC2002Cootes} demonstrated that five AAMs corresponding to five views ($-90^\circ$, $-45^\circ$, $0^\circ$, $45^\circ$, $90^\circ$) can capture the appearance variation across a wide range of rotations. Each AAM can capture faces rotated in a range of view angles. The rotation angle $\theta$ is dependent on the appearance parameters $\textbf{c}$ through a model:

\begin{equation}
\label{eq 26}
\textbf{c}=\textbf{c}_0+\cos{\theta}\textbf{c}_x+\sin{\theta}\textbf{c}_y.
\end{equation}
\citet{CVIU2012Huang} combined view-based AAM \citep{PBIVC2002Cootes} with Kalman filter to perform pose robust face tracking. Instead of model parameters controlling the shape and appearance variations, this paper only utilized the shape parameters to construct the view space. \citet{ICCV2007Gonzalez} decoupled variations on both the shape and texture into pose and expression/identity parts. The shape and texture are modeled by a bilinear model respectively.

An alternative way to model the pose rotation is by exploring 3D information \citep{PBCVPR2004Xiao}, and studies show that 3D linear face models generally have better qualifications than 2D linear models in three aspects: representational power, construction and real-time fitting \citep{IJCV2007Matthews}. \citet{IJCV2008Sung} deployed cylinder head models \citep{RPPAMI2000Cascia} to predict the global head pose parameters, which are fed to the subsequent AAM procedures. Their experimental results illustrate that face training by the combined method is more pose robust than that of AAM, having a 170\% higher tracking rate and 115\% wider pose coverage. \citet{CVPR2009Asthana} applied the 3D facial pose estimator \citep{RPTR2008Grujic} to obtain the pose range given a testing image. Facial feature points are then detected by the AAM trained from images with the corresponding view. They also exploited regression relationship from annotated frontal facial images to non-frontal facial images to handle pose and expression variation \citep{PR2011Asthana}. \citet{RPTR2008Grujic} presented a 3D AAM that does not require any pre-alignment of shapes, thanks to the inherent properties of complex spherical distributions: invariance to scale, translation and rotation. \citet{BMVC2010Martins, CVIU2013Martins} combined 3D PDM and a 2D appearance model through a full perspective projection. The fitting objective can be optimized by two methods based on the Lucas-Kanade framework \citep{PBIJCV2004Baker}: the simultaneous forwards additive algorithm and the normalization forwards additive algorithm. \citet{BMVC2011Hansen} presented a nonlinear shape model-based on Riemannian elasticity framework instead of linear subspace model (PDM) in a conventional AAM framework to handle the poor pose initialization problem. However, due to the complexity of the nonlinear shape formulation, the efficiency is reduced. \citet{FG2013Fanelli} proposed a 3D AAM-based on intensity and depth images. Random forest \citep{RPML2001Breiman} is explored to model the relationship between textures (from both intensity and depth images) and model parameters.

\textbf{Resolutions}: \citet{ECCV2006Dedeoglu, PAMI2007Dedeoglu} observed that classic AAM performed poorly in the case of low-resolution images. This is due to the image formation model of a typical charge-coupled device (CCD) camera. Consequently, they proposed a resolution-aware algorithm to adapt to low-resolution images which substitutes the classic fitting criterion of L2 norm error with a new formulation, taking the image formation model into account. \citet{BMVC2006Liu} trained several AAMs, each of which corresponds to a special resolution, to model compactness at lower resolution.

\textbf{Illumination}: Classic AAM models the texture variation with the Gaussian. Sometimes this assumption may result in errors when the illumination changes considerably. \citet{CVPR2007Kahraman} decomposed the original texture space into two orthogonal subspaces: identity and illumination subspace. Any texture can then be described by two projection vectors $\pmb{\beta}_{id}$ and $\pmb{\beta}_{illu}$:

\begin{equation}
\label{eq 27}
\textbf{a}=\textbf{a}_0+\textbf{P}_{id}\pmb{\beta}_{id}+\textbf{P}_{illu}\pmb{\beta}_{illu}.
\end{equation}
where $\textbf{P}_{id}$ and $\textbf{P}_{illu}$ consist of basis vectors which span the identity and illumination subspace respectively. \citet{IVC2010Kozakaya} explored multilinear analysis (tensor) to model the variations across face identity, pose, expression, and illumination. The tensor consists of an image tensor and a model tensor. The image tensor is utilized to estimate image variations which can be solved in a discretion or continuous manner. The model tensor is applied to construct a variation-specific AAM from a tensor representation.

\textbf{Outliers}: \citet{BMVC2007Roberts} observed that AAMs are not robust to a large set of gross outliers; they explored the Geman-McClure kernel (M-estimator) with two sets of learned scaling parameters to alleviate this problem.

\textbf{Occlusions}: AAM learns the texture model from a holistic view and faces challenges in achieving good performance, such as sensible to partial occlusion, while ASM opts for local texture descriptors. \citet{IJCV2007Sung} therefore combined ASM and AAM to give a united objective function:

\begin{equation}
\label{eq 28}
E=(1-\omega)(E_{aam}+E_{reg})+\omega{E_{asm}},
\end{equation}
where $E_{aam}$ and $E_{asm}$ are the residual errors of the AAM and ASM appearance model respectively, $E_{reg}$ is the regularization error term constrained on shape parameters, and $\omega$ is a trade-off parameter to balance the ASM residual error with other errors. \citet{CVIU2013Martins} proposed two robust fitting methods based on the Lucas-Kanade forwards additive method \citep{PBLK20Part3} to handle partial and self-occlusions.

\section{Regression-Based Methods}
\label{sec:4}
The aforementioned categories of methods mostly govern the shape variations through certain parameters, such as PDM coefficient vector $\pmb{\alpha}$ in ASM and AAM. By contrast, regression-based methods directly learn a regression function from image appearance (feature) to the target output (shape):

\begin{equation}
\label{eq 29}
\mathcal{M}:\mathcal{F}(\textbf{I})\rightarrow\textbf{x}\in\mathbb{R}^{2N},
\end{equation}
where $\mathcal{M}$ denotes the mapping from image appearance (feature) $\mathcal{F}(\textbf{I})$ to the shape $\textbf{x}$ and $\mathcal{F}$ is the feature extractor. Haar-like features \citep{RPIJCV2004Viola}, SIFT \citep{RPIJCV2004Lowe}, local binary patterns (LBP) \citep{RPPR1996Ojala} and other gradient-based features are generally used feature types.

\citet{IPMI2007Zhou} proposed a shape regression method based on boosting \citep{RPJCSS1997Freund, RPAS2000Friedman}. Their method proceeds in two stages: first, the rigid parameters are found by casting the problem as an object detection problem which is solved by a boosting-based regression method; secondly, a regularized regression function is learned from perturbed training examples to predict the non-rigid shape. Haar-like features are fed to the non-rigid shape regressors.

Kozakaya et al. \citep{CVPR2008Kozakaya, FG2008Kozakaya, IVC2010Kozakaya} proposed a weighted vector concentration approach to localize facial features without any specific prior assumption on facial shape or facial feature point configuration. In the training phase, grid sampling points are evenly placed on each face image and an extended feature vector is extracted for each sampling point of each training image. The extended feature vector is composed of histograms of oriented gradients (HOG descriptor \citep{RPCVPR2005Dalal}), $N$ directional vectors from the sampling point to all $N$ feature points, and local likelihood patterns at the feature points. In the detection phase, given an input face image, local descriptors corresponding to each sampling point are extracted. Then a nearest local pattern descriptor can be found for each sampling point of the input image among the descriptors located at the same position of training images using the approximate nearest neighbor search (ANNS) algorithm \citep{RPJACM1998Arya}. Simultaneously, a group of directional vectors and local likelihood patterns can also be obtained. Finally, feature points are computed from a weighted square distance from the point to the line through sampling points and the directional vector. Each facial feature point can be detected independently after all nearest neighbors are found by the ANNS method. This paper does not take faces with different expressions into consideration in their experiments.

In consideration of the nonlinear property of the facial feature localization problem and generalization ability, \citet{CVPR2010Valstar} deployed support vector regression to output the target point location from the input local appearance-based features (Haar-like features). To overcome the overfitting problem due to the high dimensionality of the Haar-like features, Adaboost regression is utilized to perform feature selection.

\citet{BMVC2011Kazemi} divided a face into four parts: eyes (left and right), nose and mouth. Several regression strategies such as ridge regression, ordinary least squares regression, principal component regression were then explored to regress the local appearance (a variant of HOG descriptor) of each part to the target landmark location. Their experimental results illustrate that these several regression methods, ridge regression achieves the best performance. Moreover, their method has comparable performance as to AAM methods but is more robust.

\begin{figure}[!t]
\centering
\includegraphics[width=0.8\columnwidth]{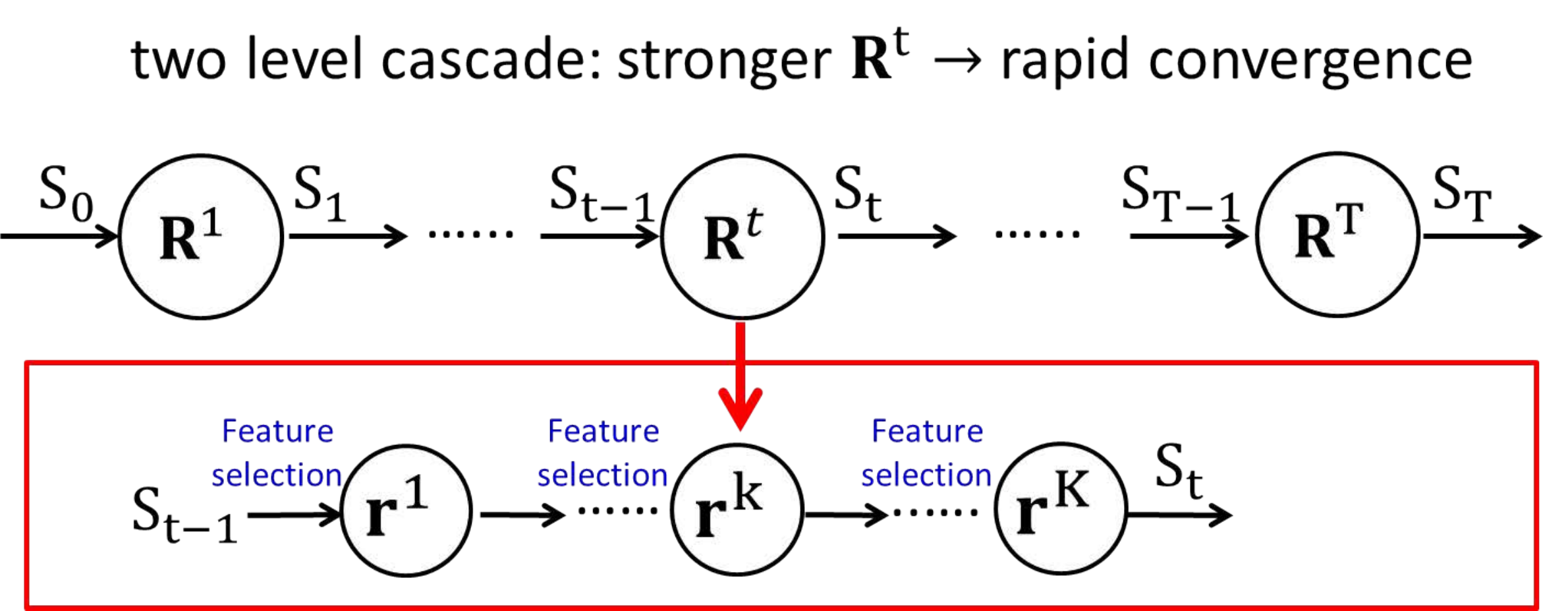}
\caption{Two-level cascaded regressor of \citep{CVPR2012Cao}.}
\label{fig:Fig 6}
\end{figure}

\citet{CVPR2012Cao} proposed a two-level cascaded learning framework (see Fig. \ref{fig:Fig 6}) based on boosted regression \citep{RPML2002Duffy}. Unlike the above method which learns the regression map of each landmark of those landmarks that correspond to the same component, this method directly learns a vectorial output for all landmarks. Shape-indexed features such as that in \citep{RPCVPR2010Dollar} are extracted from the whole image and are fed into the regressor. To reduce the complexity of feature selection but still achieve reasonable performance, the authors further proposed a correlation-based feature selection strategy. Each regressor ($\textbf{R}^t$ in Fig. \ref{fig:Fig 6}, $t=1,\cdots,T$) in the first level consists of cascaded random fern regressors ($\textbf{r}^k$ in Fig. \ref{fig:Fig 6}, $k=1,\cdots,K$) \citep{RPPAMI2010Ozuysal} in the second level. This method achieves state-of-the-art performance in a very efficient manner. In particular, it achieves the highest accuracy on the LFPW database: labeled face parts in the wild database \citep{CVPR2011Belhumeur}, images of which are taken under uncontrolled conditions.

Considering the method \citep{CVPR2012Cao} is not robust to occlusions and large shape variations, \citet{ICCV2013Burgos} improved it from three aspects. First, \citet{CVPR2012Cao} references pixel by its local coordinates with respect to its closest landmark, which is not enough against large pose variations and shape deformation. \citet{ICCV2013Burgos} proposed to reference pixels by linear interpolation between two landmarks. Secondly, Burgos-Artizzu et al. presented a strategy to incorporate the occlusion information into the regression which improves the robustness to occlusion. Thirdly, they designed a smart initialization restart scheme to deploying the similarity between different predictions resulted from different initializations. Experimental results on several existing databases in the wild and their newly constructed database illustrate the proposed method achieves state-of-the-art performance.

In view of the boosted regression in \citet{CVPR2012Cao}, which is a greedy method to approximate the function mapping from facial image appearance features to facial shape updates, \citet{CVPR2013Xiong} developed the supervised descent method (SDM) to solve a series of linear least squares problems as follows:

\begin{equation}
\label{eq 30}
\mathrm{argmin}_{\textbf{R}_k,\textbf{b}_k}\sum_{\textbf{d}^i}\sum_{\textbf{x}_k^i}\left\|\triangle\textbf{x}_*^i-\textbf{R}_k\pmb{\phi}_k^i-\textbf{b}_k\right\|^2,
\end{equation}
where $\triangle\textbf{x}_*^i=\textbf{x}_*^i-\textbf{x}_k^i$ is the ground truth difference between the truth shape $\textbf{x}_*^i$ of the $i$-th training image $\textbf{d}^i$ and the shape $\textbf{x}_k^i$ obtained from the $k$-th iteration, $\pmb{\phi}_k^i$ is the extracted SIFT features around the shape $\textbf{x}_k^i$ on the training image, $\textbf{R}_k$ is called the common descent direction in this paper and $\textbf{b}_k$ is a biased term. This method has a natural derivation process based on the Newton method. A series of $\{\textbf{R}_k,\textbf{b}_k\}$ are learned in the training stage, and in the testing stage they are applied to the SIFT features extracted from the testing image to update the shape sequentially. SDM efficiently achieves comparable performance to \citep{CVPR2012Cao} on database LFPW \citep{CVPR2011Belhumeur}.

\citet{PAMI2013Martinez} believed that each image patch evaluated by the regressors adds evidence to the target location rather than just taking the last estimate (the last iteration) into account and discarding the rest of these estimates. They aggregated all up-to-date local evidence obtained from support vector regression by an unnormalized mixture of Gaussian distributions. LBP is deployed as the local texture descriptor and a correlation-based feature selection method is introduced to reduce the dimensionality of LBP features.

\citet{CVPR2012Dantone} proposed a facial feature point detection by extending the concept of regression forests \citep{RPML2001Breiman, RPFTCGV2012Criminisi} to conditional regression forests. They claimed that it is difficult for general regression forests to learn the variations of faces with different head poses. The head pose is evaluated by regression forests. A regression forest is constructed, conditioned on the head pose (i.e. there is one regression forest corresponding to each head pose). In the testing phase, the probabilities of the head pose of an input testing image should be first calculated and, according to this distribution, the number of trees selected from each forest can be determined. Finally the position of each facial feature point can be computed through solving a mean-shift problem. Yang and Patras added structural information into the random regression and proposed a structured-output regression forest-based face parts localization method \citep{ACCV2012Yang}. Then, they \citep{ICCV2013Yang} proposed to deploy a cascade of sieves to refine the voting map obtained from random regression forest.

\citet{PR2012Rivera} casted the facial feature point detection problem as a regression problem. The input of a regressor consists of features extracted from input images, either pixel intensities or C1 features \citep{RPPAMI2007Serre}. The output of a regressor is PDM coefficients (shape parameters). The regressor is either a kernel ridge regression or $\epsilon$-support vector regression. Their experimental results show that kernel ridge regression with pixel intensities achieves the best performance when images have a low resolution.

Considering the fact existing parameterized appearance models do not sample parameter space uniformly, which may result in a biased model, \citet{ECCV2012Lozano} proposed a continuous regression method to solve this biased learning problem. Instead of discretely sampling the parameter space, this method directly integrates on the parameter space. A closed-form solution can be achieved. To alleviate the small sample size problem, the closed-form solution is further projected onto the principal components.

\section{Other Methods}
\label{sec:5}
In addition to the aforementioned three categories of methods, there are also some methods that do not belong to any of them. Some methods deploy graphical model to describe the relation between facial feature points, which are assigned to the sub-category of graphical model-based methods in the following text. Some methods align a set of facial images simultaneously, which is known as joint face alignment. Other methods may detect facial features points independently from the image texture and ignore the correlation between points, and we call this sub-category of methods independent detectors.

\subsection{Graphical Model-based Methods}
Graphical model-based FFPD methods mainly refer to tree-structure-based methods and Markov random field (MRF)-based methods. Tree-structure-based methods take each facial feature point as a node and all points as a tree. The locations of facial feature points can be optimally solved by dynamic programming. Unlike the tree-structure which has no loop, MRF-based methods model the location of all points with loops.

\citet{PBECCV2002Coughlan} developed a generative Bayesian graphical model that deployed separate models to describe shape variability (shape prior) and appearance variations (appearance likelihood) to find deformable shapes. The shape prior takes the location of each facial feature point and these points' normal orientation as a node in MRF. An edge map and an orientation map are calculated to model the appearance likelihood. A variant of the belief propagation method is utilized to optimize the problem. MRF has been also explored to constrain the relative position of all facial feature points obtained from the regression procedure in \citet{CVPR2010Valstar, PAMI2013Martinez}. \citet{CVPR2007Gu} learned a sparse Gaussian MRF structure to regularize the spatial configuration of face parts by lasso regression.

Unlike the method in \citet{PBECCV2002Coughlan} which models the shape prior only in a local neighborhood, \citet{CVPR2006Liang} proposed a method that incorporates a global shape prior directly into the Markov network. The local shape prior is enforced by denoting a line segment as a node of the constructed Markov network. Here, line segments draw from one facial point to another neighboring point. Subsequently, \citet{ECCV2006Liang} claimed that although CLM-based methods take the global shape prior into account, these methods neglect the neighboring constraint between points since they compute the response map of each point independently. Based on the thought in \citet{CVPR2006Liang}, Liang et al. further incorporated the PDM shape prior into their model.

Another work that considers both the local characteristics and global characteristics of facial shapes is the bi-stage component-based facial feature point detection method \citep{ICCV2007Huang}. The whole face shape is divided into seven parts. The shape of each part is modeled as a Markov network by taking each point as a node. Belief propagation is explored to find the locations of these components. Then, configurations of these components are constrained by the global shape prior described by the Gaussian process latent variable model.

\citet{CVPR2012Zhu} proposed a unified model for face detection, head pose estimation and landmark estimation. Their method is based on a mixture of trees, each of which corresponds to one head pose view. These different trees share a pool of parts. In the training stage, the tree structure is first estimated via Chow-Liu algorithm \citep{RPTIT1968Chow}. Then a model of a tree-structured pictorial structure \citep{RPIJCV2005Felzenszwalb} is constructed for each view. In the testing stage, the input image is scored by all tree structures respectively and the pixel locations corresponding to the tree with maximum score are the final landmark locations. \citet{VISAPP2012Michal} also modeled the relative position of facial feature points as a tree-structure. Since tree-structure-based methods only consider the local neighboring relation and neglect the global shape configuration, they may easily lead to unreasonable facial shape.

\subsection{Joint Face Alignment Methods}
Joint face alignment jointly aligns a batch of images undergoing a variety of geometric and appearance variations \citep{JACVPR2011Zhao}, motivated by the congealing-style joint alignment method \citep{RPPAMI2006Miller} and sparse and low-rank decomposition method \citep{RPPAMI2012Peng}. \citet{JACVPR2011Zhao} designed a joint AAM by assuming that the images of the same face should lie in the same linear subspace and the person-specific space should be proximate the generic appearance space. The problem is formulated as a nonlinear problem constrained by a rank term which can be transformed to a nuclear norm. An augmented Lagrangian method is explored to optimize the nonlinear problem.

\citet{JAECCV2012Smith} stated that the method \citep{JACVPR2011Zhao} breaks down under several common conditions, such as significant occlusion or shadow, image degradation, and outliers. Considering the fact that a non-parametric set of global shape models \citep{CVPR2011Belhumeur} results in excellent facial feature point localization accuracy on facial images undergoing significant occlusions, shadows, and pose and expression variation, they introduced the same shape model combined with a local appearance model into the joint alignment framework.

Different from the aforementioned two joint face alignment methods, which both incorporate the rank term into the objective, \citet{JAECCV2012Zhao} proposed a novel two-stage approach to align a set of images of the same person. The initial facial feature point estimation is first computed by an off-the-shelf approach \citep{ECCV2008Gu}. To distinguish the "good" alignments from the "bad" ones among all these initial estimations, a discriminative face alignment evaluation metric is designed by virtue of cascaded AdaBoost framework \citep{RPIJCV2004Viola} and Real AdaBoost \citep{RPAS2000Friedman}. Selected "good" alignments are utilized to improve the accuracy of "bad" ones through appearance consistency between the "bad" estimate and its selected K neighboring "good" estimates.

\citet{CVPR2009Tong, JACVIU2012Tong} proposed a semi-supervised facial landmark localization approach which utilizes a small number of manually labeled images. Their objective function is to minimize the sum of squared error of two distances: the distance between the labeled and unlabeled images, and the distance between the unlabeled images. To obtain a reasonable shape, an on-line learned PDM shape model is imposed as a constraint. To further improve the preciseness of the above model, they perform the above procedures in a coarse-to-fine manner, which proceeds by dividing the whole face into patches with different sizes at different levels.

\subsection{Independent Facial Feature Point Detectors}
The aforementioned methods predict the locations of all facial feature points or a group of points simultaneously. There are other methods which detect each point independently. Here, methods which do not rely on manually labeled images, such as approach \citep{PR2009Asteriadis}, are not included.

\citet{PBSMC2005Vukadinovic} detected each point by a local expert as utilized in CLM-based methods. Here, Gabor feature-based boosted classifier is utilized to classify the positive image patch from the negative image patch. The position with the peak response among the response map of each point is the sought location. \citet{CVPR2013Shen} proposed the detection of each facial feature point through a voting strategy on corresponding points on some exemplar images retrieved from the training dataset. The location corresponding to the peak response in each voting map is the estimated position.

Considering the fact that there is great variability among faces and facial features, such as eye centers and eye corners, \citet{CVPR2008Ding, PAMI2010Ding} employed subclass discriminant analysis \citep{RPPAMI2006Zhu} to divide vectors (features or context) of the same class into subclasses. Vectors centered on the facial feature point are called features and vectors centered on points surrounding the facial feature point are called context. The $K$-means clustering method is explored to divide each class into a number of subclasses. Given the detected face box, facial feature points can be exhaustively searched in some windows located relative to the bounding box by comparison with the learned subclasses at different scales. The final facial feature point is achieved by a voting strategy on different detected positions at different scales.

The advantage of independent facial feature point detectors is the initialization free character. One major disadvantage is the ambiguity problem. This means there exist more than one positions looking like the target landmark, especially under complex environment like deliberately disguise, occlusion or pose variation. To address this problem, \citet{ICCV2013Zhao} proposed to jointly estimate correct positions of all landmarks from some candidates obtained by independent facial feature point detectors.

\subsection{Deep Learning-Based Methods}
\citet{CVPR2012Luo} proposed a hierarchical face parsing method based on deep learning \citep{RPNC2006Hinton, RPScience2006Hinton}. They recast the facial feature point localization problem as the process of finding the label maps (segmentation) which clearly indicate the pixels belong to a certain component. The feature can then be easily obtained from the boundary of the label maps. The proposed hierarchical framework consists of four layers: face detector (the first layer), facial parts detectors (the second layer), facial component detectors (the third layer), and facial component segmentation. The structure of this model is somewhat like a pictorial structure \citep{RPIJCV2005Felzenszwalb}: the face detector can be seen as the root node and other detectors (part detectors and component detectors) as the child nodes. The objective function can be formulated in a Bayesian (maximum a posterior) form. The prior term denotes the spatial consistency between detectors of different layers and is modeled as the Gaussian distribution. The likelihood term represents the detectors and segmentation. All detectors can be learned by restricted Boltzmann machine \citep{RPNC2006Hinton} and segmentation can be learned by a deep autoencoder-like \citep{RPScience2006Hinton} method. Inspired by Luo et al. \citep{CVPR2012Luo}, \citet{CVPR2013Smith} deployed exemplar-based strategy as in \citet{CVPR2011Belhumeur} to parse a face image.

\citet{CVPR2013Sun} proposed a three-level cascaded deep convolutional network framework for point detection in a coarse-to-fine manner. Each level is composed of several numbers of convolutional networks. The first level gives an initial estimate to the point position and the following two levels then refine the obtained initial estimate to a more accurate one. Though great accuracy can be achieved, this method needs to model each point by a convolutional network which improves the complexity of the whole model. Moreover, with the increase in the number of facial feature points, the time consumption to detect all points is high.

\citet{CVPR2013Wu} explored deep belief networks to capture face shape variation due to facial expression variations and utilized a 3-way restricted Boltzmann machine to capture the relationship between frontal face shapes and non-frontal face shapes. They applied the proposed model to facial feature tracking.

\section{Evaluations}
\label{sec:6}
\subsection{Databases}
There are many face databases publically available due to the easy acquisition of images and the fast development of social networks such as Facebook, Flickr, and Google+. The ground truth facial feature points are usually labeled manually by employing workers or through crowdsourcing, e.g. the Amazon mechanical turk (MTurk). Each face image is generally labeled by several workers and the average of these labeled results is taken as the final ground truth. These face databases can be classified into two categories: databases captured in controlled conditions and databases captured in uncontrolled conditions (i.e. in the wild). Controlled databases are taken under the framework of predefined experimental settings such as the variation of illumination, occlusions, head pose and facial expressions. Databases in the wild are generally collected from websites such as Facebook and Flickr. Table \ref{tab:table 2} describes representation collections which are popularly used in empirical studies.

\begin{table*}
\centering
\caption{Datasets for Facial Feature Point Detection}
\label{tab:table 2}
\scalebox{0.8}{
\begin{tabular}{|p{17cm}|}
\hline
\textbf{Databases Collected under Well-Controlled Conditions:}
\\
\\
\textbf{CMU Multi-PIE2008} - CMU Multi-PIE \citep{DB2010MultiPIE} face database was collected in four sessions between October 2004 and March 2005. It aims to support the development of algorithms for recognition of faces across pose, illumination and expression conditions. This database contains 337 subjects and more than 750,000 images for 305 GB of data. A total of six different expressions are recorded: neutral, smile, surprise, squint, disgust and scream. Subjects were recorded across 15 views and under 19 different illumination conditions. A subset of this database has been labeled either 68 points or 39 points depending on their view but landmarks are not published online. Details on obtaining this dataset can be found at: \url{http://www.multipie.org}.\\
\\
\textbf{Extended M2VTS database1999 (XM2VTS)} - XM2VTS database \citep{DB1999XM2VTS} collected 2,360 color images, sound files and 3D face models of 295 people. The database contains four recordings of these 295 subjects taken over a period of four months. Each recording was captured when the subject was speaking or rotating his/her head. This database is available on request at: \url{www.ee.surrey.ac.uk/CVSSP/xm2vtsdb/}. These 2,360 color images are labeled with 68 landmarks and are published online: \url{http://personalpages.manchester.ac.uk/staff/timothy.f.cootes/data/xm2vts/xm2vts_markup.html}.\\
\\
\textbf{AR1998} - AR database \citep{DB1998AR} contains over 4,000 color images corresponding to the faces of 126 people (70 men and 56 women). Images were taken under strictly controlled conditions and with different facial expressions, illumination conditions, and occlusions (sunglasses and scarf). Each person appeared in two sessions, separated by two weeks. Ding and Martinez \citep{PAMI2010Ding} manually annotated 130 landmarks on each face image which have been published online with the database: \url{www2.ece.ohio-state.edu/~aleix/ARdatabase.html}.\\
\\
\textbf{IMM2004} - IMM database \citep{DB2004IMM} contains 240 color images of 40 persons (7 females and 33 males). Each image is labeled with 58 landmarks around the eyebrows, eyes, nose, mouth and jaw. Face images and landmarks can be downloaded at: \url{http://www2.imm.dtu.dk/~aam/datasets/datasets.html}.\\
\\
\textbf{MUCT2010} - MUCT2010 database \citep{DB2010MUCT} consists of 3,755 face images of 276 subjects and each image is marked with 76 manual landmarks. Faces in this database are captured under different lighting conditions, at various ages, and are of several different ethnicities. The database is available at: \url{www.milbo.org/muct/}.\\
\\
\textbf{PUT2008} - PUT database \citep{DB2008PUT} collected 9,971 high resolution images $(2048\times{1536})$ of 100 people taken in partially controlled illumination conditions with rotations along the pitch and yaw angle. Each image is labeled with 30 landmarks. A subset of 2,193 near-frontal images is provided with 194 control points. The database is available at: \url{https://biometrics.cie.put.poznan.pl/index.php?option=com_content&view=article&id=4&Itemid=2&lang=en}.\\
\hline
\textbf{Databases in the wild:}
\\
\\
\textbf{BioID2001} - BioID database \citep{DB2001BioID} was recorded in an indoor lab environment, but "real world" conditions were used. This database contains 1,521 grey level face images of 23 subjects and each image is labeled with 20 landmarks. This database is available at: \url{http://www.bioid.com/index.php?q=downloads/software/bioid-face-database.html.}\\
\\
\textbf{LFW2007} - LFW database \citep{DB2007LFW} contains 13,233 face images of 5,749 subjects collected from the web. Each face in the database has been labeled with the name of the person pictured. 1,680 of the people pictured have two or more distinct photos in the data set. The constructors of this database did not provide manually labeled landmarks but there are other available sites: \citep{VISAPP2012Michal} \url{http://cmp.felk.cvut.cz/~uricamic/flandmark/ (7 landmarks)}; \citep{CVPR2012Dantone} \url{http://www.dantone.me/datasets/facial-features-lfw/ (10 landmarks)}.\\
\\
\textbf{Annotated Facial Landmarks in the Wild 2011(AFLW)} - AFLW database \citep{DB2011AFLW} is a large-scale, multi-view, real-world face database with annotated facial feature points. Images were collected from Flickr using a wide range of face relevant key words such as face, mugshot, and profile face. This database includes 25,993 images in total and each image is labeled with 21 landmarks. It is available at: \url{http://lrs.icg.tugraz.at/research/aflw/}.\\
\\
\textbf{Labeled Face Parts in the Wild 2011 (LFPW)} - LFPW database \citep{CVPR2011Belhumeur} is composed of 1,400 face images (1,100 as the training set and the other 300 images are taken as the testing set) downloaded from the web using simple text queries on websites such as Google.com, Flickr.com, and Yahoo.com. Due to copyright issues, the authors did not distribute image files but provided a list of image URLs. However, some image links are no longer available. 35 landmarks are labeled in total;29 of them are usually utilized in literatures. More information can be found at: \url{http://homes.cs.washington.edu/~neeraj/databases/lfpw/}.\\
\\
\textbf{Annotated Faces in the Wild 2012 (AFW)} - AFW database \citep{CVPR2012Zhu} contains 205 images with a highly cluttered background and large variations both in face scale and pose. Each image is labeled with 6 landmarks and the bounding box of the corresponding face. The dataset is available at: \url{http://www.ics.uci.edu/~xzhu/face/}.\\
\\
\textbf{Helen2012} - Helen database \citep{ECCV2012Le} contains 2,300 high resolution face images collected from Flickr.com. Each face image is labeled with 194 landmarks. More information about this database can be found at: \url{http://www.ifp.illinois.edu/~vuongle2/helen/}.\\
\\
\textbf{300 Faces in-the-Wild Challenge (300-W) 2013} - 300-W database is a mixed database consisting of face images from several published databases (LFPW, Helen, AFW, and XM2VTS) and a new collected database IBUG. All these images are re-annotated with 68 landmarks. This database is published for the first Automatic Facial Landmark Detection in-the-Wild Challenge (300-W 2013) held in conjunction with the International Conference on Computer Vision 2013. This database is available at: \url{http://ibug.doc.ic.ac.uk/resources/300-W/}.\\
\\
\textbf{Caltech Occluded Faces in the Wild (COFW) 2013 }- COFW database \citep{ICCV2013Burgos} is composed of 1,007 face images showing large variations in shape and occlusions due to differences in pose, expression, use of accessories such as sunglasses and hats and interactions with objects (e.g. food, hands, microphones, \textit{etc.}). 29 points are marked for each image. The major difference between this database and other ones is that each landmark is explicitly labeled whether it is occluded. This database presents a great challenging task for facial feature point detection due to the large amount and variety of occlusions and large shape variations. This database is available at: \url{http://www.vision.caltech.edu/xpburgos/ICCV13/}.\\
\hline
\end{tabular}}
\end{table*}

\subsection{Comparisons and Discussions}
The distance from the estimated points to the ground truth normalized by the inter-ocular distance and the number of points is a common informative metric for evaluating a facial feature point detection system (named mean normalized error, MNE, in the following text). Sometimes the figure of the proportion of testing images with the increase of MNE is plotted as a comparison metric among different approaches. The performance of facial feature point detection methods cannot be verified by experimenting on each database listed as Table \ref{tab:table 2} shown since there are too many databases. Table \ref{tab:table 3} shows the published performance of representative methods of aforementioned categories on several different databases.

\begin{figure*}
\centering
\includegraphics[width=1.8\columnwidth]{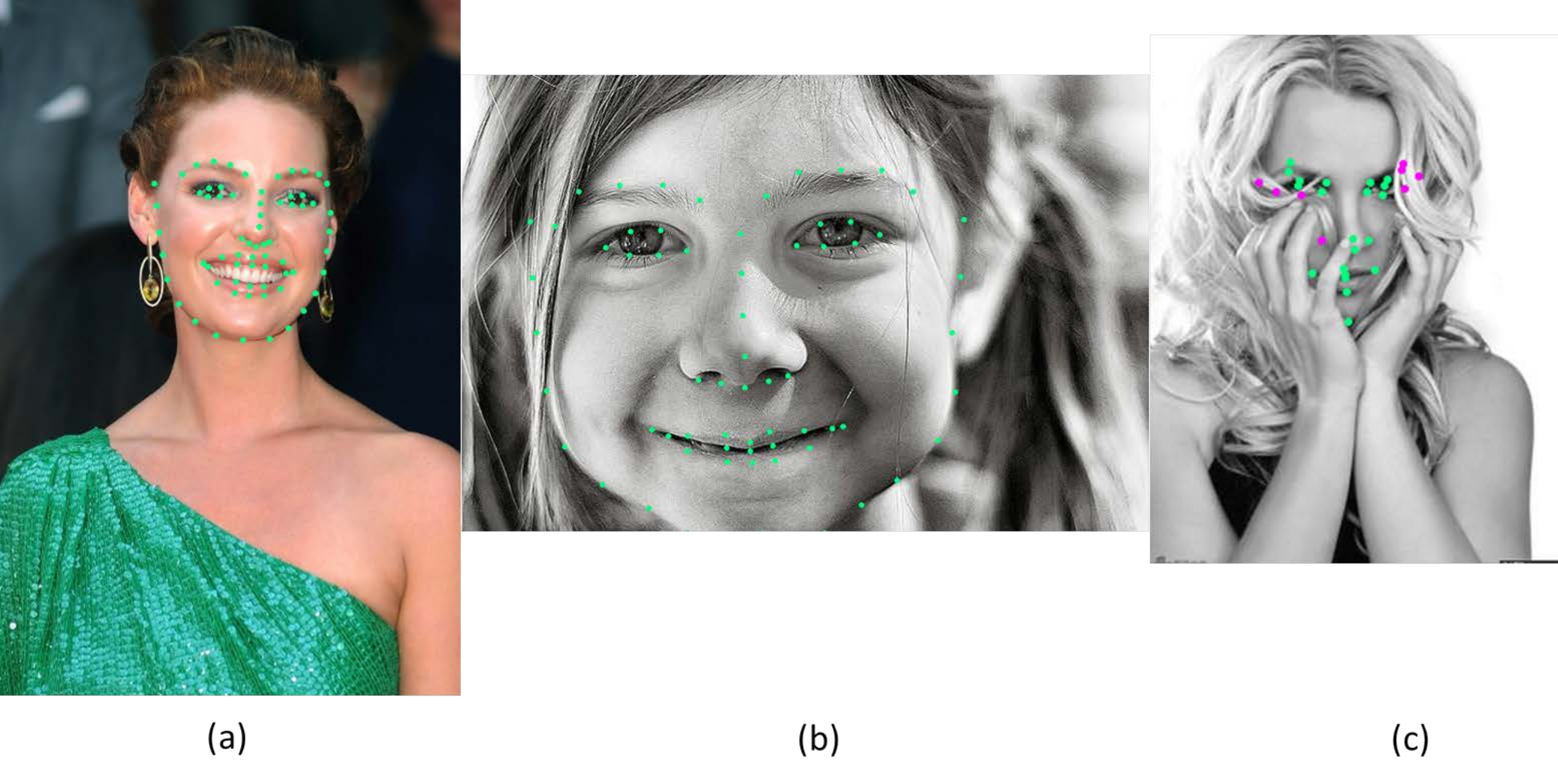}
\caption{Example faces from (a) LFPW, (b) Helen database and (c) COFW. The dots in deep red indicate corresponding points are occluded.}
\label{fig:Fig 7}
\end{figure*}

\begin{table*}
\centering
\caption{Mean Normalized Error of Representative Methods on Various Databases}
\label{tab:table 3}
\scalebox{0.7}{
\begin{tabular}{|c|c|c|c|}
\hline
\textbf{method} & \textbf{Database (\#Training Images+\#Testing Images)} & \textbf{MNE}$(\times10^{(-2)})$ & \textbf{\#Landmarks} \\
\hline
Belhumeur \citep{CVPR2011Belhumeur} & LFPW $(1100+300)$	& 3.99 & 29 \\
\hline
Cao \citep{CVPR2012Cao} & LFPW$(2000+500)$ & 3.43 & 29 \\
Xiong \citep{CVPR2013Xiong} & LFPW $(884+245)$ & 3.47 & 29 \\
\hline
Burgos-Artizzu \citep{ICCV2013Burgos} & LFPW$(845+194)$ & 3.50 & 29 \\
\hline
Xiong \citep{CVPR2013Xiong} & LFW-A\&C \citep{CVPR2011Saragih} $(604+512)$ & 2.7 & 66 \\
\hline
Xiong \citep{CVPR2013Xiong} & Multi-PIE, LFW-A\&C (training) + RU-FACS \citep{PBIJCV2004Matthews} (test) & 5.03 & 49 \\
\hline
Sukno \citep{PAMI2007sukno} & XM2VTS $(800+400)$ & 2.03 & 64 \\
\hline
Sukno \citep{PAMI2007sukno} & AR $(180+90)$ & 1.63 & 98 \\
\hline
Le \citep{ECCV2012Le} & MUCT+BioID $3755+1521$ & 4.5 & 17 \\
\hline
Le \citep{ECCV2012Le} & Helen$(2000+330)$ & 9.1 & 194 \\
\hline
Dantone \citep{CVPR2012Dantone} & LFW $(13000+1000)$ & 6.985 & 10 \\
\hline
Valstar \citep{CVPR2010Valstar} & FERET\citep{DB2000FERET}+MMI\citep{DB2010Valstar} $(360+40)$ & 5.11 & 22 \\
\hline
Martinez \citep{PAMI2013Martinez} & MMI\citep{DB2010Valstar}+FERET\citep{DB2000FERET}+XM2VTS+BioID $946+1000$ & 3.575 & 20\\
\hline
Ding \citep{CVPR2008Ding} & Americal Sign Lnguage Sentences (766 frames) & 6.23 & 98 \\
\hline
Ding \citep{PAMI2010Ding} & Collected training database+AR+XM2VTS$(51664+1200)$ & 8.4 & 98 \\
\hline
Wu \citep{CVPR2013Wu} & MMI\citep{DB2010Valstar}(196) & 5.5275 & 26 \\
\hline
Michal \citep{VISAPP2012Michal} & LFW $(6919+2316)$ & 5.4606 & 8 \\
\hline
\end{tabular}}
\end{table*}
To further illustrate the characteristics of various categories of methods, we have collected some software published online and listed them as shown in Table \ref{tab:table 4}.
\begin{table*}
\centering
\caption{Published Software Collection}
\label{tab:table 4}
\scalebox{0.8}{
\begin{tabular}{|c|c|}
\hline
\textbf{Method} & \textbf{Website} \\
\hline
Vukadinovic \citep{PBSMC2005Vukadinovic} & \url{http://ibug.doc.ic.ac.uk/resources/fiducial-facial-point-detector-20052007/} \\
\hline
Milborrow \citep{ECCV2008Milborrow} & \url{http://www.milbo.users.sonic.net/stasm/}\\
\hline
Inverse compositional AAM & \url{http://sourceforge.net/projects/icaam/files/}\\
\hline
Valstar \citep{CVPR2010Valstar} & \url{http://ibug.doc.ic.ac.uk/resources/facial-point-detector-2010/} \\
\hline
Hansen \citep{BMVC2011Hansen} & \url{https://svn.imm.dtu.dk/AAMLab/svn/AAMLab/trunk/ (username: guest, password: aamlab)} \\
\hline
Saragih \citep{IJCV2011Saragih} & \url{https://github.com/kylemcdonald/FaceTracker}\\
\hline
Tzimiropoulos \citep{ACCV2012Tzimiropoulos} & \url{http://ibug.doc.ic.ac.uk/resources/aoms-generic-face-alignment/} \\
\hline
Rivera \citep{PR2012Rivera} & \url{http://cbcsl.ece.ohio-state.edu/downloads.html} \\
\hline
Zhu \citep{CVPR2012Zhu} & \url{http://www.ics.uci.edu/~xzhu/face/} \\
\hline
Dantone \citep{CVPR2012Dantone} & \url{http://www.dantone.me/projects-2/facial-feature-detection/} \\
\hline
Michal \citep{VISAPP2012Michal} & \url{http://cmp.felk.cvut.cz/~uricamic/flandmark/} \\
\hline
Sun \citep{CVPR2013Sun} & \url{http://mmlab.ie.cuhk.edu.hk/archive/CNN_FacePoint.htm} \\
\hline
Asthana \citep{CVPR2013Asthana} & \url{https://sites.google.com/site/akshayasthana/clm-wild-code?} \\
\hline
Xiong \citep{CVPR2013Xiong} & \url{www.humansensing.cs.cmu.edu/intraface} \\
\hline
Martinez \citep{PAMI2013Martinez} & \url{http://ibug.doc.ic.ac.uk/resources/facial-point-detector-2010/} \\
\hline
Yu \citep{ICCV2013Yu} & \url{http://www.research.rutgers.edu/?xiangyu/face_align.html} \\
\hline
Tzimiropoulos \citep{ICCV2013Tzimiropoulos} & \url{http://ibug.doc.ic.ac.uk/resources} \\
\hline
Burgos-Artizzu \citep{ICCV2013Burgos} & \url{http://www.vision.caltech.edu/xpburgos/ICCV13/} \\
\hline
\end{tabular}}
\end{table*}
Eight representative methods were chosen for study: DRMF-CLM \citep{CVPR2013Asthana}, OPM-CLM \citep{ICCV2013Yu}, FF-AAM \citep{ICCV2013Tzimiropoulos}, CNN-DL \citep{CVPR2013Sun}, the graphical model (GM) method  \citep{CVPR2012Zhu}, BorMan-Regression \citep{CVPR2010Valstar}, SDM-Regression \citep{CVPR2013Xiong}, and RCPR-Regression \citep{ICCV2013Burgos}. We localized FFPs in three databases, COFW \citep{ICCV2013Burgos}, LFPW \citep{CVPR2011Belhumeur}, and Helen \citep{ECCV2012Le}, using the published software. The 68 re-annotated landmarks of the "300 Faces in-the-Wild Challenge" were used as the ground truth for images in LFPW and Helen. For COFW, we used the augmented version presented in \citep{ICCV2013Burgos}, which contains 1345 training images and 507 test images. Examples from these three databases are shown in Fig. \ref{fig:Fig 7}.

Since only the trained models, and not the source code, were published in some cases, it was difficult to make equitable comparisons (for example, some software contained different face detectors). In addition, different methods labeled different numbers of facial landmarks (see Table \ref{tab:table 5}). In Table \ref{tab:table 5}, "any" denotes that the authors published the training code, and thus the models could be trained for different numbers of facial landmarks. Face detection rates were quantified according to the percentage of detected faces being labeled in the corresponding database. "GM-99" and "GM-1050" indicate graphical models composed of 99 and 1050 parts, respectively. Errors were measured as the percentage of the interocular distance $d_{io}$, as shown in equation (\ref{eq 31}), i.e., the mean normalized error (MNE), where $\textbf{x}_{(i)}^e$ is the $i$-th estimated point and $\textbf{x}_{(i)}^g$ is its corresponding ground truth:

\begin{equation}
\label{eq 31}
e=\frac{\sum_{i=1}^N\left\|\textbf{x}_{(i)}^e-\textbf{x}_{(i)}^g\right\|_2}{N\times{d_{io}}}\times{100\%},
\end{equation}

\begin{table*}
\newcommand{\tabincell}[2]{\begin{tabular}{@{}#1@{}}#2\end{tabular}}
\centering
\caption{Mean Normalized Error of Eight Representative Methods on Three Databases: COFW, LFPW, and Helen}
\label{tab:table 5}
\scalebox{0.8}{
\begin{tabular}{|c|c|p{1.5cm}<{\centering}|p{1.5cm}<{\centering}|p{1.5cm}<{\centering}|p{1.5cm}<{\centering}|p{1.5cm}<{\centering}|p{1.5cm}<{\centering}|p{1.5cm}<{\centering}|p{1.5cm}<{\centering}|p{1.5cm}<{\centering}|}
\hline
\multicolumn{2}{|c|}{ }& DRMF \citep{CVPR2013Asthana} & OPM \citep{ICCV2013Yu} & FF \citep{ICCV2013Tzimiropoulos} & CNN \citep{CVPR2013Sun} & GM-99 \citep{CVPR2012Zhu} & GM-1050 \citep{CVPR2012Zhu} & Borman \citep{CVPR2010Valstar} & SDM \citep{CVPR2013Xiong} & RCPR \citep{ICCV2013Burgos} \\
\hline
\multicolumn{2}{|c|}{\#landmarks} & 66 & 66 & any & 5 & 68 & 68 & 29 & 49 & any\\
\hline
\multirow{3}{*}{\tabincell{c}{Face detection \\rate (\%)}} & COFW & 70.22 & 86.00 & 100 & 72.98 & 79.68 & 79.68 & 50.30 & 71.40 & 100\\
\cline{2-11}
 &LFPW & 73.21 & 92.86 & 100 & 96.88 & 89.29 & 88.39 & 76.34 & 87.95 & 100 \\
\cline{2-11}
 &Helen & 63.03 & 89.39 & 100 & 95.76 & 92.73 & 92.42 & 65.15 & 93.64 & 100\\
\hline
\multirow{3}{*}{Error (\%)} & COFW & 9.3666 & 11.1453 & 12.2417 & 5.4457 & 12.3449 & 11.8249 & 12.8179 & 6.9927 & 8.7382 \\
\cline{2-11}
&LFPW & 7.2202 & 10.3122 & 7.3907 & 5.7649 & 14.1085 & 14.5332 & 10.7461 & 5.3600 & 6.4350 \\
\cline{2-11}
& Helen & 8.2878 & 11.5897 & 8.9364 & 3.9133 & 13.4897 & 13.4176 & 11.2004 & 5.8397 & 5.4654 \\
\hline
\end{tabular}}
\end{table*}

\begin{figure}
\centering
\includegraphics[width=0.8\columnwidth]{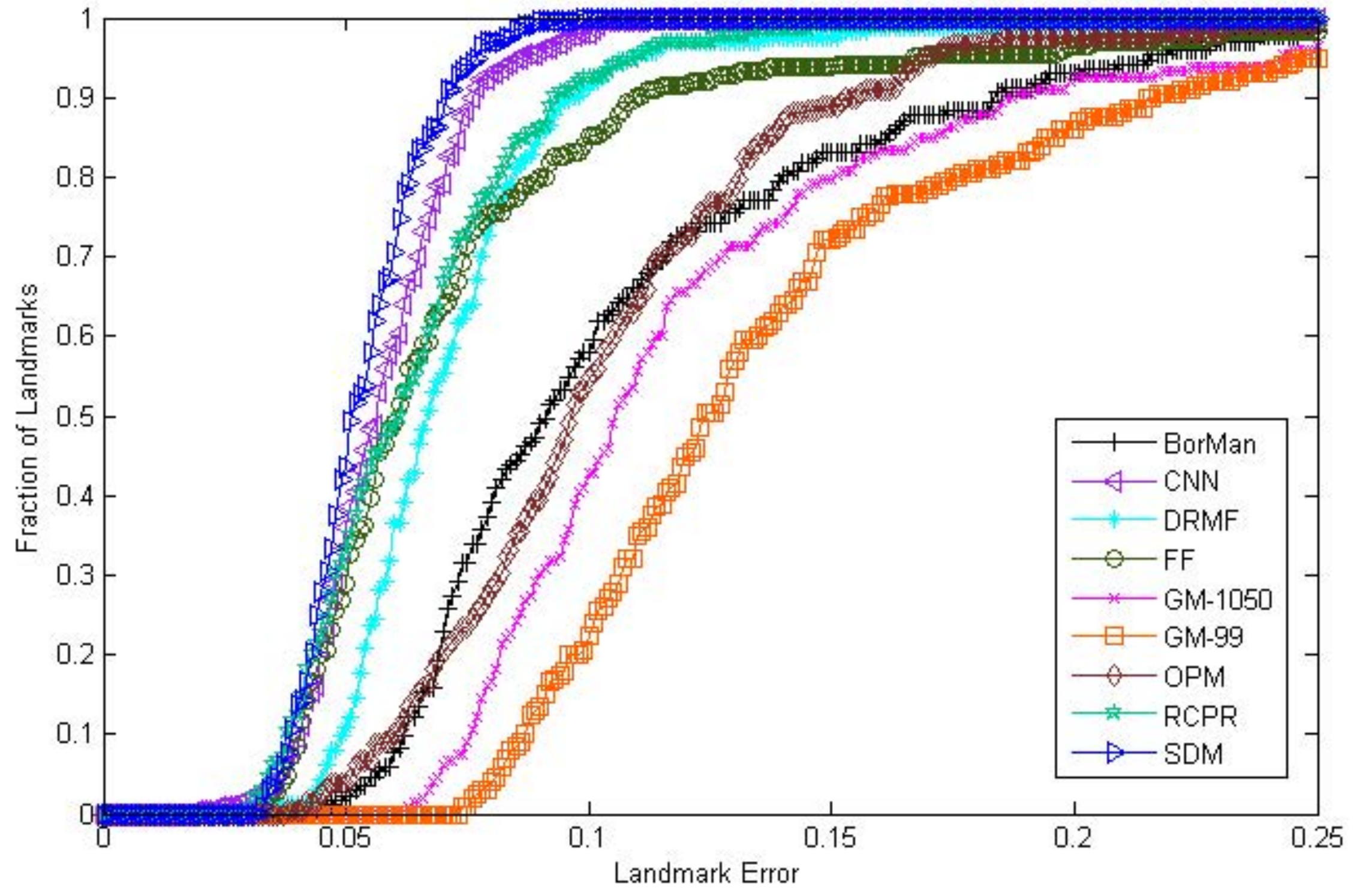}
\caption{Cumulative error curves on the LFPW database.}
\label{fig:Fig 8}
\end{figure}

\begin{figure}
\centering
\includegraphics[width=0.8\columnwidth]{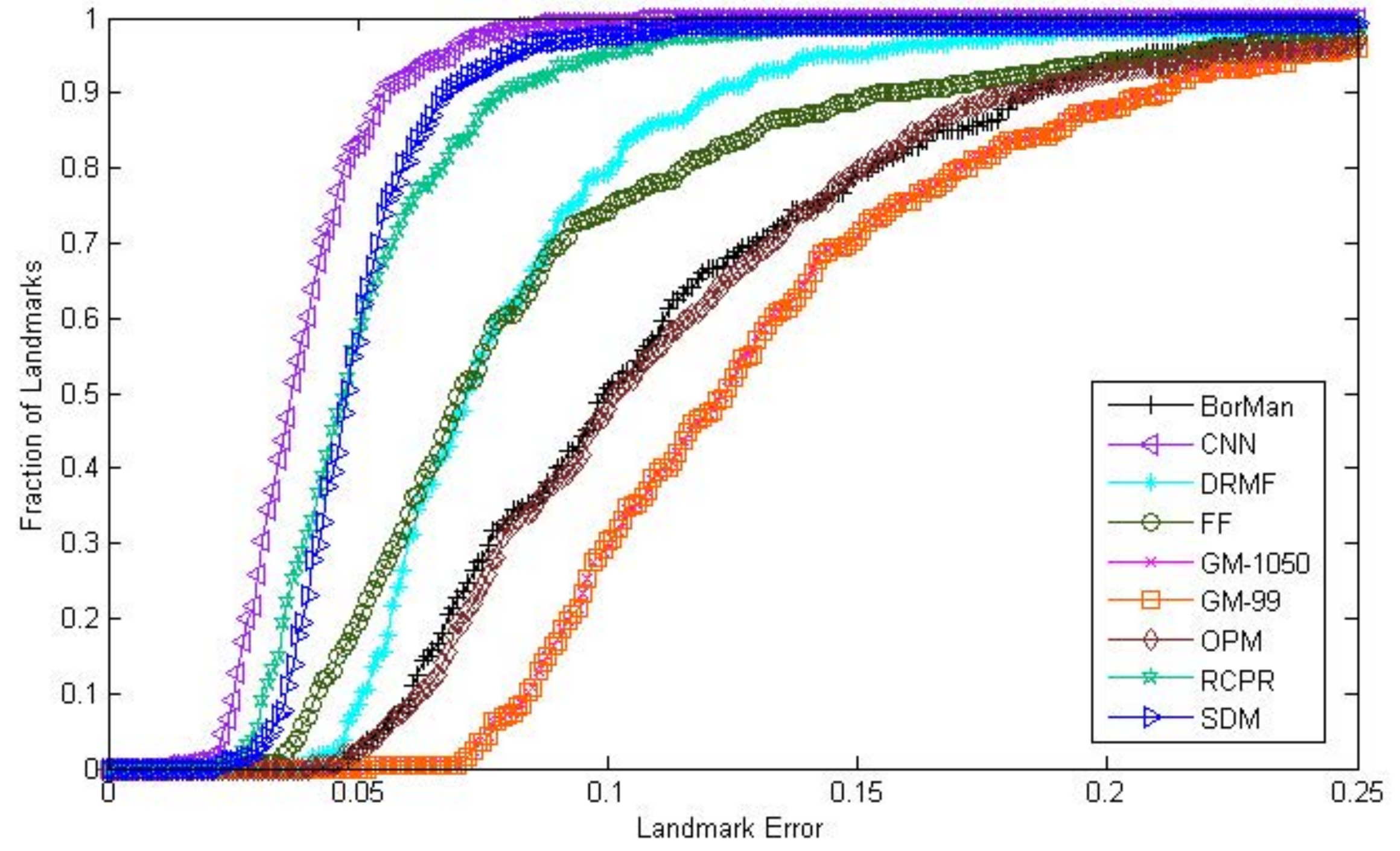}
\caption{Cumulative error curves on the Helen database.}
\label{fig:Fig 9}
\end{figure}

\begin{figure}
\centering
\includegraphics[width=0.8\columnwidth]{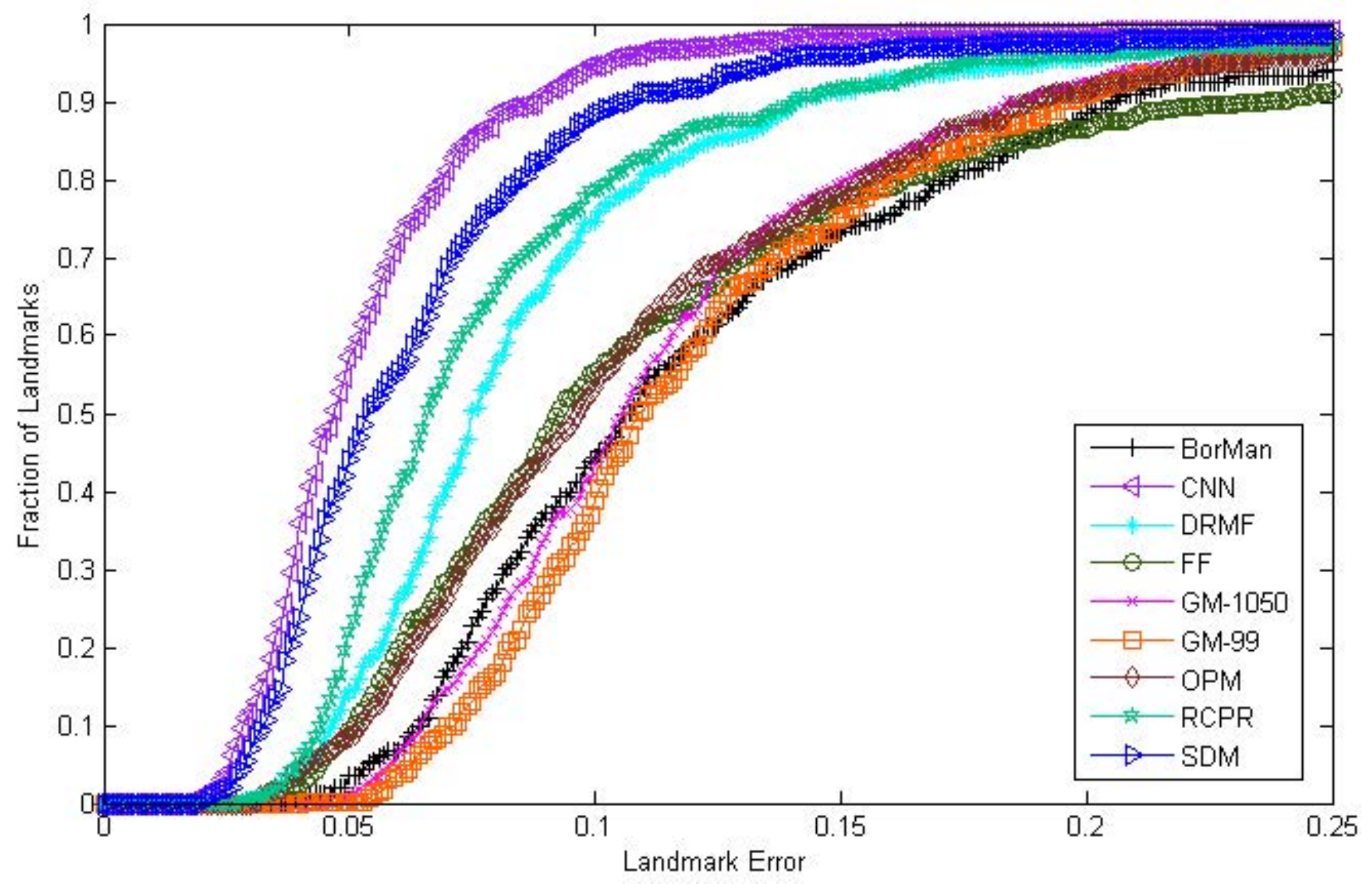}
\caption{Cumulative error curves on the COFW database.}
\label{fig:Fig 10}
\end{figure}

Fig. \ref{fig:Fig 8}, Fig. \ref{fig:Fig 9}, and Fig. \ref{fig:Fig 10} show the cumulative error curves for the above three databases. It can be seen that CNN \citep{CVPR2013Sun} achieves promising performance on all three databases. There are two main reasons for this: first, deep learning is highly capable of performing feature learning followed by classification or detection, especially when there are many training samples (CNN utilizes approximately ten thousand training samples); secondly, CNN detects five characteristic points: the center of the two pupils, the nose tip, and the two eye corners, which are relatively easy to detect. The cascaded regression method, SDM \citep{CVPR2013Xiong}, also achieves good performance for detecting 49 facial points distributed around the eyebrows, eyes, nose, and mouth, and without points around the outline of the face. RCPR, another cascaded regression method, also appears promising, although inferior to SDM; this is likely to be because SDM fails to detect several difficult test images and detects 49 points without the facial outline. Table \ref{tab:table 6} shows a comparison of the normalized error of RCPR retrained on 49 points on the same faces detected by SDM. The model could not be retrained on COFW, since 29 points label the faces in this database. The recomputed normalized error of RCPR on the COFW database was therefore calculated on the faces detected by SDM, and the details are shown in Table \ref{tab:table 6}.

\begin{table}
\centering
\caption{Recalculation of mean normalized error RCPR method.}
\label{tab:table 6}
\begin{tabular}{|p{3cm}<{\centering}|c|c|c|}
\hline
 & COFW & LFPW & Helen \\
\hline
RCPR \citep{ICCV2013Burgos} (\%) & 6.9557 & 5.0030 & 4.2730 \\
\hline
SDM \citep{CVPR2013Xiong} (\%) & 6.9927 & 5.3600 & 5.8397 \\
\hline
\end{tabular}
\end{table}

GM \citep{CVPR2012Zhu} is trained on the Multi-PIE database \citep{DB2010MultiPIE}, which is captured under laboratory conditions, but has inferior performance on real-world databases. Although the fast AAM fitting (FF) method \citep{ICCV2013Tzimiropoulos} achieves moderate performance, in our experience this method is very sensitive to the initialization. Of the four categories of methods, cascaded regression-based methods (e.g., \citep{CVPR2012Cao, CVPR2013Xiong, ICCV2013Burgos}) and CNN \citep{CVPR2013Sun} have the best performance.

\begin{figure}[!t]
\centering
\includegraphics[width=1\columnwidth]{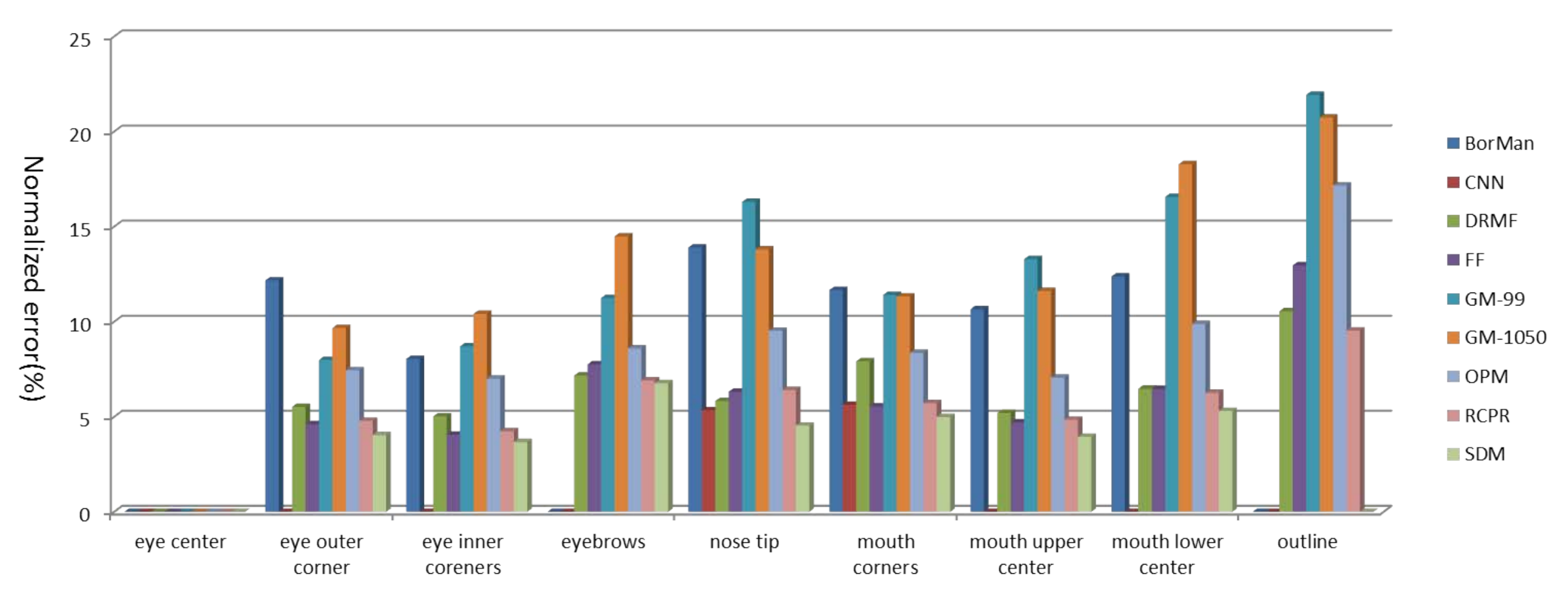}
\caption{Comparison of detection error of different landmarks (or parts): LFPW database.}
\label{fig:Fig 11}
\end{figure}

\begin{figure}[!t]
\centering
\includegraphics[width=1\columnwidth]{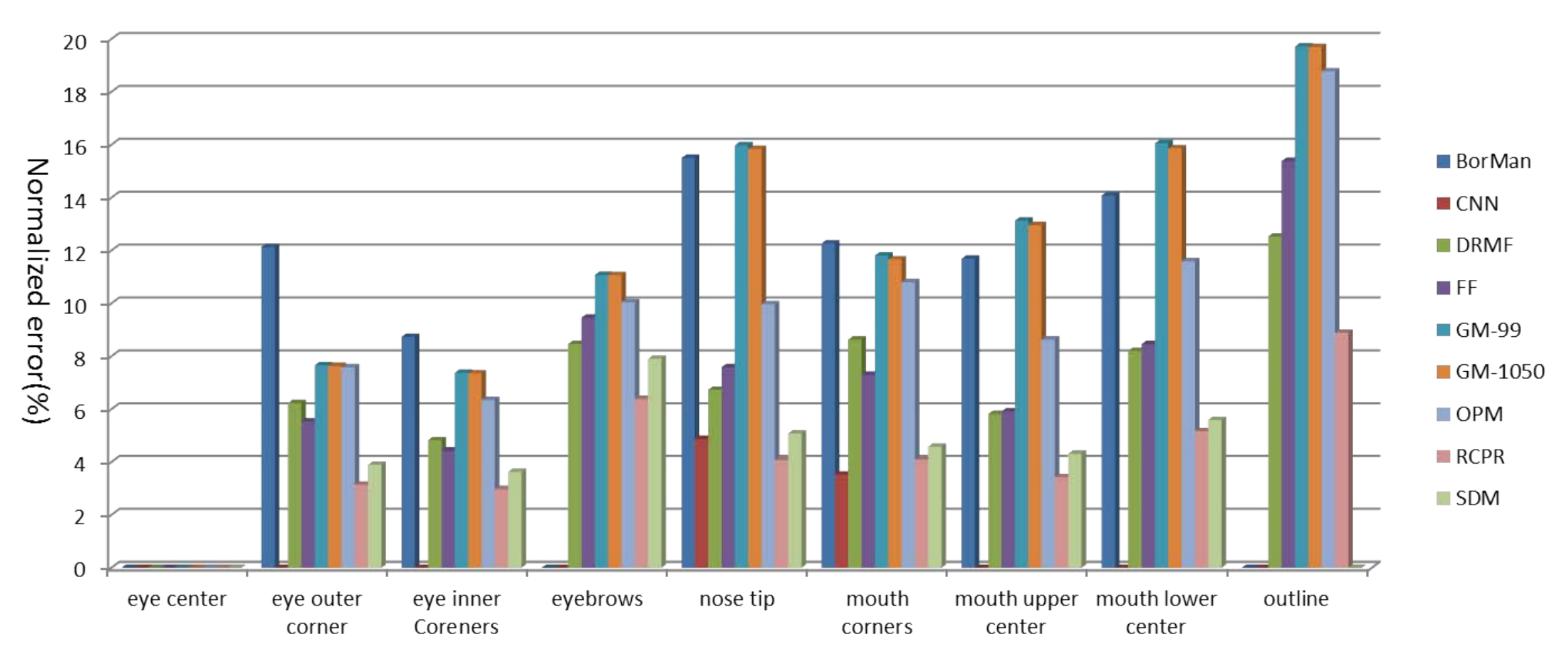}
\caption{Comparison of detection error of different landmarks (or parts): Helen database.}
\label{fig:Fig 12}
\end{figure}

\begin{figure}[!t]
\centering
\includegraphics[width=1\columnwidth]{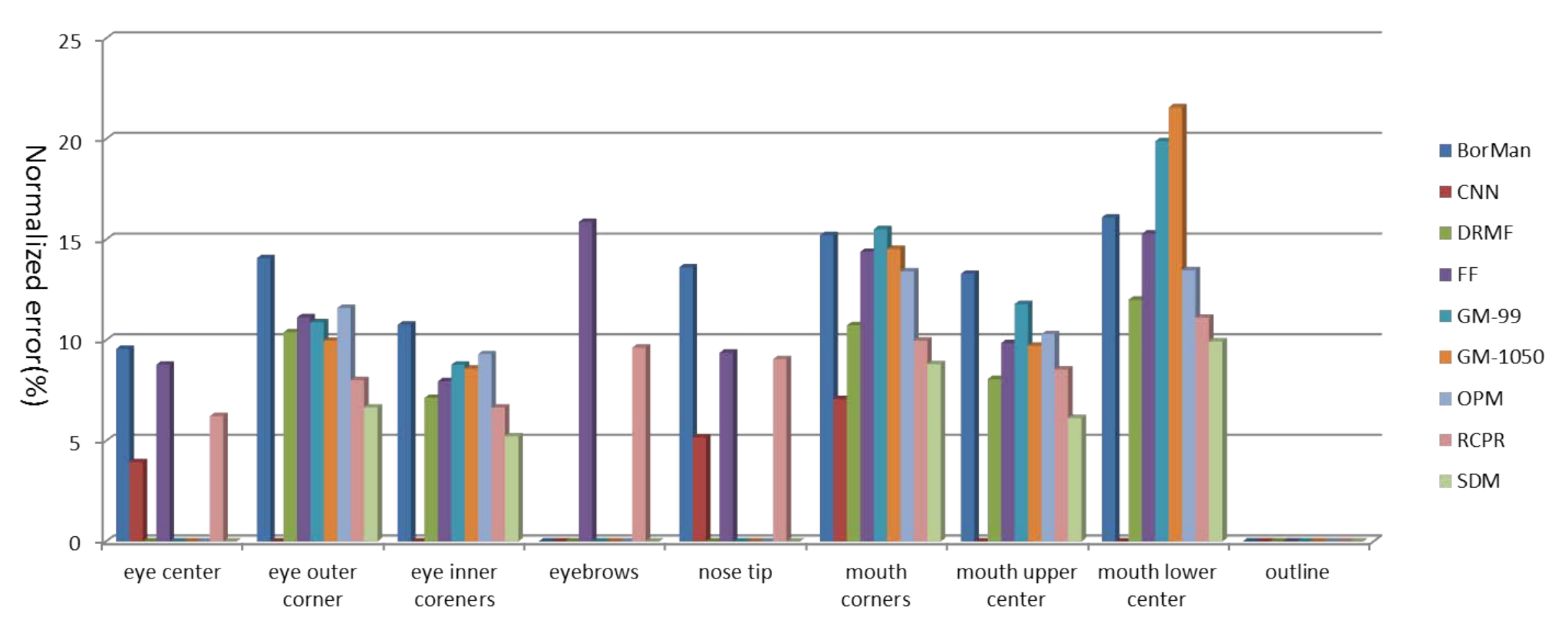}
\caption{Comparison of detection error of different landmarks (or parts): COFW database.}
\label{fig:Fig 13}
\end{figure}

Fig. \ref{fig:Fig 11}, Fig. \ref{fig:Fig 12}, and Fig. \ref{fig:Fig 13} show the detection errors for different landmarks or parts. Here, zero error means that the corresponding method does not detect a point (part) or that the corresponding database does not label a landmark (part). From Fig. \ref{fig:Fig 11} and Fig. \ref{fig:Fig 12}, it can be seen that landmarks around the outline of the face are the most difficult to accurately detect by all the tested methods. This is because the outline is easily affected by pose variation and occlusion. In contrast, the inner/outer corners of the eyes and the nose tips are relatively easy to localize, since these points are hardly affected by facial expressions, while the points around the mouth are heavily dependent on facial expressions.

Some methods are reported to have similar performance to human beings \citep{PAMI2013Belhumeur, ICCV2013Burgos}. However, occlusion and large shape variations in face images still provide significant challenges to successful and accurate detection, which is why in our experiments the above methods achieve better performance on the LFPW and Helen databases than on the COFW database.

It is also important to consider whether FFPD methods can detect facial landmarks in real-time. Model training is usually time-consuming in deep learning-based methods. The C++ implementation of CNN \citep{CVPR2013Sun} took 0.12s to process a single image on a 3.30 GHz CPU, excluding face detection and image resizing. CLM-based methods generally take training time to learn local experts (e.g., learning weights using a linear SVM). State-of-the-art CLM methods \citep{IJCV2011Saragih, CVPR2008Wang, ECCV2008Gu} are reported to take 0.120s, 0.098s, and 2.410s, respectively, on a 2.5 GHz Intel Core 2 Duo processor. Since publication of the seminal work in this area \citep{PBIJCV2004Matthews}, inverse composition fitting has significantly developed \citep{ICCV2013Tzimiropoulos}, and this state-of-the-art fitting algorithm reaches near real-time performance on real-world databases. Recently, cascaded regression methods have attracted a lot of attention, not only due to their favorable performance, but also because of their training and detection speed. Cao et al. \citep{CVPR2012Cao} reported that their method took only 20 minutes to train a model of 2000 images, with testing taking 0.015s with C++ implementation on an Intel Core i7 2.93 GHz CPU. RCPR \citep{ICCV2013Burgos} has even better performance than \citet{CVPR2012Cao}, which is also a cascaded fern-based method.

\section{Conclusion}
\label{sec:7}
Most existing methods improve the robustness dependent on carefully designed features such as pixel difference features \citep{CVPR2012Cao, ICCV2013Burgos} and SIFT features \citep{CVPR2013Xiong}. Though these features achieve some success, they still cannot adaptively deal with various shape variations and appearance variations. Recently, \citet{CVPR2014Ren} presented an effective way to learn a set of local binary features to represent the facial image. Another promising way to adaptively learn features is by virtue of deep learning \citep{PAMI2013Bengio} which achieves state-of-the-art performance on many computer vision tasks.

Besides feature learning, the model structure is another important issue related to the detection performance. Conventional ASM and AAM based methods assume that shape variations are statistically distribute as multivariate Gaussian, i.e. the linear PCA shape model. These explicit shape constraints actually have limited shape representation ability. Recent studies show that a cascaded set of simple linear regressors could achieve promising performance \citep{CVPR2013Xiong, CVPR2014Ren}. Implicit shape constraint would be automatically hold if the initial shape is a legal face shape \citep{CVPR2012Cao}.

In this paper we reviewed FFPD methods, which can be grouped into four major categories: constrained local model-based, active appearance model-based, regression-based, and \textit{other} methods. \textit{Other} methods could be further divided into four minor categories: graphical model-based methods, joint face alignment methods, independent FFP detectors, and deep learning-based methods. By virtue of a comprehensive analysis and comparison of these methods, we found that cascaded regression-based methods achieved promising performance in the experimental setting. Although some state-of-the-art methods are ostensibly comparable to humans on some databases, there remain challenges in detecting occluded faces or those with large shape variation. Furthermore, most existing real-world databases are composed of frontal or near frontal images. Automatic FFP detection remains a distant promise.

%
%


\bibliographystyle{spbasic}
\bibliography{FFPDS}   

\begin{thebibliography}{202}
\providecommand{\natexlab}[1]{#1}
\providecommand{\url}[1]{{#1}}
\providecommand{\urlprefix}{URL }
\expandafter\ifx\csname urlstyle\endcsname\relax
  \providecommand{\doi}[1]{DOI~\discretionary{}{}{}#1}\else
  \providecommand{\doi}{DOI~\discretionary{}{}{}\begingroup
  \urlstyle{rm}\Url}\fi
\providecommand{\eprint}[2][]{\url{#2}}

\bibitem[{Aizenberg et~al(2000)Aizenberg, Aizenberg, and
  Vandewalle}]{RPBook2000Aizenberg}
Aizenberg I, Aizenberg N, Vandewalle J (2000) Multi-valued and universal binary
  neurons: theory, learning, applications. Kluwer Academic

\bibitem[{Amberg and Better(2011)}]{ICCV2011Amberg}
Amberg B, Better T (2011) Optimal landmark detection using shape models and
  branch and bound. In: Proceedings of IEEE International Conference on
  Computer Vision, pp 455--462

\bibitem[{Amberg et~al(2009)Amberg, Andrew, and Thomas}]{CVPR2009Brian}
Amberg B, Andrew B, Thomas V (2009) On compositional image alignment, with an
  application to active appearance models. In: Proceedings of IEEE
  International Conference on Computer Vision and Pattern Recognition, pp
  1714--1721

\bibitem[{Anderson et~al(2013)Anderson, Stenger, Cipolla, and
  Wan}]{CVPR2013Anderson}
Anderson R, Stenger B, Cipolla R, Wan V (2013) Expressive visual text-to-speech
  using active appearance models. In: Proceedings of IEEE Conference on
  Computer Vision and Pattern Recognition, pp 3382--3389

\bibitem[{Arya et~al(1998)Arya, Mount, Silverman, and Wu}]{RPJACM1998Arya}
Arya S, Mount D, Silverman R, Wu A (1998) An optimal algorithm for approximate
  nearest neighbor searching. Journal of ACM 45(6):891--923

\bibitem[{Ashraf et~al(2010)Ashraf, Lucey, and Chen}]{CVPR2010Ashraf}
Ashraf A, Lucey S, Chen T (2010) Fast image alignment in the {Fourier} domain.
  In: Proceedings of IEEE Conference on Computer Vision and Pattern
  Recognition, pp 2480--2487

\bibitem[{Asteriadis et~al(2009)Asteriadis, Nikolaidis, and
  Pitas}]{PR2009Asteriadis}
Asteriadis S, Nikolaidis N, Pitas I (2009) Facial feature detection using
  distance vector fields. Pattern Recognition 42(7):1388--1398

\bibitem[{Asthana et~al(2009)Asthana, Goecke, Quadrianto, and
  Gedeon}]{CVPR2009Asthana}
Asthana A, Goecke R, Quadrianto N, Gedeon T (2009) Learning based automatic
  face annotation for arbitrary poses and expression from frontal images only.
  In: Proceedings of IEEE International Conference on Computer Vision and
  Pattern Recognition, pp 1635--1642

\bibitem[{Asthana et~al(2011)Asthana, Lucey, and Goecke}]{PR2011Asthana}
Asthana A, Lucey S, Goecke R (2011) Regression based automatic face annotation
  for deformable model building. Pattern Recognition 44(10-11):2598--2613

\bibitem[{Asthana et~al(2013)Asthana, Cheng, Zafeiriou, and
  Pantic}]{CVPR2013Asthana}
Asthana A, Cheng S, Zafeiriou S, Pantic M (2013) Robust discriminative response
  map fitting with constrained local models. In: Proceedings of IEEE Conference
  on Computer Vision and Pattern Recognition, pp 3444--3451

\bibitem[{Baker and Matthews(2004)}]{PBIJCV2004Baker}
Baker S, Matthews I (2004) Lucas-kanade 20 years on: A unifying framework.
  International Journal of Computer Vision 56(1):221--255

\bibitem[{Baker et~al(2003)Baker, Gross, and Matthews}]{PBLK20Part3}
Baker S, Gross R, Matthews I (2003) Lucas-kanade 20 years on: a unifying
  framework: part 3. Tech. rep., Carnegie Mellon University

\bibitem[{Baltrusaitis et~al(2012)Baltrusaitis, Robinson, and
  Morency}]{CVPR2012Baltrusaitis}
Baltrusaitis T, Robinson P, Morency L (2012) 3{D} constrained local model for
  rigid and non-rigid facial tracking. In: Proceedings of IEEE Conference on
  Computer Vision and Pattern Recognition, pp 2610--2617

\bibitem[{Batur and Hayes(2003)}]{PBCVPR2003Batur}
Batur A, Hayes M (2003) A novel convergence scheme for active appearance
  models. In: Proceedings of IEEE Conference on Computer Vision and Pattern
  Recogntion, pp 359--368

\bibitem[{Batur and Hayes(2005)}]{PBTIP2005Batur}
Batur A, Hayes M (2005) Adaptive active appearance models. IEEE Transactions on
  Image Processing 14(11):1707--1721

\bibitem[{Belhumeur et~al(2011)Belhumeur, Jacobs, Kriegman, and
  Kumar}]{CVPR2011Belhumeur}
Belhumeur P, Jacobs D, Kriegman D, Kumar N (2011) Localizing parts of faces
  using a consensus of exemplars. In: Proceedings of IEEE Conference on
  Computer Vision and Pattern Recognition, pp 545--552

\bibitem[{Belhumeur et~al(2013)Belhumeur, Jacobs, Kriegman, and
  Kumar}]{PAMI2013Belhumeur}
Belhumeur P, Jacobs D, Kriegman D, Kumar N (2013) Localizing parts of faces
  using a consensus of exemplars. IEEE Transactions on Pattern Analysis and
  Machine Intelligence 35(12):2930--2940

\bibitem[{Bengio et~al(2013)Bengio, Courville, and Vincent}]{PAMI2013Bengio}
Bengio Y, Courville A, Vincent P (2013) Representation learning: A review and
  new perspectives. IEEE Transactions on Pattern Analysis and Machine
  Intelligence 35(8):1798--1828

\bibitem[{Bitouk et~al(2008)Bitouk, Kumar, Dhillon, Belhumeur, and
  Nayar}]{ASIGGRAPH2008Bitouk}
Bitouk D, Kumar N, Dhillon S, Belhumeur P, Nayar S (2008) Face swapping:
  automatically replacing faces in photographs. In: Proceedings of SIGGRAPH, pp
  39.1--39.8

\bibitem[{Blanz and Vetter(1999)}]{RPSIGGRAPH1999Blanz}
Blanz V, Vetter T (1999) A morphable model for the synthesis of 3d faces. In:
  Proceedings of SIGGRAPH, pp 187--194

\bibitem[{Breiman(1984)}]{RPBook1984Breiman}
Breiman L (1984) Classification and regression trees. Boca Raton: Chapman $\&$
  Hall/CRC

\bibitem[{Breiman(2001)}]{RPML2001Breiman}
Breiman L (2001) Random forests. Machine Learning 45(1):5--32

\bibitem[{Burgos-Artizzu et~al(2013)Burgos-Artizzu, Perona, and
  Dollar}]{ICCV2013Burgos}
Burgos-Artizzu X, Perona P, Dollar P (2013) Robust face landmark estimation
  under occlusion. In: Proceedings of IEEE International Conference on Computer
  Vision, pp 1513--1520

\bibitem[{Butakoff and Frangi(2006)}]{PAMI2006Butakoff}
Butakoff C, Frangi A (2006) A framework for weighted fusion of multiple
  statistical models of shape and appearance. IEEE Transactions on Pattern
  Analysis and Machine Intelligence 28(11):1847--1857

\bibitem[{Butakoff and Frangi(2010)}]{CVIU2010Butakoff}
Butakoff C, Frangi A (2010) Multi-view face segmentation using fusion of
  statistical shape and appearance models. Computer Vision and Image
  Understanding 114(3):311--321

\bibitem[{Cao et~al(2012)Cao, Wei, Wen, and Sun}]{CVPR2012Cao}
Cao X, Wei Y, Wen F, Sun J (2012) Face alignment by explicit shape regression.
  In: Proceedings of IEEE Conference on Computer Vision and Pattern
  Recognition, pp 2887--2894

\bibitem[{Chen et~al(2001)Chen, Xu, Shum, Zhu, and Zheng}]{AICCV2001Chen}
Chen H, Xu Y, Shum H, Zhu S, Zheng N (2001) Example-based facial sketch
  generation with non-parametric sampling. In: Proceedings of IEEE
  International Conference on Computer Vision, pp 433--438

\bibitem[{Chew et~al(2011)Chew, Lucey, Lucey, Saragih, Cohn, and
  Sridharan}]{FG2011Chew}
Chew S, Lucey P, Lucey S, Saragih J, Cohn J, Sridharan S (2011)
  Person-independent facial expression detection using constrained local
  models. In: Proceedings of Inteernational Conference on Automatic Face and
  Gesture Recognition, pp 915--920

\bibitem[{Chow and Liu(1968)}]{RPTIT1968Chow}
Chow C, Liu C (1968) Approximating discrete probability distributions with
  dependence trees. IEEE Transactions on Information Theory 14(3):462--467

\bibitem[{Cootes and Taylor(1992)}]{PBBMVC1992Cootes}
Cootes T, Taylor C (1992) Active shape models-'smart snakes'. In: Proceedings
  of British Machine Vision Conference, pp 266--275

\bibitem[{Cootes and Taylor(1993)}]{PBBMVC1993Cootes}
Cootes T, Taylor C (1993) Active shape model search using local grey-level
  models: a quantitative evaluation. In: Proceedings of British Machine Vision
  Conference, pp 639--648

\bibitem[{Cootes and Taylor(1999)}]{PBIVC1999Cootes}
Cootes T, Taylor C (1999) A mixture model for representing shape variation.
  Image and Vision Computing 17(8):567--573

\bibitem[{Cootes and Taylor(2001)}]{PBICCV2001Cootes}
Cootes T, Taylor C (2001) Constrained active appearance models. In: Proceedings
  of IEEE International Conference on Computer Vision, pp 748--754

\bibitem[{Cootes and Taylor(2006)}]{BMVC2006Cootes}
Cootes T, Taylor C (2006) An algorithm for tuning an active appearance model to
  new data. In: Proceedings of British Machine Vision Conference, pp 919--928

\bibitem[{Cootes et~al(1994)Cootes, Taylor, and Lanitis}]{PBBMVC1994Cootes}
Cootes T, Taylor C, Lanitis A (1994) Active shape models: evaluation of a
  multi-resolution method for improving image search. In: Proceedings of
  British Machine Vision Conference, pp 327--336

\bibitem[{Cootes et~al(1995)Cootes, Taylor, Cooper, and
  Graham}]{PBCVIU1995Cootes}
Cootes T, Taylor C, Cooper D, Graham J (1995) Active shape models-their
  training and application. Computer Vision and Image Understanding
  61(1):38--59

\bibitem[{Cootes et~al(1998{\natexlab{a}})Cootes, Edwards, and
  Taylor}]{PBECCV1998Cootes}
Cootes T, Edwards G, Taylor C (1998{\natexlab{a}}) Active appearance models.
  In: Proceedings of European Conference on Computer Vision, pp 484--498

\bibitem[{Cootes et~al(1998{\natexlab{b}})Cootes, Edwards, and
  Taylor}]{PBBMVC1998Cootes}
Cootes T, Edwards G, Taylor C (1998{\natexlab{b}}) A comparative evaluation of
  active appearance model algorithms. In: Proceedings of British Machine Vision
  Conference, pp 680--689

\bibitem[{Cootes et~al(2001)Cootes, Edwards, and Taylor}]{PBPAMI2001Cootes}
Cootes T, Edwards G, Taylor C (2001) Active appearance models. IEEE
  Transactions on Pattern Analysis and Machine Intelligence 23(6):681--685

\bibitem[{Cootes et~al(2002)Cootes, Wheeler, Walker, and
  Taylor}]{PBIVC2002Cootes}
Cootes T, Wheeler G, Walker K, Taylor C (2002) View-based active appearance
  models. Image and Vision Computing 20(9-10):657--664

\bibitem[{Cootes et~al(2012)Cootes, Lonita, Lindner, and
  Sauer}]{ECCV2012Cootes}
Cootes T, Lonita M, Lindner C, Sauer P (2012) Robust and accurate shape model
  fitting using random forest regression voting. In: Proceedings of European
  Conference on Computer Vision, pp 278--291

\bibitem[{Coughlan and Ferreira(2002)}]{PBECCV2002Coughlan}
Coughlan J, Ferreira S (2002) Finding deformable shapes using loopy belief
  propagation. In: Proceedings of European Conference on Computer Vision, pp
  453--468

\bibitem[{Criminisi et~al(2012)Criminisi, Shotton, and
  Konukoglu}]{RPFTCGV2012Criminisi}
Criminisi A, Shotton J, Konukoglu E (2012) Decision forests: a unified
  framework for classification, regression, density estimation, manifold
  learning and semi-supervised learning. Foundations and Trends in Computer
  Graphics and Vision 7(2-3):81--227

\bibitem[{Cristinacce and Cootes(2003)}]{PBBMVC2003Cristinacce}
Cristinacce D, Cootes T (2003) Facial feature detection using {AdaBoost} with
  shape constraints. In: Proceedings of British Machine Vision Conference, pp
  24.1--24.10

\bibitem[{Cristinacce and Cootes(2004)}]{PBFG2004Cristinacce}
Cristinacce D, Cootes T (2004) A comparison of shape constrained facial feature
  detectors. In: Proceedings of Inteernational Conference on Automatic Face and
  Gesture Recognition, pp 375--380

\bibitem[{Cristinacce and Cootes(2006{\natexlab{a}})}]{FG2006Cristinacce}
Cristinacce D, Cootes T (2006{\natexlab{a}}) Facial feature detection and
  tracking with automatic template selection. In: Proceedings of Inteernational
  Conference on Automatic Face and Gesture Recognition, pp 429--434

\bibitem[{Cristinacce and Cootes(2006{\natexlab{b}})}]{BMVC2006Cristinacce}
Cristinacce D, Cootes T (2006{\natexlab{b}}) Feature detection and tracking
  with constrained local models. In: Proceedings of British Machine Vision
  Conference, pp 929--938

\bibitem[{Cristinacce and Cootes(2007)}]{BMVC2007Cristinacce}
Cristinacce D, Cootes T (2007) Boosted regression active shape models. In:
  Proceedings of British Machine Vision Conference, pp 1--10

\bibitem[{Cristinacce and Cootes(2008)}]{PR2008Cristinacce}
Cristinacce D, Cootes T (2008) Automatic feature localization with constrained
  local models. Pattern Recognition 41(10):3054--3067

\bibitem[{Cristinacce et~al(2004)Cristinacce, Cootes, and
  Scott}]{PBBMVC2004Cristinacce}
Cristinacce D, Cootes T, Scott I (2004) A multi-stage approach to facial
  feature detection. In: Proceedings of British Machine Vision Conference, pp
  231--240

\bibitem[{Dalal and Triggs(2005)}]{RPCVPR2005Dalal}
Dalal N, Triggs B (2005) Histograms of oriented gradients for human detection.
  In: Proceedings of IEEE Conference on Computer Vision and Pattern
  Recognition, pp 886--893

\bibitem[{Dantone et~al(2012)Dantone, Gall, Fanelli, and van
  Gool}]{CVPR2012Dantone}
Dantone M, Gall J, Fanelli G, van Gool L (2012) Real-time facial feature
  detection using conditional regression forests. In: Proceedings of IEEE
  Conference on Computer Vision and Pattern Recognition, pp 2578--2585

\bibitem[{Dedeoglu et~al(2006)Dedeoglu, Baker, and Kanade}]{ECCV2006Dedeoglu}
Dedeoglu G, Baker S, Kanade T (2006) Resolution-aware fitting of active
  appearance models to low resolution images. In: Proceedings of European
  Conference on Computer Vision, pp 83--97

\bibitem[{Dedeoglu et~al(2007)Dedeoglu, Kanade, and Baker}]{PAMI2007Dedeoglu}
Dedeoglu G, Kanade T, Baker S (2007) The asymmetry of image registration and
  its application to face tracking. IEEE Transactions on Pattern Analysis and
  Machine Intelligence 29(5):807--823

\bibitem[{Ding and Martinez(2008)}]{CVPR2008Ding}
Ding L, Martinez A (2008) Precise detailed detection of faces and facial
  features. In: Proceedings of IEEE International Conference on Computer Vision
  and Pattern Recognition, pp 1--7

\bibitem[{Ding and Martinez(2010)}]{PAMI2010Ding}
Ding L, Martinez A (2010) Features versus context: an approach for precise and
  detailed detection and delineation of faces and facial features. IEEE
  Transactions on Pattern Analysis and Machine Intelligence 32(11):2022--2038

\bibitem[{Dollar et~al(2010)Dollar, Welinder, and Perona}]{RPCVPR2010Dollar}
Dollar P, Welinder P, Perona P (2010) Cascaded pose regression. In: Proceedings
  of IEEE Conference on Computer Vision and Pattern Recognition, pp 1078--1085

\bibitem[{Donner et~al(2006)Donner, Reiter, Georg, Peloschek, and
  Bischof}]{PAMI2006Donner}
Donner R, Reiter M, Georg L, Peloschek P, Bischof H (2006) Fast active
  appearance model search using canonical correlation analysis. IEEE
  Transactions on Pattern Analysis and Machine Intelligence 28(10):1690--1694

\bibitem[{Duffy(2002)}]{RPML2002Duffy}
Duffy N (2002) Boosting methods for regression. Machine Learning
  47(2-3):153--200

\bibitem[{Everingham et~al(2006)Everingham, Sivic, and
  Zisserman}]{BMVC2006Everingham}
Everingham M, Sivic J, Zisserman A (2006) "hello! {My} name is ...
  {Buffy}"-automatic naming of characters in tv video. In: Proceedings of
  British Machine Vision Conference, pp 899--908

\bibitem[{Fanelli et~al(2013)Fanelli, Dantone, and Gool}]{FG2013Fanelli}
Fanelli G, Dantone M, Gool L (2013) Real time {3D} face alignment with random
  forests-based active appearance models. In: Proceedings of Inteernational
  Conference on Automatic Face and Gesture Recognition, pp 1--8

\bibitem[{Felzenszwalb and Huttenlocher(2005)}]{RPIJCV2005Felzenszwalb}
Felzenszwalb P, Huttenlocher D (2005) Pictorial structures for object
  recognition. International Journal of Computer Vision 61(1):55--79

\bibitem[{Fischler and Bolles(1981)}]{RPCACM1981Fischler}
Fischler M, Bolles R (1981) Random sample consensus: a paradigm for model
  fitting with application to image analysis and automated cartography.
  Communications of the ACM 24(6):381--395

\bibitem[{Freund and Schapire(1997)}]{RPJCSS1997Freund}
Freund Y, Schapire R (1997) A decision-theoretic generalization of on-line
  learning and an application to boosting. Journal of Computer and System
  Science 55:119--139

\bibitem[{Freund et~al(2003)Freund, Iyer, Schapire, and
  Singer}]{RPJMLR2003Freund}
Freund Y, Iyer R, Schapire R, Singer Y (2003) An efficient boosting algorithm
  for combining preferences. Journal of Machine Learning Research 4(6):933--969

\bibitem[{Friedman(2001)}]{AS2001Friedman}
Friedman J (2001) Greedy function approximation: a gradient boosting machine.
  The Annals of Statistics 29(5):1189--1232

\bibitem[{Friedman et~al(2000)Friedman, Hastie, and
  Tibshiani}]{RPAS2000Friedman}
Friedman J, Hastie T, Tibshiani R (2000) Additive logistic regression: a
  statistical view of boosting. The Annals of Statistics 38(2):337--374

\bibitem[{Gao et~al(2011)Gao, Ekenel, Fischer, and Stiefelhagen}]{BMVC2011Gao}
Gao H, Ekenel H, Fischer M, Stiefelhagen R (2011) Boosting pseudo census
  transform feature for face alignment. In: Proceedings of British Machine
  Vision Conference, pp 54.1--54.11

\bibitem[{Gao et~al(2012)Gao, Ekenel, and Stiefelhagen}]{BMVC2012Gao}
Gao H, Ekenel H, Stiefelhagen R (2012) Face alignment using a ranking model
  based on regression trees. In: Proceedings of British Machine Vision
  Conference, pp 118.1--118.11

\bibitem[{Gao et~al(2010)Gao, Su, Li, and Tao}]{SMCC2010Gao}
Gao X, Su Y, Li X, Tao D (2010) A review of active appearance models. IEEE
  Transactions on Systems, Man, and Cybernetics-Part C: Applications and
  Reviews 40(2):145--158

\bibitem[{van Ginneken et~al(2002)van Ginneken, Frangi, Staal, Romeny, and
  Viergever}]{PBTMI2002Ginneken}
van Ginneken B, Frangi A, Staal J, Romeny B, Viergever M (2002) Active shape
  model segmentation with optimal features. IEEE Transactions on Medical
  Imaging 21(8):924--933

\bibitem[{Gonzalez-Mora et~al(2007)Gonzalez-Mora, De~la Torre, Murthi, Guil,
  and Zapata}]{ICCV2007Gonzalez}
Gonzalez-Mora J, De~la Torre F, Murthi R, Guil N, Zapata E (2007) Bilinear
  acitve appearance models. In: Proceedings of IEEE International Conference on
  Computer Vision, pp 1--8

\bibitem[{Gross et~al(2005)Gross, Mattews, and Baker}]{PBIVC2005Gross}
Gross R, Mattews I, Baker S (2005) Generic vs. person specific active
  appearance models. Image and Vision Computing 23(12):1080--1093

\bibitem[{Gross et~al(2010)Gross, Matthews, Cohn, Kanade, and
  Baker}]{DB2010MultiPIE}
Gross R, Matthews I, Cohn J, Kanade T, Baker S (2010) Multi-pie. Image and
  Vision Computing 28(5):807--813

\bibitem[{Grujic et~al(2008)Grujic, Ilic, Lepetit, and Fua}]{RPTR2008Grujic}
Grujic N, Ilic S, Lepetit V, Fua P (2008) {3D} facial pose estimation by image
  retrieval. Tech. rep., Deutsche Telekom Laboratories

\bibitem[{Gu and Kanade(2006)}]{CVPR2006Gu}
Gu L, Kanade T (2006) 3{D} alignment of face in a single image. In: Proceedings
  of IEEE International Conference on Computer Vision and Pattern Recognition,
  pp 1305--1312

\bibitem[{Gu and Kanade(2008)}]{ECCV2008Gu}
Gu L, Kanade T (2008) A generative shape regularization model for robust face
  alignment. In: Proceedings of European Conference on Computer Vision, pp
  413--426

\bibitem[{Gu et~al(2007)Gu, Xing, and Kanade}]{CVPR2007Gu}
Gu L, Xing E, Kanade T (2007) Learning {GMRF} structures for spatial priors.
  In: Proceedings of IEEE International Conference on Computer Vision and
  Pattern Recognition, pp 1--6

\bibitem[{Hall et~al(2000)Hall, D., and Martin}]{RPPAMI2000Hall}
Hall P, D M, Martin R (2000) Merging and splitting eigenspace models. IEEE
  Transactions on Pattern Analysis and Machine Intelligence 22(9):1042--1049

\bibitem[{Hamsici and Martinez(2009)}]{ICCV2009Hamsici}
Hamsici O, Martinez A (2009) Active appearance models with rotation invariant
  kernels. In: Proceedings of IEEE International Conference on Computer Vision,
  pp 1003--1009

\bibitem[{Hansen et~al(2011)Hansen, Fagertun, and Larsen}]{BMVC2011Hansen}
Hansen M, Fagertun J, Larsen R (2011) Elastic appearance models. In:
  Proceedings of British Machine Vision Conference, pp 91.1--91.12

\bibitem[{Hinton and Salakhutdinov(2006)}]{RPScience2006Hinton}
Hinton G, Salakhutdinov R (2006) Reducing the dimensionality of data with
  neural networks. Science 313(5786):504--507

\bibitem[{Hinton et~al(2006)Hinton, Osindero, and Teh}]{RPNC2006Hinton}
Hinton G, Osindero S, Teh Y (2006) A fast learning algorithm for deep belief
  nets. Neural Computation 18(7):1527--1554

\bibitem[{Hou et~al(2001)Hou, Li, Zhang, and Cheng}]{PBCVPR2001Hou}
Hou X, Li S, Zhang H, Cheng Q (2001) Direct appearance models. In: Proceedings
  of IEEE Conference on Computer Vision and Pattern Recognition, pp 828--833

\bibitem[{Huang et~al(2012)Huang, Ding, and Fang}]{CVIU2012Huang}
Huang C, Ding X, Fang C (2012) Pose robust face tracking by combining
  view-based {AAMs} and temporal filters. Computer Vision and Image
  Understanding 116(7):777--792

\bibitem[{Huang et~al(2007{\natexlab{a}})Huang, Ramesh, Berg, and
  Learned-Miller}]{DB2007LFW}
Huang G, Ramesh M, Berg T, Learned-Miller E (2007{\natexlab{a}}) Labeled faces
  in the wild: A database for studying face recognition in unconstrained
  environments. Tech. Rep. 07-49, University of Massachusetts

\bibitem[{Huang et~al(2007{\natexlab{b}})Huang, Liu, and
  Metaxas}]{ICCV2007Huang}
Huang Y, Liu Q, Metaxas D (2007{\natexlab{b}}) A component based deformable
  model for generalized face alignment. In: Proceedings of IEEE International
  Conference on Computer Vision, pp 1--8

\bibitem[{Jesorsky et~al(2001)Jesorsky, Kirchberg, and
  Frischholz}]{DB2001BioID}
Jesorsky O, Kirchberg K, Frischholz R (2001) Robust face detection using the
  hausdoff distance. In: International Conference on Audio- and Video-based
  Biometric Person Authentication, pp 90--95

\bibitem[{Kahraman et~al(2007)Kahraman, Gokmen, Darkner, and
  Larsen}]{CVPR2007Kahraman}
Kahraman F, Gokmen M, Darkner S, Larsen R (2007) An active illumination and
  appearance {(AIA)} model for face alignment. In: Proceedings of IEEE
  International Conference on Computer Vision and Pattern Recognition, pp 1--7

\bibitem[{Kasinski et~al(2008)Kasinski, Florek, and Schmidt}]{DB2008PUT}
Kasinski A, Florek A, Schmidt A (2008) The {PUT} face database. Image
  Processing and Communications 13(3):59--64

\bibitem[{Kazemi and Cullivan(2011)}]{BMVC2011Kazemi}
Kazemi V, Cullivan J (2011) Face alignment with part-based modeling. In:
  Proceedings of British Machine Vision Conference, pp 27.1--27.10

\bibitem[{Kinoshita et~al(2012)Kinoshita, Konishi, Kawade, and
  Murase}]{ECCV2012Kinoshita}
Kinoshita K, Konishi Y, Kawade M, Murase H (2012) Facial model fitting based on
  perturbation learning and it's evaluation on challenging real-world
  diversities images. In: Proceedings of European Conference on Computer Vision
  Workshop, pp 153--162

\bibitem[{Kostinger et~al(2011)Kostinger, Wohlhart, Roth, and
  Bischof}]{DB2011AFLW}
Kostinger M, Wohlhart P, Roth P, Bischof H (2011) Annotated facial landmarks in
  the wild: a large-scale, real-world database for facial landmark
  localization. In: International Conference on Computer Vision WorkShops, pp
  2144--2151

\bibitem[{Kozakaya et~al(2008{\natexlab{a}})Kozakaya, Shibata, Takeguchi, and
  Nishiura}]{CVPR2008Kozakaya}
Kozakaya T, Shibata T, Takeguchi T, Nishiura M (2008{\natexlab{a}}) Fully
  automatic feature localization for medical images using a global vector
  concentration approach. In: Proceedings of IEEE International Conference on
  Computer Vision and Pattern Recognition, pp 1--6

\bibitem[{Kozakaya et~al(2008{\natexlab{b}})Kozakaya, Shibata, Yuasa, and
  Yamaguchi}]{FG2008Kozakaya}
Kozakaya T, Shibata T, Yuasa M, Yamaguchi O (2008{\natexlab{b}}) Facial feature
  localization using weighted vector concentration approach. In: Proceedings of
  Inteernational Conference on Automatic Face and Gesture Recognition, pp 1--6

\bibitem[{Kozakaya et~al(2010)Kozakaya, Shibata, Yuasa, and
  Yamaguchi}]{IVC2010Kozakaya}
Kozakaya T, Shibata T, Yuasa M, Yamaguchi O (2010) Facial feature localization
  using weighted vector concentration approach. Image and Vision Computing
  28(5):772--780

\bibitem[{La~Cascia et~al(2000)La~Cascia, Sclaroff, and
  Athitsoss}]{RPPAMI2000Cascia}
La~Cascia M, Sclaroff S, Athitsoss V (2000) Fast, reliable head tracking under
  varying illumination: an approach based on registration of texture-mapped 3d
  models. IEEE Transactions on Pattern Analysis and Machine Intelligence
  22(4):322--336

\bibitem[{Le et~al(2012)Le, Brandt, Lin, Bourdev, and Huang}]{ECCV2012Le}
Le V, Brandt J, Lin Z, Bourdev L, Huang T (2012) Interactive facial feature
  localization. In: Proceedings of European Conference on Computer Vision, pp
  679--692

\bibitem[{Learned-Miller(2006)}]{RPPAMI2006Miller}
Learned-Miller E (2006) Data driven image models through continuous joint
  alignment. IEEE Transactions on Pattern Analysis and Machine Intelligence
  28(2):236--250

\bibitem[{Lee and Kim(2009)}]{PAMI2009Lee}
Lee H, Kim D (2009) Tensor-based {AAM} with continuous variation estimation:
  application to variation-robust face recognition. IEEE Transactions on
  Pattern Analysis and Machine Intelligence 31(6):1102--1116

\bibitem[{Li et~al(2009)Li, Gu, and Kanade}]{CVPR2009Li}
Li Y, Gu L, Kanade T (2009) A robust shape model for multi-view car alignment.
  In: Proceedings of IEEE International Conference on Computer Vision and
  Pattern Recognition, pp 2466--2473

\bibitem[{Li et~al(2011)Li, Gu, and Kanade}]{PAMI2011Li}
Li Y, Gu L, Kanade T (2011) Robustly aligning a shape model and its application
  to car alignment of unknown pose. IEEE Transactions on Pattern Analysis and
  Machine Intelligence 33(9):1860--1876

\bibitem[{Liang et~al(2006{\natexlab{a}})Liang, F., Xu, Tang, and
  Shum}]{CVPR2006Liang}
Liang L, F W, Xu Y, Tang X, Shum H (2006{\natexlab{a}}) Accurate face alignment
  using shape constrained {Markov} network. In: Proceedings of IEEE
  International Conference on Computer Vision and Pattern Recognition, pp
  1313--1319

\bibitem[{Liang et~al(2006{\natexlab{b}})Liang, Wen, Tang, and
  Xu}]{ECCV2006Liang}
Liang L, Wen F, Tang X, Xu Y (2006{\natexlab{b}}) An integrated model for
  accurate shape alignment. In: Proceedings of European Conference on Computer
  Vision, pp 333--346

\bibitem[{Liang et~al(2008)Liang, Xiao, Wen, and Sun}]{ECCV2008Liang}
Liang L, Xiao R, Wen F, Sun J (2008) Face alignment via component-based
  discriminative search. In: Proceedings of European Conference on Computer
  Vision, pp 72--85

\bibitem[{Liu(2007)}]{CVPR2007Liu}
Liu X (2007) Generic face alignment using boosted appearance model. In:
  Proceedings of IEEE International Conference on Computer Vision and Pattern
  Recognition, pp 1--8

\bibitem[{Liu(2009)}]{PAMI2009Liu}
Liu X (2009) Discriminative face alignment. IEEE Transactions on Pattern
  Analysis and Machine Intelligence 31(11):1941--1954

\bibitem[{Liu et~al(2006)Liu, Tu, and Wheeler}]{BMVC2006Liu}
Liu X, Tu P, Wheeler F (2006) Face model fitting on low resolution images. In:
  Proceedings of British Machine Vision Conference, pp 1079--1088

\bibitem[{Lowe(2004)}]{RPIJCV2004Lowe}
Lowe D (2004) Distinctive image features from scale-invariant keypoints.
  International Journal of Computer Vision 60(2):91--110

\bibitem[{Lucey et~al(2009)Lucey, Wang, Cox, Sridharan, and
  Cohn}]{IVC2009Lucey}
Lucey S, Wang Y, Cox M, Sridharan S, Cohn J (2009) Efficient constrained local
  model fitting for non-rigid face alignment. Image and Vision Computing
  27(12):1804--1813

\bibitem[{Lucey et~al(2013)Lucey, Navarathna, Ashraf, and
  Sridharan}]{PAMI2013Lucey}
Lucey S, Navarathna R, Ashraf A, Sridharan S (2013) Fourier {Lucas-Kanade}
  algorithm. IEEE Transactions on Pattern Analysis and Machine Intelligence
  35(6):1383--1396

\bibitem[{Luo et~al(2012)Luo, Wang, and Tang}]{CVPR2012Luo}
Luo P, Wang X, Tang X (2012) Hierarchical face parsing via deep learning. In:
  Proceedings of IEEE Conference on Computer Vision and Pattern Recognition, pp
  2480--2487

\bibitem[{Martinez and Benavente(1998)}]{DB1998AR}
Martinez A, Benavente R (1998) The {AR} face database. Tech. rep., University
  of Barcelona

\bibitem[{Martinez et~al(2013)Martinez, Valstar, Binefa, and
  Pantic}]{PAMI2013Martinez}
Martinez B, Valstar M, Binefa X, Pantic M (2013) Local evidence aggregation for
  regression-based facial point detection. IEEE Transactions on Pattern
  Analysis and Machine Intelligence 35(5):1149--1163

\bibitem[{Martins et~al(2010)Martins, Caseiro, and Batista}]{BMVC2010Martins}
Martins P, Caseiro R, Batista J (2010) Face alignment through {2.5D} active
  appearance models. In: Proceedings of British Machine Vision Conference, pp
  1--12

\bibitem[{Martins et~al(2012{\natexlab{a}})Martins, Caseiro, Henriques, and
  Batista}]{ECCV2012Martins}
Martins P, Caseiro R, Henriques J, Batista J (2012{\natexlab{a}})
  Discriminative {Bayesian} active shape models. In: Proceedings of European
  Conference on Computer Vision, pp 57--70

\bibitem[{Martins et~al(2012{\natexlab{b}})Martins, Caseiro, Henriques, and
  Batista}]{BMVC2012Martins}
Martins P, Caseiro R, Henriques J, Batista J (2012{\natexlab{b}}) Let the shape
  speak-discriminative face alignment using conjugate priors. In: Proceedings
  of British Machine Vision Conference, pp 118.1--118.11

\bibitem[{Martins et~al(2013)Martins, Caseiro, and Batista}]{CVIU2013Martins}
Martins P, Caseiro R, Batista J (2013) Generative face alignment through {2.5D}
  active appearance models. Computer Vision and Image Understanding
  117(3):250--268

\bibitem[{Matthews and Baker(2004)}]{PBIJCV2004Matthews}
Matthews I, Baker S (2004) Active appearance models revisited. International
  Journal of Computer Vision 60(2):135--164

\bibitem[{Matthews et~al(2007)Matthews, Xiao, and Baker}]{IJCV2007Matthews}
Matthews I, Xiao J, Baker S (2007) {2D} vs. {3D} deformable face models:
  representational power, construction, and real-time fitting. International
  Journal of Computer Vision 75(1):93--113

\bibitem[{Mei et~al(2008)Mei, Figl, Darzi, Rueckert, and Edwards}]{ECCV2008Mei}
Mei L, Figl M, Darzi A, Rueckert D, Edwards P (2008) Sample sufficiency and
  {PCA} dimension for statistical shape models. In: Proceedings of European
  Conference on Computer Vision, pp 492--503

\bibitem[{Messer et~al(1999)Messer, Matas, Kittler, Luettin, and
  Maitre}]{DB1999XM2VTS}
Messer K, Matas J, Kittler J, Luettin J, Maitre G (1999) {XM2VTSDB}:the
  extended {M2VTS} database. In: International Conference on Audio- and
  Video-based Biometric Person Authentication, pp 72--77

\bibitem[{Miborrow and F.(2008)}]{ECCV2008Milborrow}
Miborrow S, F N (2008) Locating facial features with an extended active shape
  model. In: Proceedings of European Conference on Computer Vision, pp 504--513

\bibitem[{Miborrow et~al(2010)Miborrow, Morkel, and Nicolls}]{DB2010MUCT}
Miborrow S, Morkel J, Nicolls F (2010) The {MUCT} landmarked face database. In:
  Proceedings of Pattern Recognition Association of South Africa, pp 1--6

\bibitem[{Michal et~al(2012)Michal, Franc, and
  Hlav{\'{a}{\v{c}}}}]{VISAPP2012Michal}
Michal U, Franc V, Hlav{\'{a}{\v{c}}} V (2012) Detector of facial landmarks
  learned by the structured output {SVM}. In: Proceedings of International
  Conference on Computer Vision Theory and Applications, pp 547--556

\bibitem[{Navarathna et~al(2011)Navarathna, Sridharan, and
  Lucey}]{ICCV2011Navarathna}
Navarathna R, Sridharan S, Lucey S (2011) Fourier active appearance models. In:
  Proceedings of IEEE International Conference on Computer Vision, pp
  1919--1926

\bibitem[{Nelder and Mead(1965)}]{RPComputerJournal1965}
Nelder J, Mead R (1965) A simplex method for function minimization. Computer
  Journals 7(4):308--313

\bibitem[{Nguyen and Torre(2008)}]{FG2008Nguyen}
Nguyen M, Torre F (2008) Learning image alignment without local minima for face
  detection and tracking. In: Proceedings of Inteernational Conference on
  Automatic Face and Gesture Recognition, pp 1--7

\bibitem[{Nguyen and De~la Torre(2008)}]{CVPR2008Nguyen}
Nguyen M, De~la Torre F (2008) Local minima free parameterized appearance
  models. In: Proceedings of IEEE International Conference on Computer Vision
  and Pattern Recognition, pp 1--8

\bibitem[{Nguyen and Torre(2010)}]{IJCV2010Nguyen}
Nguyen M, Torre F (2010) Metric learning for image alignment. International
  Journal of Computer Vision 88(1):69--84

\bibitem[{Nordstrom et~al(2004)Nordstrom, Larsen, Sierakowski, and
  Stegmann}]{DB2004IMM}
Nordstrom M, Larsen M, Sierakowski J, Stegmann M (2004) The {IMM} face
  database-an annotated dataset of 240 face images. Tech. rep., Technical
  University of Denmark

\bibitem[{Ojala et~al(1996)Ojala, Pietikainen, and Harwood}]{RPPR1996Ojala}
Ojala T, Pietikainen M, Harwood D (1996) A comparative study of texute measures
  with classification based on featured distributions. Pattern Recognition
  29(1):51--59

\bibitem[{Ozuysal et~al(2010)Ozuysal, Calonder, Lepetit, and
  Fua}]{RPPAMI2010Ozuysal}
Ozuysal M, Calonder M, Lepetit V, Fua P (2010) Fast keypoint recognition using
  random ferns. IEEE Transactions on Pattern Analysis and Machine Intelligence
  32(3):448--461

\bibitem[{Papandreou and Maragos(2008)}]{CVPR2008Papandreou}
Papandreou G, Maragos P (2008) Adaptive and constrained algorithms for inverse
  compositional active appearance model fitting. In: Proceedings of IEEE
  International Conference on Computer Vision and Pattern Recognition, pp 1--8

\bibitem[{Paquet(2009)}]{CVPR2009Paquet}
Paquet U (2009) Convexity and {Bayesian} constrained local models. In:
  Proceedings of IEEE International Conference on Computer Vision and Pattern
  Recognition, pp 1193--1199

\bibitem[{Peng et~al(2012)Peng, Ganesh, Wright, Xu, and Ma}]{RPPAMI2012Peng}
Peng Y, Ganesh A, Wright J, Xu W, Ma Y (2012) {RASL}: robust alignment by
  sparse and low-rank decomposition for linearly correlated images. IEEE
  Transactions on Pattern Analysis and Machine Intelligence 34(11):2233--2246

\bibitem[{Peyras et~al(2007)Peyras, Bartoli, Mercier, and
  Dalle}]{BMVC2007Peyras}
Peyras J, Bartoli A, Mercier H, Dalle P (2007) Segmented {AAMs} improve
  person-independent face fitting. In: Proceedings of British Machine Vision
  Conference, pp 1--10

\bibitem[{Phillips et~al(2000)Phillips, Moon, Rauss, and Rizvi}]{DB2000FERET}
Phillips P, Moon H, Rauss P, Rizvi S (2000) The {FERET} evaluation methodology
  for face recognition algorithms. IEEE Transactions on Pattern Analysis and
  Machine Intelligence 22(10):1090--1104

\bibitem[{Ren et~al(in press, 2014)Ren, Cao, Wei, and Sun}]{CVPR2014Ren}
Ren S, Cao X, Wei Y, Sun J (in press, 2014) Face alignment at 3000 fps via
  regressing local binary features. In: Proceedings of IEEE Conference on
  Computer Vision and Pattern Recognition

\bibitem[{Rivera and Martinez(2012)}]{PR2012Rivera}
Rivera S, Martinez A (2012) Learning deformable shape manifolds. Pattern
  Recognition 45(4):1792--1801

\bibitem[{Roberts et~al(2007)Roberts, Cootes, and Adams}]{BMVC2007Roberts}
Roberts M, Cootes T, Adams J (2007) Robust active appearance models with
  iteratively rescaled kernels. In: Proceedings of British Machine Vision
  Conference, pp 17.1--17.10

\bibitem[{Roh et~al(2011)Roh, Oguri, and Kanade}]{FG2011Roh}
Roh M, Oguri T, Kanade T (2011) Face alignment robust to occlusion. In:
  Proceedings of Inteernational Conference on Automatic Face and Gesture
  Recognition, pp 239--244

\bibitem[{Sanchez-Lozano et~al(2012)Sanchez-Lozano, De~la Torre, and
  Gonzalez-Jimenez}]{ECCV2012Lozano}
Sanchez-Lozano E, De~la Torre F, Gonzalez-Jimenez D (2012) Continuous
  regression for non-rigid image alignment. In: Proceedings of European
  Conference on Computer Vision, pp 250--263

\bibitem[{Saragih(2011)}]{CVPR2011Saragih}
Saragih J (2011) Principal regression analysis. In: Proceedings of IEEE
  Conference on Computer Vision and Pattern Recognition, pp 2881--2888

\bibitem[{Saragih and Gocke(2009)}]{PR2009Saragih}
Saragih J, Gocke R (2009) Learning {AAM} fitting through simulation. Pattern
  Recognition 42(11):2628--2636

\bibitem[{Saragih and Goecke(2007)}]{ICCV2007Saragih}
Saragih J, Goecke R (2007) A nonlinear discriminative approach to {AAM}
  fitting. In: Proceedings of IEEE International Conference on Computer Vision,
  pp 1--8

\bibitem[{Saragih et~al(2008)Saragih, Lucey, and Cohn}]{FG2008Saragih}
Saragih J, Lucey S, Cohn J (2008) Deformable face fitting with soft
  correspondence constraints. In: Proceedings of Inteernational Conference on
  Automatic Face and Gesture Recognition, pp 1--8

\bibitem[{Saragih et~al(2009{\natexlab{a}})Saragih, Lucey, and
  Cohn}]{ICCV2009Saragih02}
Saragih J, Lucey S, Cohn J (2009{\natexlab{a}}) Deformable model fitting with a
  mixture of local experts. In: Proceedings of IEEE International Conference on
  Computer Vision, pp 2248--2255

\bibitem[{Saragih et~al(2009{\natexlab{b}})Saragih, Lucey, and
  Cohn}]{ICCV2009Saragih01}
Saragih J, Lucey S, Cohn J (2009{\natexlab{b}}) Face alignment through subspace
  constrained mean-shifts. In: Proceedings of IEEE International Conference on
  Computer Vision, pp 1034--1041

\bibitem[{Saragih et~al(2009{\natexlab{c}})Saragih, Lucey, and
  Cohn}]{ICCV2009Saragih03}
Saragih J, Lucey S, Cohn J (2009{\natexlab{c}}) Probabilistic constrained
  adaptive local displacement experts. In: Proceedings of IEEE International
  Conference on Computer Vision Workshops, pp 288--295

\bibitem[{Saragih et~al(2011)Saragih, Lucey, and Cohn}]{IJCV2011Saragih}
Saragih J, Lucey S, Cohn J (2011) Deformable model fitting by regularized
  landmark mean-shift. International Journal of Computer Vision 91(2):200--215

\bibitem[{Sauer et~al(2011)Sauer, Cootes, and Taylor}]{BMVC2011Sauer}
Sauer P, Cootes T, Taylor C (2011) Accurate regression procedures for active
  appearance models. In: Proceedings of British Machine Vision Conference, pp
  1--11

\bibitem[{Serre et~al(2007)Serre, Wolf, Bileschi, Riesenhuber, and
  Poggio}]{RPPAMI2007Serre}
Serre T, Wolf L, Bileschi S, Riesenhuber M, Poggio T (2007) Robust object
  recognition with cortex-like mechanisms. IEEE Transactions on Pattern
  Analysis and Machine Intelligence 29(3):411--426

\bibitem[{Shen et~al(2013)Shen, Lin, Brandt, and Wu}]{CVPR2013Shen}
Shen X, Lin Z, Brandt J, Wu Y (2013) Detecting and aligning faces by image
  retrieval. In: Proceedings of IEEE Conference on Computer Vision and Pattern
  Recognition, pp 3460--3467

\bibitem[{Sivic et~al(2009)Sivic, Everingham, and Zisserman}]{CVPR2009Sivic}
Sivic J, Everingham M, Zisserman A (2009) "who are you?"-learning person
  specific classifiers from video. In: Proceedings of IEEE International
  Conference on Computer Vision and Pattern Recognition, pp 1145--1152

\bibitem[{Smith and Zhang(2012)}]{JAECCV2012Smith}
Smith B, Zhang L (2012) Joint face alignment with non-parametric shape models.
  In: Proceedings of European Conference on Computer Vision, pp 43--56

\bibitem[{Smith et~al(2013)Smith, Zhang, Brandt, Lin, and Yang}]{CVPR2013Smith}
Smith B, Zhang L, Brandt J, Lin Z, Yang J (2013) Exemplar-based face parsing.
  In: Proceedings of IEEE Conference on Computer Vision and Pattern
  Recognition, pp 3484--3491

\bibitem[{Sozou et~al(1995)Sozou, Cootes, Taylor, and Mauro}]{PBIVC1995Sozou}
Sozou P, Cootes T, Taylor C, Mauro E (1995) A non-linear generalization of
  point distribution models using polynomial regression. Image and Vision
  Computing 13(5):451--457

\bibitem[{Sozou et~al(1997)Sozou, Cootes, Taylor, and Mauro}]{PBIVC1997Sozou}
Sozou P, Cootes T, Taylor C, Mauro E (1997) Non-linear point distribution
  modeling using a multi-layer perceptron. Image and Vision Computing
  15(6):457--463

\bibitem[{Stegmann et~al(2003)Stegmann, Ersboll, and
  Larsen}]{PBTMI2003Stegmann}
Stegmann M, Ersboll B, Larsen R (2003) {FAME}-a flexible appearance modeling
  environment. IEEE Transactions on Medical Imaging 22(10):1319--1331

\bibitem[{Sukno et~al(2007)Sukno, Ordas, Butakoff, Cruz, and
  Frangi}]{PAMI2007sukno}
Sukno F, Ordas S, Butakoff C, Cruz S, Frangi A (2007) Active shape models with
  invariant optimal features: application to facial analysis. IEEE Transactions
  on Pattern Analysis and Machine Intelligence 29(7):1105--1117

\bibitem[{Sun et~al(2013)Sun, Wang, and Tang}]{CVPR2013Sun}
Sun Y, Wang X, Tang X (2013) Deep convolutional network cascade for facial
  point detection. In: Proceedings of IEEE Conference on Computer Vision and
  Pattern Recognition, pp 3476--3483

\bibitem[{Sung and Kim(2008)}]{ASMCA2008Sung}
Sung J, Kim D (2008) Pose-robust facial expression recognition using view-based
  2d+3d {AAM}. IEEE Transactions on Systems, Man and Cybernetics, Part A:
  Systems and Humans 38(4):852--866

\bibitem[{Sung et~al(2007)Sung, Kanade, and Kim}]{IJCV2007Sung}
Sung J, Kanade T, Kim D (2007) A unified gradient-based approach for combining
  {ASM} into {AAM}. International Journal of Computer Vision 75(2):297--309

\bibitem[{Sung et~al(2008)Sung, Kanade, and Kim}]{IJCV2008Sung}
Sung J, Kanade T, Kim D (2008) Pose robust face tracking by combining active
  appearance models and cylinder head models. International Journal of Computer
  Vision 80(2):260--274

\bibitem[{Tong et~al(2009)Tong, Liu, Wheeler, and Tu}]{CVPR2009Tong}
Tong Y, Liu X, Wheeler F, Tu P (2009) Automatic facial landmark labeling with
  minimal supervision. In: Proceedings of IEEE International Conference on
  Computer Vision and Pattern Recognition, pp 2097--2104

\bibitem[{Tong et~al(2012)Tong, Liu, Wheeler, and Tu}]{JACVIU2012Tong}
Tong Y, Liu X, Wheeler F, Tu P (2012) Semi-supervised facial landmark
  annotation. Computer Vision and Image Understanding 116(8):922--935

\bibitem[{De~la Torre and Nguyen(2008)}]{CVPR2008De}
De~la Torre F, Nguyen M (2008) Parameterized kernel principal component
  analysis: theory and applications to supervised and unsupervised image
  alignment. In: Proceedings of IEEE International Conference on Computer
  Vision and Pattern Recognition, pp 1--8

\bibitem[{Tresadern et~al(2009)Tresadern, Bhaskar, Adeshina, Taylor, and
  Cootes}]{BMVC2009Tresadern}
Tresadern P, Bhaskar H, Adeshina S, Taylor C, Cootes T (2009) Combining local
  and global shape models for deformable object matching. In: Proceedings of
  British Machine Vision Conference, pp 1--12

\bibitem[{Tresadern et~al(2010)Tresadern, Sauer, and
  Cootes}]{BMVC2010Tresadern}
Tresadern P, Sauer P, Cootes T (2010) Additive update predictors in active
  appearance models. In: Proceedings of British Machine Vision Conference, pp
  1--12

\bibitem[{Tresadern et~al(2012)Tresadern, Ionita, and
  Cootes}]{IJCV2012Tresadern}
Tresadern P, Ionita M, Cootes T (2012) Real-time facial feature tracking on a
  mobile device. International Journal of Computer Vision 96(3):280--289

\bibitem[{Tzimiropoulos and Pantic(2013)}]{ICCV2013Tzimiropoulos}
Tzimiropoulos G, Pantic M (2013) Optimization problems for fast {AAM} fitting
  in-the-wild. In: Proceedings of IEEE International Conference on Computer
  Vision, pp 593--600

\bibitem[{Tzimiropoulos et~al(2011)Tzimiropoulos, Zafeiriou, and
  Pantic}]{ICCV2011Tzimiropoulos}
Tzimiropoulos G, Zafeiriou S, Pantic M (2011) Robust and efficient parametric
  face alignment. In: Proceedings of IEEE International Conference on Computer
  Vision, pp 1847--1854

\bibitem[{Tzimiropoulos et~al(2012)Tzimiropoulos, Alabort-i Medina, Zafeiriou,
  and Pantic}]{ACCV2012Tzimiropoulos}
Tzimiropoulos G, Alabort-i Medina J, Zafeiriou S, Pantic M (2012) Generic
  active appearance models revisited. In: Proceedings of Asian Conference on
  Computer Vision, pp 650--663

\bibitem[{Valstar and Pantic(2010)}]{DB2010Valstar}
Valstar M, Pantic M (2010) Induced disgust, happiness and surprise: an addition
  to the {MMI} facial expression database. In: Proceedings of International
  Conference on Language Resources and Evaluation, Workshop EMOTION, pp 65--70

\bibitem[{Valstar et~al(2010)Valstar, Martinez, Binefa, and
  Pantic}]{CVPR2010Valstar}
Valstar M, Martinez B, Binefa X, Pantic M (2010) Facial point detection using
  boosted regression and graph models. In: Proceedings of IEEE International
  Conference on Computer Vision and Pattern Recognition, pp 2729--2736

\bibitem[{Viola and Jones(2004)}]{RPIJCV2004Viola}
Viola P, Jones M (2004) Robust real-time face detection. International Journal
  of Computer Vision 57(2):137--154

\bibitem[{Vogler et~al(2007)Vogler, Li, and Kanaujia}]{ICCV2007Vogler}
Vogler C, Li Z, Kanaujia A (2007) The best of both worlds: combining 3d
  deformable models with active shape models. In: Proceedings of IEEE
  International Conference on Computer Vision, pp 1--7

\bibitem[{Vukadinovic and Pantic(2005)}]{PBSMC2005Vukadinovic}
Vukadinovic D, Pantic M (2005) Fully automatic facial feature point detection
  using {Gabor} feature based boosted classifiers. In: Proceedings of
  International Conference on Systems, Man, and Cybernetics, pp 1692--1698

\bibitem[{Wang et~al(2014)Wang, Tao, Gao, Li, and Li}]{AIJCV2013Wang}
Wang N, Tao D, Gao X, Li X, Li J (2014) A comprehensive survey to face
  hallucination. International Journal of Computer Vision 106(1):9--30

\bibitem[{Wang et~al(2008{\natexlab{a}})Wang, Lucey, and Cohn}]{CVPR2008Wang}
Wang Y, Lucey S, Cohn J (2008{\natexlab{a}}) Enforcing convexity for improved
  alignment with constrained local models. In: Proceedings of IEEE
  International Conference on Computer Vision and Pattern Recognition, pp 1--8

\bibitem[{Wang et~al(2008{\natexlab{b}})Wang, Lucey, Cohn, and
  Saragih}]{FG2008Wang}
Wang Y, Lucey S, Cohn J, Saragih J (2008{\natexlab{b}}) Non-rigid face tracking
  with local appearance consistency constraint. In: Proceedings of
  Inteernational Conference on Automatic Face and Gesture Recognition, pp 1--8

\bibitem[{Weise et~al(2011)Weise, Bouaziz, Li, and Pauly}]{ASIGGRAPH2011Weise}
Weise T, Bouaziz S, Li H, Pauly M (2011) Realtime performance-based facial
  animation. In: Proceedings of SIGGRAPH, pp 77.1--77.9

\bibitem[{Wimmer et~al(2008)Wimmer, Stulp, Pietzsch, and
  Radig}]{PAMI2008Wimmer}
Wimmer M, Stulp F, Pietzsch S, Radig B (2008) Learning local objective
  functions for robust face model fitting. IEEE Transactions on Pattern
  Analysis and Machine Intelligence 30(8):1357--1370

\bibitem[{Wu et~al(2008)Wu, Liu, and Doretto}]{CVPR2008Wu}
Wu H, Liu X, Doretto G (2008) Face alignment via boosted ranking model. In:
  Proceedings of IEEE International Conference on Computer Vision and Pattern
  Recognition, pp 1--8

\bibitem[{Wu et~al(2013)Wu, Wang, and Ji}]{CVPR2013Wu}
Wu Y, Wang Z, Ji Q (2013) Facial feature tracking under varying facial
  expressions and face poses based on restricted {Boltzmann} machine. In:
  Proceedings of IEEE Conference on Computer Vision and Pattern Recognition, pp
  3452--3459

\bibitem[{Xiao et~al(2004)Xiao, Baker, Matthews, and Kanade}]{PBCVPR2004Xiao}
Xiao J, Baker S, Matthews I, Kanade T (2004) Real-time combined {2D}+{3D}
  active appearance models. In: Proceedings of IEEE Conference on Computer
  Vision and Pattern Recogntion, pp 535--542

\bibitem[{Xiong and De~la Torre(2013)}]{CVPR2013Xiong}
Xiong X, De~la Torre F (2013) Supervised descent method and its application to
  face alignment. In: Proceedings of IEEE Conference on Computer Vision and
  Pattern Recognition, pp 532--539

\bibitem[{Yang and Patras(2012)}]{ACCV2012Yang}
Yang H, Patras I (2012) Face parts localization using structured output
  regression forests. In: Proceedings of Asian Conference on Computer Vision,
  pp 667--679

\bibitem[{Yang and Patras(2013)}]{ICCV2013Yang}
Yang H, Patras I (2013) Sieving regression forest votes for facial feature
  detection in the wild. In: Proceedings of IEEE International Conference on
  Computer Vision, pp 1936--1943

\bibitem[{Yang et~al(2002)Yang, Kriegman, and Ahuja}]{RPPAMI2002Yang}
Yang M, Kriegman D, Ahuja N (2002) Detecting faces in images: a survey. IEEE
  Transactions on Pattern Analysis and Machine Intelligence 24(1):34--58

\bibitem[{Yu et~al(2013)Yu, Huang, Zhuang, Yan, and Metaxas}]{ICCV2013Yu}
Yu X, Huang J, Zhuang S, Yan W, Metaxas D (2013) Pose-free facial landmark
  fitting via optimized part mixtures and cascaded deformable shape model. In:
  Proceedings of IEEE International Conference on Computer Vision, pp
  1944--1951

\bibitem[{Zhang et~al(2009)Zhang, Liu, Poel, and Nijholt}]{ACCV2009Zhang}
Zhang H, Liu D, Poel M, Nijholt A (2009) Face alignment using boosting and
  evolutionary search. In: Proceedings of Asian Conference on Computer Vision,
  pp 110--119

\bibitem[{Zhao et~al(2011)Zhao, Cham, and Wang}]{JACVPR2011Zhao}
Zhao C, Cham W, Wang X (2011) Joint face alignment with a generic deformable
  face model. In: Proceedings of IEEE Conference on Computer Vision and Pattern
  Recognition, pp 561--568

\bibitem[{Zhao et~al(2012)Zhao, Chai, and Shan}]{JAECCV2012Zhao}
Zhao X, Chai X, Shan S (2012) Joint face alignment: rescue bad alignments with
  good ones by regularized re-fitting. In: Proceedings of European Conference
  on Computer Vision, pp 616--630

\bibitem[{Zhao et~al(2013)Zhao, Shan, Chai, and Chen}]{ICCV2013Zhao}
Zhao X, Shan S, Chai X, Chen X (2013) Cascaded shape space pruning for robust
  facial landmark detection. In: Proceedings of IEEE International Conference
  on Computer Vision, pp 1033--1040

\bibitem[{Zheng et~al(2006)Zheng, Zhou, Georgescu, Zhou, and
  Comaniciu}]{ECCV2006Zheng}
Zheng Y, Zhou X, Georgescu B, Zhou S, Comaniciu D (2006) Example based
  non-rigid shape detection. In: Proceedings of European Conference on Computer
  Vision, pp 423--436

\bibitem[{Zhou and Comaniciu(2007)}]{IPMI2007Zhou}
Zhou S, Comaniciu D (2007) Shape regression machine. In: Information Proceeding
  in Medical Imaging, pp 13--25

\bibitem[{Zhou et~al(2005)Zhou, Comaniciu, and Gupta}]{APAMI2005Zhou}
Zhou X, Comaniciu D, Gupta A (2005) An information fusion framework for robust
  shape tracking. IEEE Transactions on Pattern Analysis and Machine
  Intelligence 27(1):115--129

\bibitem[{Zhou et~al(2003)Zhou, Gu, and Zhang}]{PBCVPR2003Zhou}
Zhou Y, Gu L, Zhang H (2003) Bayesian tangent shape model: estimating shape and
  pose parameters via {Bayesian} inference. In: Proceedings of IEEE Conference
  on Computer Vision and Pattern Recognition, pp 109--116

\bibitem[{Zhu and Martinez(2006)}]{RPPAMI2006Zhu}
Zhu M, Martinez A (2006) Subclass discriminant analysis. IEEE Transactions on
  Pattern Analysis and Machine Intelligence 28(8):1274--1286

\bibitem[{Zhu and Ramanan(2012)}]{CVPR2012Zhu}
Zhu X, Ramanan D (2012) Face detection, pose estimation, and landmark
  localization in the wild. In: Proceedings of IEEE Conference on Computer
  Vision and Pattern Recognition, pp 2879--2886

\end{thebibliography}

%
%

\end{document}